\documentclass[accepted]{uai2026} %

\usepackage[american]{babel}

\usepackage[utf8]{inputenc} %
\usepackage[T1]{fontenc}    %
\usepackage{hyperref}       %
\usepackage{url}            %
\usepackage{booktabs}       %
\usepackage{amsfonts}       %
\usepackage{nicefrac}       %
\usepackage{microtype}      %
\usepackage{xcolor}         %

\usepackage{natbib} %
\usepackage{mathtools} %
\usepackage{booktabs} %
\usepackage{tikz} %
\usetikzlibrary{arrows.meta} %
\usepackage{algorithm}
\usepackage{graphicx}
\usepackage{amsmath}
\usepackage{amssymb}
\usepackage{mathtools}
\usepackage{amsthm}

\usepackage{algpseudocode}
\usepackage{xspace}
\usepackage{tcolorbox}

\DeclareMathOperator*{\argmax}{arg\,max}

\newcommand{\division}{\mkern-\medmuskip\rotatebox[origin=c]{45}{\scalebox{0.9}{$-$}}\mkern-\medmuskip}
\newcommand{\methodname}{Active-MoSH\xspace}
\newcommand{\trustmethod}{C-MoSH\xspace}
\newcommand{\methodfullname}{Active-C-MoSH\xspace}
\newtheorem{theorem}{Theorem}

\title{Interactive Multi-Objective Probabilistic Preference \\ Learning with Soft and Hard Bounds}

\author[1]{Edward Chen}
\author[1]{Sang T. Truong}
\author[1]{Natalie Dullerud}
\author[1]{Sanmi Koyejo}
\author[1]{Carlos Guestrin}
\affil[1]{%
    Department of Computer Science\\
    Stanford University\\
    Stanford, California, USA
}
  
\begin{document}
\maketitle

\begin{abstract}
High-stakes decision-making involves navigating multiple competing objectives with expensive evaluations. For instance, in brachytherapy, clinicians must balance maximizing tumor coverage (e.g., an aspirational target or \textit{soft bound} of $>$95\% coverage) against strict organ dose limits (e.g., a non-negotiable \textit{hard bound} of $<$601~cGy to the bladder). Selecting Pareto-optimal solutions that match implicit preferences is challenging, as exhaustive Pareto frontier exploration is computationally and cognitively prohibitive, necessitating interactive frameworks to guide users. While decision-makers (DMs) often possess domain knowledge to narrow the search via such soft-hard bounds, current methods often lack systematic approaches to iteratively refine these multi-faceted preference structures. Furthermore, DMs often require confidence that they have not overlooked superior alternatives, a paramount necessity in high-stakes scenarios. We present \methodname, an interactive local-global framework designed for this process. Its \textit{local component} integrates probabilistic preference learning with an active sampling strategy to adaptively refine Pareto subsets while minimizing cognitive burden. To bolster decision confidence, \methodname's \textit{global component}, \trustmethod, leverages multi-objective sensitivity analysis to identify potentially overlooked, high-value points beyond immediate feedback. We demonstrate \methodname's performance benefits through diverse synthetic and real-world applications. A high-stakes case study with real cervical cancer brachytherapy treatment plans and an image selection user study further validate our hypotheses regarding the framework's ability to improve convergence, enhance DM confidence, and provide expressive preference articulation.
\end{abstract}

\section{Introduction}\label{sec:intro}

Countless critical decision-making tasks within domains like healthcare and engineering design involve finding optimal tradeoffs among multiple competing, often expensive-to-evaluate, objectives $f_1(x),...,f_L(x)$ \cite{fromer_computer-aided_2023, luukkonen_artificial_2023, xie_mars_2021, yu_multi-objective_2000, papadimitriou_multiobjective_2001}. Decision-makers (DMs) typically seek a Pareto-optimal (PO) point that aligns with their hidden preferences. However, exhaustively searching the Pareto frontier is often infeasible due to computational costs, lengthy validation, and limited human attention. Consequently, DMs usually navigate the frontier interactively, validating subsets of points — often defined by initial, possibly flexible, constraints or target regions — to find their ideal solution \cite{paria_flexible_2019, ozaki_multi-objective_2023, wilde_learning_2022}.

In healthcare, for instance, brachytherapy planning for cervical cancer treatment requires clinicians to balance maximizing radiation to tumors against minimizing exposure to healthy organs \cite{deufel_pnav_2020}. The vast tradeoff space and time-consuming nature of plan evaluation (potentially hours to densely populate the tradeoff space) make direct exploration impractical, especially while the patient is under anesthesia \cite{cui_multi-criteria_2018, cui_multi-criteria_2018-1}. Crucially, clinicians often possess significant prior knowledge to define initial regions of interest. Rather than just as values to be maximized or minimized, this prior knowledge often comes in the form of both desired targets (or soft bounds) and strict, non-negotiable boundaries (or hard bounds). For example, a clinician may aim for a high level of tumor coverage (soft bound: $>95\%$) while also needing to ensure that coverage does not fall below a critical minimum threshold (hard bound: $>90\%$), and that radiation to a healthy organ like the bladder remains strictly below a maximum (hard bound: $<601$ centigrays (cGY)) and preferred dose (soft bound: $<513$ cGY) \cite{chen_modeling_2026}. As clinicians observe presented treatment plans and gain new insights into achievable tradeoffs, they iteratively refine these initial preferences. For instance, if achieving the $>95\%$ tumor coverage consistently results in plans where the bladder dose approaches or exceeds 601 cGY, the clinician may directly adjust their minimum or desired tumor coverage downwards to explore solutions that offer a more viable and safe balance. This iterative adjustment of their articulated goals is central to navigating complex decision spaces. A critical challenge, however, is building DM confidence: DMs must be confident that they have not overlooked potentially superior PO points outside of their current tradeoff space. While multi-objective preference learning has advanced, methods explicitly supporting such iterative refinement using expressive, multi-faceted preference structures and robust confidence-building within Pareto frontier subsets remain underexplored.

This paper formalizes this interactive process through \methodname, a novel framework with distinct \textit{local and global} components to efficiently guide DMs to optimal solutions while fostering confidence in their final choice. The \textit{local component} of \methodname iteratively refines the DM's search within the tradeoff space. To effectively capture and adapt to nuanced DM preferences (often comprising aspirational targets and strict limits), it extends the soft-hard utility function (SHF) concept \cite{chen_modeling_2026} to a dynamic, interactive setting. This SHF-based approach enables probabilistic modeling of the DM's latent preference vector and their evolving soft-hard bounds, thereby adaptively refining the explored Pareto subset and focusing computational resources on preference-aligned regions. An integrated active sampling strategy further optimizes this local search by balancing exploration-exploitation and minimizing computational/cognitive load. Addressing potential DM doubts about unobserved superior alternatives, the \textit{global component}, \trustmethod, enhances decision confidence. By leveraging multi-objective sensitivity analysis, \trustmethod systematically quantifies solution robustness to stated bounds and proactively identifies potentially overlooked, high-value Pareto regions \textit{beyond immediate DM feedback}, thereby building strong DM confidence. We rigorously validate \methodname via simulations on synthetic benchmarks and real-world applications, including a brachytherapy planning case study utilizing three real patient cases. Furthermore, a user study involving AI-generated image selection for educational magazine covers statistically confirms our hypotheses regarding the framework's improved efficiency, decision-making effectiveness, and ability to bolster decision confidence.

Although there exists extensive work on interactive multi-objective decision-making (MODM), many methods focus on feedback mechanisms that may not fully support intuitive exploration of specific tradeoff sub-regions \cite{wilde_learning_2022, biyik_asking_2020, astudillo_multi-attribute_2020, ozaki_multi-objective_2023}. While some recent approaches allow DMs to impose priors on Pareto subsets, they lack emphasis on the interactive feedback loop with expressive bound structures like soft-hard bounds, and the crucial aspect of building DM confidence within this dynamic context \cite{paria_flexible_2019, malkomes_beyond_2021}. In summary, our contributions comprise the following:

\begin{enumerate}
    \item \textbf{\methodname}: An interactive framework with a \textit{local component} that helps DMs iteratively refine preferred Pareto regions using probabilistic preference learning over soft-hard bounds and active sampling.
    \item \textbf{\trustmethod}: The \textit{global component} of \methodname, which builds DM confidence via multi-objective sensitivity analysis to quantify robustness and proactively uncover overlooked optimal regions.
    \item \textbf{Empirical Validation}: Rigorous evaluation across synthetic problems, a high-stakes brachytherapy case study, and a user study on AI-generated images, demonstrating statistically significant improvements in decision efficiency and confidence.
\end{enumerate}

\usetikzlibrary{arrows.meta, positioning, calc, backgrounds, fit, decorations.pathmorphing, shapes.geometric, patterns}

\definecolor{paretoBlue}{RGB}{44,107,176}
\definecolor{softGreen}{RGB}{46,139,87}
\definecolor{hardRed}{RGB}{192,47,47}
\definecolor{sampleOrange}{RGB}{230,140,30}
\definecolor{idealGold}{RGB}{218,165,32}
\definecolor{localBg}{RGB}{228,239,253}
\definecolor{globalBg}{RGB}{253,237,220}
\definecolor{panelBorder}{RGB}{180,180,190}
\definecolor{arrowGray}{RGB}{100,100,110}
\definecolor{trustPurple}{RGB}{128,90,170}

\begin{figure*}[t]
\centering
\begin{tikzpicture}[
    >=Stealth,
    every node/.style={font=\small},
    panel/.style={
        rectangle, rounded corners=5pt,
        draw=panelBorder, line width=0.6pt,
        fill=white, inner sep=0pt,
        minimum width=3.3cm, minimum height=3.9cm
    },
    panelTitle/.style={
        font=\footnotesize\bfseries, anchor=north,
        inner sep=2pt, text=black!85
    },
    mainArr/.style={
        -{Stealth[length=5pt,width=4pt]},
        line width=1pt, color=arrowGray
    },
    loopArr/.style={
        -{Stealth[length=4pt,width=3.5pt]},
        line width=0.8pt, color=paretoBlue!80, densely dashed
    },
    compBox/.style={
        rectangle, rounded corners=8pt,
        draw=#1!50, line width=0.9pt,
        fill=#1, inner sep=5pt
    },
    tinylbl/.style={font=\scriptsize},
    microLbl/.style={font=\fontsize{5.5}{6.5}\selectfont}
]

\node[panel] (panelA) at (-0.4,0) {};
\node[panelTitle] at ([yshift=-3pt]panelA.north) {{Probabilistic Modeling}};

\begin{scope}[shift={($(panelA.south west) + (0.45, 0.53)$)}]
    \draw[->, thin, black!70] (0,0) -- (2.4,0) node[right, tinylbl] {$f_1$};
    \draw[->, thin, black!70] (0,0) -- (0,2.6) node[above, tinylbl] {$f_2$};

    \fill[softGreen!15] (1.0,0) rectangle (2.4,1.55);

    \draw[very thick, paretoBlue, opacity=0.85]
        plot[smooth, tension=0.7]
        coordinates {(0.2,2.4) (0.5,2.0) (0.85,1.55) (1.25,1.1) (1.65,0.7) (2.0,0.45) (2.3,0.3)};

    \draw[hardRed, line width=1pt] (0.4,0) -- (0.4,2.6);
    \draw[hardRed, line width=1pt] (0,2.3) -- (2.4,2.3);

    \draw[softGreen, line width=1pt, dashed] (1.0,0) -- (1.0,2.6);
    \draw[softGreen, line width=1pt, dashed] (0,1.55) -- (2.4,1.55);

    \foreach \x/\y in {0.6/1.9, 0.95/1.45, 1.35/1.0, 1.75/0.6, 2.1/0.4} {
        \fill[paretoBlue!70] (\x,\y) circle (1.3pt);
    }

    \draw[hardRed!80, fill=hardRed!18, line width=0.3pt]
        plot[domain=0.05:0.75, samples=30, smooth]
        (\x, {-0.32*exp(-(\x-0.4)^2/(2*0.018))})
        -- (0.75,0) -- (0.05,0) -- cycle;
    \node[microLbl, hardRed, anchor=north] at (0.3,-0.18) {$p(\alpha_H)$};

    \draw[softGreen!70!black, fill=softGreen!15, line width=0.3pt]
        plot[domain=0.65:1.35, samples=30, smooth]
        (\x, {-0.32*exp(-(\x-1.0)^2/(2*0.018))})
        -- (1.35,0) -- (0.65,0) -- cycle;
    \node[microLbl, softGreen!80!black, anchor=north] at (1.0,-0.18) {$p(\alpha_S)$};

    \draw[softGreen!70!black, fill=softGreen!15, line width=0.3pt]
        plot[domain=1.2:1.9, samples=30, smooth]
        ({-0.32*exp(-(\x-1.55)^2/(2*0.018))}, \x)
        -- (0,1.9) -- (0,1.2) -- cycle;
    \node[microLbl, softGreen!80!black, anchor=east] at (-0.31,1.55) {$p(\alpha_S)$};

    \draw[hardRed!80, fill=hardRed!18, line width=0.3pt]
        plot[domain=1.95:2.6, samples=30, smooth]
        ({-0.32*exp(-(\x-2.3)^2/(2*0.018))}, \x)
        -- (0,2.6) -- (0,1.95) -- cycle;
    \node[microLbl, hardRed, anchor=east] at (-0.31,2.3) {$p(\alpha_H)$};

    \node[microLbl, softGreen!50!black, fill=white, inner sep=1pt,
          rounded corners=1pt] (focusLbl) at (2.15,2.1) {Ideal Region};
    \draw[->, softGreen!50!black, line width=0.4pt]
        (focusLbl.south) -- (1.8,1.2);

    \node[microLbl, paretoBlue!90!black, fill=white, inner sep=1pt,
          rounded corners=1pt, anchor=north west] at (1.7,0.28) {Pareto};
\end{scope}

\node[font=\fontsize{7}{8.5}\selectfont, text=black!60, anchor=north] at ([yshift=0pt]panelA.south) {DM sets initial bounds};

\node[panel] (panelB) at (3.7,0) {};
\node[panelTitle] at ([yshift=-3pt]panelB.north) {Active Sampling};

\begin{scope}[shift={($(panelB.south west) + (0.45, 0.53)$)}]
    \draw[->, thin, black!70] (0,0) -- (2.4,0) node[right, tinylbl] {$f_1$};
    \draw[->, thin, black!70] (0,0) -- (0,2.6) node[above, tinylbl] {$f_2$};

    \fill[softGreen!12] (1.0,0) rectangle (2.4,1.55);

    \draw[paretoBlue, opacity=0.3, line width=0.8pt]
        plot[smooth, tension=0.7]
        coordinates {(0.2,2.4) (0.5,2.0) (0.85,1.55) (1.25,1.1) (1.65,0.7) (2.0,0.45) (2.3,0.3)};

    \draw[hardRed, line width=0.5pt, opacity=0.4] (0.4,0) -- (0.4,2.6);
    \draw[hardRed, line width=0.5pt, opacity=0.4] (0,2.3) -- (2.4,2.3);
    \draw[softGreen, line width=0.5pt, dashed, opacity=0.4] (1.0,0) -- (1.0,2.6);
    \draw[softGreen, line width=0.5pt, dashed, opacity=0.4] (0,1.55) -- (2.4,1.55);

    \foreach \x/\y in {0.6/1.9, 0.95/1.45, 1.35/1.0, 1.75/0.6, 2.1/0.4} {
        \fill[paretoBlue!30] (\x,\y) circle (1.3pt);
    }

    \fill[sampleOrange] (1.2,1.15) circle (2.2pt);
    \fill[sampleOrange] (1.5,0.85) circle (2.2pt);
    \fill[sampleOrange] (1.8,0.55) circle (2.2pt);

    \draw[->, sampleOrange!80, line width=0.6pt] (0.35,0.5) to[out=25,in=200] (1.13,1.07);
    \draw[->, sampleOrange!80, line width=0.6pt] (0.35,0.3) to[out=15,in=215] (1.43,0.77);

    \node[microLbl, sampleOrange!80!black, fill=white, inner sep=1pt,
          rounded corners=1pt] at (1.75,1.45) {Queries};
\end{scope}

\node[font=\fontsize{7}{8.5}\selectfont, text=black!60, anchor=north] at ([yshift=0pt]panelB.south) {Select informative queries};

\node[panel] (panelC) at (7.8,0) {};
\node[panelTitle] at ([yshift=-3pt]panelC.north) {DM Feedback};

\begin{scope}[shift={($(panelC.south west) + (0.45, 0.53)$)}]
    \draw[->, thin, black!70] (0,0) -- (2.4,0) node[right, tinylbl] {$f_1$};
    \draw[->, thin, black!70] (0,0) -- (0,2.6) node[above, tinylbl] {$f_2$};

    \fill[softGreen!15] (0.8,0) rectangle (2.4,1.8);

    \draw[paretoBlue, opacity=0.3, line width=0.8pt]
        plot[smooth, tension=0.7]
        coordinates {(0.2,2.4) (0.5,2.0) (0.85,1.55) (1.25,1.1) (1.65,0.7) (2.0,0.45) (2.3,0.3)};

    \draw[hardRed, line width=0.4pt, opacity=0.25] (0.4,0) -- (0.4,2.6);
    \draw[softGreen, line width=0.4pt, dashed, opacity=0.25] (1.0,0) -- (1.0,2.6);
    \draw[hardRed, line width=0.4pt, opacity=0.25] (0,2.3) -- (2.4,2.3);
    \draw[softGreen, line width=0.4pt, dashed, opacity=0.25] (0,1.55) -- (2.4,1.55);

    \draw[hardRed, line width=1pt] (0.25,0) -- (0.25,2.6);
    \draw[softGreen, line width=1pt, dashed] (0.8,0) -- (0.8,2.6);
    \draw[hardRed, line width=1pt] (0,2.45) -- (2.4,2.45);
    \draw[softGreen, line width=1pt, dashed] (0,1.8) -- (2.4,1.8);

    \draw[-{Stealth[length=4pt,width=3pt]}, hardRed, line width=1.2pt]
        (0.4,2.55) -- (0.25,2.55);
    \draw[-{Stealth[length=4pt,width=3pt]}, softGreen!80!black, line width=1.2pt]
        (1.0,2.55) -- (0.8,2.55);
    \draw[-{Stealth[length=4pt,width=3pt]}, hardRed, line width=1.2pt]
        (2.3,2.3) -- (2.3,2.45);
    \draw[-{Stealth[length=4pt,width=3pt]}, softGreen!80!black, line width=1.2pt]
        (2.3,1.55) -- (2.3,1.8);

    \fill[sampleOrange!50] (1.2,1.15) circle (1.5pt);
    \fill[sampleOrange!50] (1.5,0.85) circle (1.5pt);
    \fill[sampleOrange!50] (1.8,0.55) circle (1.5pt);

    \draw[hardRed!80, fill=hardRed!18, line width=0.3pt]
        plot[domain=0.1:0.4, samples=25, smooth]
        (\x, {-0.32*exp(-(\x-0.25)^2/(2*0.006))})
        -- (0.4,0) -- (0.1,0) -- cycle;
    \node[microLbl, hardRed, anchor=north] at (0.13,-0.18) {$p_m(\alpha_H)$};

    \draw[softGreen!70!black, fill=softGreen!15, line width=0.3pt]
        plot[domain=0.55:1.05, samples=25, smooth]
        (\x, {-0.32*exp(-(\x-0.8)^2/(2*0.006))})
        -- (1.05,0) -- (0.55,0) -- cycle;
    \node[microLbl, softGreen!80!black, anchor=north] at (1.1,-0.18) {$p_m(\alpha_S)$};

    \draw[softGreen!70!black, fill=softGreen!15, line width=0.3pt]
        plot[domain=1.55:2.05, samples=25, smooth]
        ({-0.32*exp(-(\x-1.8)^2/(2*0.006))}, \x)
        -- (0,2.05) -- (0,1.55) -- cycle;
    \node[microLbl, softGreen!80!black, anchor=east] at (-0.31,1.8) {$p_m(\alpha_S)$};

    \draw[hardRed!80, fill=hardRed!18, line width=0.3pt]
        plot[domain=2.2:2.6, samples=25, smooth]
        ({-0.32*exp(-(\x-2.45)^2/(2*0.006))}, \x)
        -- (0,2.6) -- (0,2.2) -- cycle;
    \node[microLbl, hardRed, anchor=east] at (-0.31,2.45) {$p_m(\alpha_H)$};

    \node[microLbl, black!70, fill=white, inner sep=1pt,
          rounded corners=1pt] at (1.6,1.55) {Bounds Shift};
\end{scope}

\node[font=\fontsize{7}{8.5}\selectfont, text=black!60, anchor=north] at ([yshift=0pt]panelC.south) {Refine soft-hard bounds};

\node[panel] (panelD) at (11.9,0) {};
\node[panelTitle, font=\fontsize{8.3}{9.3}\selectfont\bfseries] 
    at ([yshift=-3pt]panelD.north) {Confidence Enhancement};

\begin{scope}[shift={($(panelD.south west) + (0.45, 0.53)$)}]
    \draw[->, thin, black!70] (0,0) -- (2.4,0) node[right, tinylbl] {$f_1$};
    \draw[->, thin, black!70] (0,0) -- (0,2.6) node[above, tinylbl] {$f_2$};

    \draw[paretoBlue, opacity=0.4, line width=0.8pt]
        plot[smooth, tension=0.7]
        coordinates {(0.2,2.4) (0.5,2.0) (0.85,1.55) (1.25,1.1) (1.65,0.7) (2.0,0.45) (2.3,0.3)};

    \draw[hardRed, line width=0.5pt, opacity=0.5] (0.25,0) -- (0.25,2.6);
    \draw[softGreen, line width=0.5pt, dashed, opacity=0.5] (0.8,0) -- (0.8,2.6);
    \draw[hardRed, line width=0.5pt, opacity=0.5] (0,2.45) -- (2.4,2.45);
    \draw[softGreen, line width=0.5pt, dashed, opacity=0.5] (0,1.8) -- (2.4,1.8);

    \draw[trustPurple, line width=0.6pt, densely dotted, opacity=0.5]
        (1.25,1.05) ellipse (0.85 and 0.65);
    \fill[trustPurple, opacity=0.06]
        (1.25,1.05) ellipse (0.85 and 0.65);

    \draw[trustPurple, line width=0.7pt, opacity=0.7]
        (1.25,1.05) ellipse (0.45 and 0.35);
    \fill[trustPurple, opacity=0.1]
        (1.25,1.05) ellipse (0.45 and 0.35);

    \node[star, star points=5, star point ratio=2.3,
          fill=idealGold, draw=idealGold!70!black, line width=0.3pt,
          inner sep=1.2pt] at (1.25,1.05) {};

    \foreach \x/\y in {0.55/1.95, 0.85/1.55, 1.65/0.75, 2.0/0.5} {
        \fill[paretoBlue!25] (\x,\y) circle (1.3pt);
    }

    \node[microLbl, trustPurple!80!black] at (1.25,1.95) {Sensitivity};
    \node[microLbl, trustPurple!80!black] at (1.25,1.78) {Region};

    \draw[->, trustPurple!60, line width=0.4pt] (1.7,1.25) -- (2.1,1.55);
    \draw[->, trustPurple!60, line width=0.4pt] (0.65,0.75) -- (0.3,0.55);
\end{scope}

\node[font=\fontsize{7}{8.5}\selectfont, text=black!60, anchor=north] at ([yshift=0pt]panelD.south) {\trustmethod sensitivity check};

\draw[mainArr] ([xshift=3pt]panelA.east) -- ([xshift=-3pt]panelB.west);
\draw[mainArr] ([xshift=3pt]panelB.east) -- ([xshift=-3pt]panelC.west);
\draw[mainArr] ([xshift=3pt]panelC.east) -- ([xshift=-3pt]panelD.west);

\draw[loopArr]
    ([yshift=-8pt]panelC.south)
    to[out=-90, in=-90, looseness=0.6]
    node[midway, below, font=\fontsize{6}{7}\selectfont, text=paretoBlue!80!black] {Iterate}
    ([yshift=-7.5pt]panelB.south);

\begin{scope}[on background layer]
    \node[compBox=localBg, fit=(panelA)(panelB)(panelC),
          inner sep=10pt, yshift=-2pt] (localGroup) {};
\end{scope}
\node[font=\footnotesize\bfseries, text=paretoBlue!80!black, anchor=south]
    at ([yshift=1pt]localGroup.north) {\methodname: Local Component};

\begin{scope}[on background layer]
    \node[compBox=globalBg, fit=(panelD),
          inner sep=10pt, yshift=-2pt] (globalGroup) {};
\end{scope}
\node[font=\footnotesize\bfseries, text=sampleOrange!80!black, anchor=south]
    at ([yshift=1pt]globalGroup.north) {\trustmethod: Global Component};

\begin{scope}[shift={($(localGroup.south) + (0,-0.85)$)}]
    \def\lsp{2.7}  %

    \draw[very thick, paretoBlue, opacity=0.85] (-2*\lsp,0) -- ++(0.35,0);
    \node[tinylbl, anchor=west] at (-2*\lsp+0.4,0) {Pareto Frontier};

    \draw[hardRed, line width=0.9pt] (-1*\lsp,0) -- ++(0.35,0);
    \node[tinylbl, anchor=west] at (-1*\lsp+0.4,0) {Hard Bound};

    \draw[softGreen, line width=0.9pt, dashed] (0,0) -- ++(0.35,0);
    \node[tinylbl, anchor=west] at (0.4,0) {Soft Bound};

    \fill[sampleOrange] (1*\lsp+0.15,0) circle (2pt);
    \node[tinylbl, anchor=west] at (1*\lsp+0.4,0) {Candidate};

    \node[star, star points=5, star point ratio=2.3,
          fill=idealGold, draw=idealGold!70!black, line width=0.2pt,
          inner sep=0.9pt] at (2*\lsp+0.15,0) {};
    \node[tinylbl, anchor=west] at (2*\lsp+0.4,0) {Selected Point};
\end{scope}

\end{tikzpicture}
\caption{
    Overview of \methodname.
    \textbf{(Left three panels - Local Component):}
    The DM defines initial soft bounds (desired targets, dashed green) and hard bounds (strict limits, solid red) on objectives, narrowing the Pareto frontier to a focus region.
    \methodname actively selects informative queries (orange) within the soft-hard bound defined region.
    The DM observes these queries and adjusts bounds based on observed tradeoffs (arrows show bound shifts), iterating until convergence.
    \textbf{(Right panel - Global Component):}
    \trustmethod performs MOO sensitivity analysis around the current best solution (gold star), probing regions beyond stated bounds to confirm no superior alternatives were overlooked, thereby building DM confidence.
}
\label{overall_pipeline}
\end{figure*}

\section{Multi-Objective Preference Learning with Soft-Hard Bounds}

\subsection{Background}

Multi-objective optimization (MOO) seeks to jointly maximize $L$ expensive-to-query black-box objective functions $f_1(x),...,f_L(x)$ over an input space $X \subset \mathbb{R}^d$. Since solutions rarely optimize all objectives simultaneously, MOO focuses on identifying points on the Pareto frontier. Scalarization techniques, $s_{\pmb{\lambda}}(y): \mathbb{R}^L \rightarrow \mathbb{R}$, parameterized by $\pmb{\lambda} \in \Lambda$ representing relative preferences, combine objectives into a single value to find PO solutions \citep{roijers_survey_2013}.

Decision-makers (DMs) often have priors on desirable objective space regions, expressible as soft and hard bounds \citep{chen_modeling_2026}. A soft bound, $\alpha_{\ell, S}$, describes an ideal target for objective $f_\ell(x)$, while a hard bound, $\alpha_{\ell, H}$, defines a limit $f_\ell(x)$ must not violate. Since ideal targets $\alpha_{\ell, S}$ might be unreachable due to competing objectives, hard bounds are crucial. To operationalize these dual-level preferences, \citet{chen_modeling_2026} proposed \textit{soft-hard} utility functions (SHFs), $u_{\pmb{\alpha}}(x)$, which map each $f_{\ell}$ to a bounded utility space based on bounds $\{\alpha_{1, S}, \alpha_{1, H}, ..., \alpha_{L, S}, \alpha_{L, H}\}$. \textit{Additional details may be found in Appendix \ref{appendix:shfs}}.

Points $y \in Y$ on the SHF-defined Pareto frontier are found by solving
\begin{equation}\label{eq:shf-pareto-optimal}
    \max_{x \in X} s_{\pmb{\lambda}}(u_{\pmb{\alpha}}(x)),
\end{equation}
where $u_{\pmb{\alpha}} := [u_{\alpha_1}, ..., u_{\alpha_L}]$, $u_{\alpha_\ell}$ is the SHF for $f_\ell$ (parameterized by $\alpha_{\ell,S}, \alpha_{\ell,H}$), and $s_{\pmb{\lambda}}(u_{\pmb{\alpha}}(x))$ is assumed monotonically increasing in $u_{\pmb{\alpha}}(x)$.

\textbf{Novel Contributions.} While \citet{chen_modeling_2026} introduced the SHF framework for obtaining a single set of easily navigable PO points, extending the static SHF formulation to a dynamic, interactive setting is non-trivial. It necessitates a principled approach for modeling evolving decision-maker preferences under uncertainty, a mechanism to actively build confidence that superior solutions are not being overlooked, and a rigorous framework for evaluating the interactive system's effectiveness. This work introduces novel local-global components to systematically address these challenges.

\subsection{Problem Definition}\label{sec:problem-def}

\textbf{Problem Setting.}
As DM evaluation of each point $\pmb{y} \in Y$ is expensive, it is infeasible to observe the full $Y$ on the entire Pareto frontier. We assume the DM has a hidden ideal set of soft and hard bounds, \{$\pmb{\alpha}^*_S$, $\pmb{\alpha}^*_H$\} (abbreviating \{$\alpha^*_{1,S}$, ..., $\alpha^*_{L,S}$\} and \{$\alpha^*_{1,H}$, ..., $\alpha^*_{L,H}$\}), which best characterizes the set of tradeoffs most appropriate for their decision-making task. This implies a hidden ideal point $\pmb{y}^* = f(x^*)$ on the SHF-defined Pareto frontier, corresponding to the DM's hidden preferences $\pmb{\lambda}^*$. Formally, $x^* = \argmax_{x \in X }s_{\pmb{\lambda}^*}(u_{\pmb{\alpha}^*}(x))$, where $\pmb{y} = f(x)$.

\textbf{Feedback Mechanism.} We assume that $\pmb{\lambda}$ and the set of bounds $\{\alpha_{1,S}, \dots, \alpha_{L,H}\}$, abbreviated as $\{\pmb{\alpha}_S, \pmb{\alpha}_H\}$, are random variables. We aim to learn $\pmb{\lambda}^*$ and $\{\pmb{\alpha}^*_S, \pmb{\alpha}^*_H\}$ through $M$ iterations of interaction. At each iteration $m$, the DM provides feedback by specifying point values for soft and hard bounds, which we denote as observations $\pmb{\alpha}_{S,m}$ and $\pmb{\alpha}_{H,m}$. As exact numerical feedback is often imprecise, we express uncertainty by modeling these specified values as noisy observations of the true latent bounds rather than fixed parameters. Specifically, we treat $\pmb{\alpha}_{S,m}$ and $\pmb{\alpha}_{H,m}$ as samples drawn from the true bound distributions with fixed observation noise $\Sigma_{\text{noise}}$. We maintain posterior beliefs, e.g., $p_m(\pmb{\alpha}_H) \sim \mathcal{N}(\pmb{\mu}_{H,m}, \Sigma_{H,m})$, and update the parameters $\{\pmb{\mu}_{H,m}, \Sigma_{H,m}\}$ using these observations.

The interactive process starts at iteration $m=0$: the DM inputs their prior distributions for the bounds, $p(\pmb{\alpha}_S)$ and $p(\pmb{\alpha}_H)$, by specifying an initial set of soft and hard bounds, \{$\pmb{\alpha}_{S,0}, \pmb{\alpha}_{H,0}$\}. For each subsequent iteration $m = 1, \dots, M$: (1) based on the current beliefs of $\pmb{\lambda}$ and $\pmb{\alpha}$, a set of $k$ PO points, $Y_m = \{\pmb{y}_1, ..., \pmb{y}_k\}$, is presented to the DM as a query, (2) the DM returns feedback $\pmb{\alpha}_{m}$ by adjusting one soft or hard bound for any of the objectives, and (3) the posteriors of $\pmb{\lambda}$ and $\pmb{\alpha}$ are then updated using Bayes' rule, incorporating the DM feedback.

Due to potentially complex relationships across the multiple objectives overwhelming the DM, we assume a single modification of any of the soft and hard bounds to be the full iteration. At the end of $M$ feedback iterations, we assume the DM selects any previously seen PO points from the $M$ preference queries. As a result, our overall objective is to optimize the following SHF Utility Ratio:
\begin{equation}\label{shf-utility-ratio}
    \mathbb{E} \Bigg[ \dfrac{\max_{x \in \{X_1 \cup ... \cup X_M\}} s_{\pmb{\lambda}}(u_{\pmb{\alpha}}(x))}{\max_{x \in X} s_{\pmb{\lambda}}(u_{\pmb{\alpha}}(x))} \Bigg]
\end{equation}
Intuitively, the SHF Utility Ratio is maximized when the points in $f(X_m)$ are PO and span the high utility regions of the PF, as defined by the SHFs $u_{\pmb{\alpha}}$. We use Eq. (\ref{shf-utility-ratio}) to evaluate our results in Section \ref{sec:experimental-results}. Figure \ref{overall_pipeline} illustrates this interactive process.

\section{\methodname}\label{sec:prob-modeling-sampling}

To learn the DM's ideal preferences $\pmb{\lambda}^*$ and bounds $\{\pmb{\alpha}^*_S, \pmb{\alpha}^*_H\}$, we maintain and iteratively update posterior probability distributions over these quantities. This probabilistic approach captures our uncertainty and informs the active sampling of preference queries $Y$, making up the local component of \methodname. \textit{More details App. \ref{appendix:prob-modeling-sampling}}.

\subsection{Probabilistic Modeling of Preferences and Bounds}\label{sec:prob-modeling}

\textbf{Preferences Modeling.} 
Given the DM's history of feedbacks, we update our belief over $\pmb{\lambda}$ using Bayes' rule:
\begin{equation}\label{eq:lambda-bayes}
    p_m(\pmb{\lambda}) \propto p(\pmb{\lambda}) \prod_m L(\pmb{\pi}_{m} ; X_m, Y_m, \pmb{\alpha}_{m}, \pmb{\lambda})
\end{equation}
$p_m(\pmb{\lambda})$ is the posterior distribution over $\pmb{\lambda}$ after $m$ feedback iterations and $p(\pmb{\lambda})$ is the uniform prior probability distribution over $\pmb{\lambda}$. The likelihood term $L(\pmb{\pi}_{m} ; X_m, Y_m, \pmb{\alpha}_{m}, \pmb{\lambda})$ represents the probability of the DM's feedback implying the ranking $\pmb{\pi}_m$ in response to the preference query $Y_m$ (Section \ref{sec:model-soft-hard-bounds}) and current bounds and parameters.

\textbf{Bounds Modeling.} As DMs provide feedback by imprecisely modifying soft-hard bounds, we update our posterior beliefs over the true ideal bounds $\pmb{\alpha}_S, \pmb{\alpha}_H$. We model the DM's specified bound at iteration $m$, denoted $\pmb{\alpha}_{H,m}$, as a noisy observation of the true bound $\pmb{\alpha}_H$, governed by the likelihood function $L(\pmb{\alpha}_{H,m} | \pmb{\alpha}_H) = \mathcal{N}(\pmb{\alpha}_{H,m} | \pmb{\alpha}_H, \Sigma_{\text{noise}})$. Assuming a conjugate Gaussian prior $p(\pmb{\alpha}_H) \sim \mathcal{N}(\pmb{\mu}_{H}, \Sigma_{H})$, the posterior is updated analytically via Bayes' rule (similar update applies to $p_m(\pmb{\alpha}_S)$):
\begin{equation}\label{eq:alpha-bayes}
    p_m(\pmb{\alpha}_H) \propto p(\pmb{\alpha}_H) \prod_{i=1}^m \mathcal{N}(\pmb{\alpha}_{H,i} | \pmb{\alpha}_H, \Sigma_{\text{noise}})
\end{equation}
$\Sigma_{\text{noise}}$ is the variance of the DM's feedback consistency.

\subsection{Modeling Soft-Hard Feedback}\label{sec:model-soft-hard-bounds}

\textbf{Feedback Interpretation.}
Our framework interprets DM soft-hard bounds feedback as implicit signal about their current region of interest on the Pareto frontier. This interpretation imposes a ranking over the set of query points and directly influences the likelihood model described in Section \ref{sec:prob-modeling}; it is consistent with reference-dependent decision making: changing a constraint/target shifts the DM’s reference point or aspiration/reservation levels, which is known to change relative evaluations and the induced preference ordering \citep{deb_reference_2006, kahneman_prospect_2012}. As with reference points \citep{vesikar_reference_2018}, the points in $Y_m$ closest to this newly defined ``edge'' or ``target'' of the feasible/desirable region are considered most actionable or informative, hence ranked highest. \textit{We provide detailed motivation for the specific ranking interpretations of all feedback scenarios in Appendix \ref{appendix:modeling-soft-hard-bounds}}.

\begin{figure}[hbt!]
    \centering
    \includegraphics[width=1\linewidth]{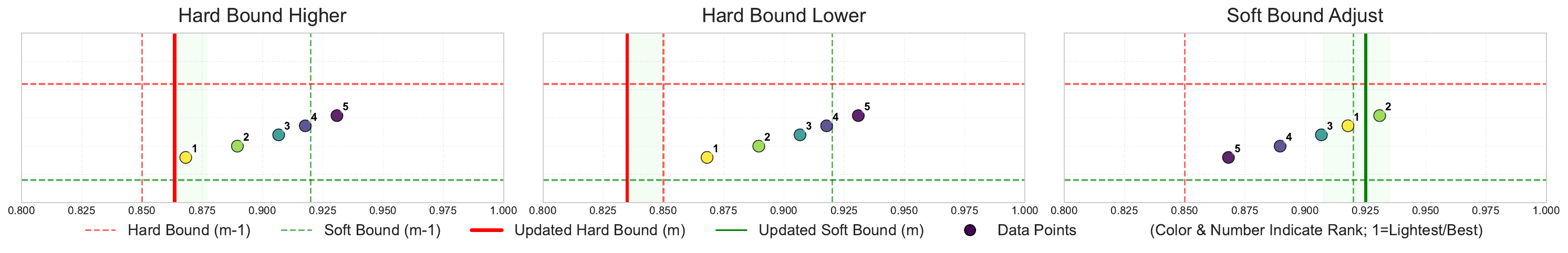}
    \caption{Bounds Feedback Interpretation. Adjustment of bounds imposes a ranking over the query points, determined by Euclidean distance from the shifted bound. The green shaded region represents actionable immediate tradeoff areas of interest implied by the feedback, which subsequently is where the distribution of preference vectors shifts towards.}
    \label{fig:feedback-interp}
\end{figure}

\textbf{Likelihood Model.}
As mentioned, we assume the DM's feedback (bound modification) implicitly ranks points in the query $Y_m$ based on their proximity to the newly adjusted bound, making those points more relevant. To operationalize the interpreted ranking, we use Euclidean distance from the shifted bound and leverage the Plackett-Luce model for the likelihood function in Eq.~\eqref{eq:lambda-bayes} \citep{luce_individual_1959}:
\begin{equation}
\label{eq:plackett_luce}
L(\pi ; X, Y, \pmb{\alpha}, \pmb{\lambda}) = \prod_{j=1}^{k} \frac{\exp\Bigl( s_{\pmb{\lambda}}( u_{\pmb{\alpha}}\bigl(X_{\pi(j)}\bigr)) \Bigr)}{ \sum_{h=j}^{k} \exp\Bigl( s_{\pmb{\lambda}} (u_{\pmb{\alpha}}\bigl(X_{\pi(h)}\bigr)) \Bigr)}
\end{equation}
where $\pi = \bigl[\pi(1), \dots, \pi(k)\bigr]$ is the DM-induced ranking of the $k$ points in $Y$, $X_{\pi(j)}$ is the input value for the point in $Y$ with rank $j$, and $u_{\pmb{\alpha}}$ is the SHF utility using the bounds.

\subsection{Active Query Sampling}\label{sec:active-sampling}

To query the DM effectively, we aim to select a navigable set of PO points $Y_m$ that reflects their current preferences and reduces cognitive load. The central challenge is our uncertainty over both the DM's latent preference vector ($\pmb{\lambda}^*$) and their ideal soft-hard bounds ($\pmb{\alpha}^*$). Distinct from static methods \citep{chen_modeling_2026}, our active sampling must be robust to this dual uncertainty. We therefore adapt a two-step process: (1) obtain a dense set of points by maximizing the expected SHF Utility Ratio (Eq. \ref{shf-utility-ratio}), where the expectation is over our posteriors of both $\pmb{\lambda}$ and $\pmb{\alpha}$, then (2) sparsify this set to present a small, diverse subset to the DM.

\textbf{Step 1: Dense Set of Points.} As the first step of the two-step process, we aim to obtain a dense and diverse set of candidate points $Y_{m_D}$ which maximizes the expected SHF Utility Ratio, while being robust to current uncertainty in the DM's preferences and soft-hard bounds. We model the expensive black-box objectives $f_\ell(x)$ with Gaussian Processes (GPs) \citep{williams_gaussian_1995}. Since $\pmb{\lambda}^*$ and $\pmb{\alpha}^*$ are both unknown to us, we want $Y_{m_D}$ to be robust to any potential $\pmb{\lambda}^*$ and $\pmb{\alpha}^*$, weighted according to their posterior distributions. As a result, we first aim to obtain a dense and diverse set of candidate points $Y_{m_D}$ by maximizing the expected SHF Utility Ratio (Eq.~\eqref{shf-utility-ratio} from Section~\ref{sec:problem-def}) over choices of $X_{m_D}$:

\begin{equation}\label{eq:bayesian-formulation}
    \max_{\substack{X_{m_D} \subseteq X, \\ |X_{m_D}| \leq k_\mathcal{D}}} 
    \mathbb{E}_{\substack{\pmb{\lambda} \sim p_m(\pmb{\lambda}), \\ {\pmb{\alpha}} \sim p_m(\pmb{\alpha})}}
    \Bigg[\dfrac{\max_{x \in X_{m_D}} s_{\pmb{\lambda}}(u_{{\pmb{\alpha}}}(x))}{\max_{x' \in X} s_{\pmb{\lambda}}(u_{{\pmb{\alpha}}}(x'))}\Bigg]
\end{equation}
where the expectation is over the current posterior distributions of $\pmb{\lambda}$ and $\pmb{\alpha}$.

To solve Equation (\ref{eq:bayesian-formulation}), we adapt the MoSH-Dense algorithm from \citet{chen_modeling_2026}. In short, our algorithm uses the notion of random scalarizations to sample a $\pmb{\lambda}$ and $\pmb{\alpha}$ from their posterior distributions at each iteration, which is then used to compute a multi-objective acquisition function \citep{paria_flexible_2019}. The maximizer of the acquisition function (Appendix \ref{appendix:active-sampling-for-queries}) is then chosen as the next sample input to be evaluated with the expensive black-box function, resulting in a single PO point. This is continued for a total of $T$ iterations, resulting in a dense set of $T$ PO points. For simplicity, we denote $X_{m_D}$ with $D$. The full algorithm is shown in Algorithm \ref{algorithm:mosh}.

\begin{algorithm}
\caption{Active-MoSH-Dense: Dense Sampling}
\begin{algorithmic}[1]

\Procedure{Active-MoSH-Dense}{}
    \State Initialize $D^{(0)} = \emptyset$
    \State Initialize $GP_{\ell}^{(0)} = GP(0, \kappa)$ $\forall$ $\ell \in [L]$
    \For{t = 1 $\rightarrow$ T}
        \State Obtain $\pmb{\lambda_t} \sim p_t(\pmb{\lambda})$ and $\pmb{\alpha_t} \sim p_t(\pmb{\alpha})$
        \State $x_t = \argmax_{x \in X} \text{acq}(u, \pmb{\lambda_t}, \pmb{\alpha_t}, x)$ \Comment{App. \ref{appendix:active-sampling-for-queries}}
        \State Obtain $y = f(x_t)$
        \State $D^{(t)} = D^{(t-1)} \cup \{(x_t, y)\}$
        \State $GP_\ell^{(t)} = \text{post}(GP_\ell^{t-1} | (x_t, y))$ $\forall$ $\ell \in [L]$
    \EndFor
    \State Return $D^{(T)}$
\EndProcedure
\end{algorithmic}
\label{algorithm:mosh}
\end{algorithm}

\textbf{Step 2: Sparse Set of Points.} To reduce the cognitive load on the DM, we then wish to sparsify the set of points $X_{m_D}$. We adapt Equation \ref{eq:bayesian-formulation} and formulate the sparsification as a robust submodular observation selection (RSOS) problem \citep{krause_robust_2008}:
\begin{equation}\label{submodular-formulation2}
    \max_{X_{m} \subseteq X_{m_D}, |X_{m}| \leq k} \min_{\pmb{\lambda} \in \Lambda}\underbrace{\Bigg[\dfrac{\max_{x \in X_{m}} s_{\pmb{\lambda}}(u_{{\pmb{\alpha}}}(x))}{\max_{x' \in X_{m_D}} s_{\pmb{\lambda}}(u_{{\pmb{\alpha}}}(x'))}\Bigg]}_{F_{\pmb{\lambda}}}
\end{equation}
where $X_{m_D}$ represents the set obtained from Active-MoSH-Dense, Algorithm \ref{algorithm:mosh}, and $f(X_{m})$ is the sparse set of SHF-defined PO points returned to the DM. To solve Equation \ref{submodular-formulation2}, we apply MoSH-Sparse, from \citet{chen_modeling_2026}, on $X_{m_D}$. MoSH-Sparse leverages the notion of submodularity, which encapsulates the concept of diminishing returns in utility for each additional point the DM validates \citep{chen_modeling_2026, nemhauser_analysis_1978, krause_robust_2008}. We leverage the proof from \citet{chen_modeling_2026} to show that the SHF Utility Ratio, $F_{\pmb{\lambda}}$, is a submodular function. As a result, MoSH-Sparse theoretically guarantees for our sampled preference queries to obtain a set of points $Y_{m}$ from $Y_{m_D}$ which achieves the optimal coverage of the unknown preferences, albeit with a slightly greater number of points than $k$. We provide the full details in Algorithm \ref{algorithm:saturate}. For simplicity of notation, we denote $X_m$ with $C$ and $X_{m_D}$ with $D$. $\psi$ is defined in Theorem \ref{krause-theorem}. Additional details of MoSH-Sparse may be found in \citet{chen_modeling_2026}.

\begin{algorithm}[H]
    \centering
    \caption{MoSH-Sparse: Pareto Sparsification}
    \begin{algorithmic}[1]
    \Procedure{MoSH-Sparse($F_1,...,F_{|\Lambda|}, k, \psi$)}{}
        \State $q_{min} = 0$; $q_{max} = \min_i F_i(D)$; $C_{best} = 0$
        \While{$(q_{max}-q_{min}) \geq 1 \division |\Lambda|$}
            \State $q = (q_{min}+q_{max})/2$
            \State Define $\Bar{F}_q(C) = 1 \division |\Lambda| \sum_i \min \{F_i(C), q\}$
            \State $\hat{C} = GPC(\Bar{F}_q, q)$ 
            \Comment{App. \ref{appendix:active-sampling-for-queries}: Algorithm \ref{algorithm:gpc}}
            \If{$|\hat{C}| > \psi k$}
                \State $q_{max} = q$
            \Else
                \State $q_{min} = q$; $C_{best} = \hat{C}$
            \EndIf
        \EndWhile
    \EndProcedure
    \end{algorithmic}
    \label{algorithm:saturate}
\end{algorithm}

\begin{theorem}\label{krause-theorem}~\citep{krause_robust_2008} For any integer k, MoSH-Sparse finds a solution $C_S$ such that $\min_i F_i(C_S) \geq \max_{|C| \leq k} \min_i F_i(C)$ and $|C_S| \leq \psi k$, for $\psi=1+\log(\max_{x \in D} \sum_i F_i(\{x\}))$. The total number of submodular function evaluations is $\mathcal{O}(|D|^2 m \log(m \min_i F_i(D)))$, where $m = |\Lambda|$.
\end{theorem}

\section{\trustmethod: Enhancing Confidence}\label{sec:t-mosh}

While the local component of \methodname focuses on refining the DM's immediate region of interest, DMs may still question if their current solution is truly optimal or if slight bound adjustments might reveal superior alternatives. To address this, we introduce \trustmethod, the \textit{global component} of our framework, designed to build DM confidence through multi-objective sensitivity analysis \citep{rappaport_sensitivity_1967}. This component expands the search beyond the DM's direct feedback to uncover potentially overlooked regions. Grounded in decision sciences \citep{desender_postdecisional_2019, desender_subjective_2018, wu_online_2014}, providing this broader validation of solution quality helps to foster DM confidence.

\trustmethod informs the DM about the sensitivity of each objective $f_l$ to perturbations in another objective $f_{l'}$. This allows the DM to understand how "active" an objective $l$ is with respect to $l'$ (i.e., how much $f_l$ changes if $f_{l'}$ is varied) and whether such modifications could lead to significantly better outcomes in $f_l$. An objective is \textit{active} if perturbing another results in substantial changes to it. Given that function evaluations $f(x)$ are expensive, \trustmethod conservatively focuses on identifying samples that promise improvement without requiring a full re-run of \methodname's local active sampling (Section \ref{sec:active-sampling}). Specifically, we are interested in samples $x$ that improve upon the current best-known value for $f_l$, denoted $f_l(x^*)$. The improvement in dimension $l$ is:
\begin{equation}
    I_l(x) = \max\!\Bigl\{ f_l(x) - f_l\bigl(x^*\bigr), 0 \Bigr\}.
    \label{eq:Ilx}
\end{equation}
This measures how much $f_l(x)$ exceeds $f_l(x^*)$. We then consider the \emph{expected improvement} $\text{EI}_{l}(x) = \mathbb{E}\!\bigl[I_l(x)\bigr]$, calculated over a posterior model for $f_l(x)$ (e.g., a Gaussian Process) \citep{jones_efficient_1998}. $\text{EI}_{l}(x)$ can often be computed in closed form. If $\text{EI}_l(x) > 0$, $x$ is a worthwhile candidate to evaluate and potentially highlight to the DM. Otherwise, if no improvement in $f_l$ is likely, objective $l$ is considered not active with respect to the perturbed dimension $l'$.
To identify such promising candidates for building confidence, \trustmethod solves:
\begin{equation}\label{eq:t-mosh-formulation}
\begin{split}
    \max_{x \in X} \text{EI}(x) \quad \text{s.t.} \quad & u_{\alpha_{\ell}}(x) \geq 0 \quad \forall \ell \in [L] \setminus \{\ell'\}, \\
    & \text{and } u_{\alpha_{\ell',H}-\epsilon}(x) \geq 0
\end{split}
\end{equation}
This formulation targets samples with positive expected improvement that also satisfy the DM's current soft and hard bounds (slightly relaxed for the perturbed hard bound $\alpha_{\ell',H}$, via $\epsilon$) sampled from their posterior distributions. By focusing computational resources on such promising candidates, \trustmethod complements the local search by effectively guiding exploration towards regions that confirm solution robustness or reveal valuable alternatives, aligning with the DM’s broader priorities. Further details are in Appendix~\ref{appendix:t-mosh}.

\section{Overview of Experiments}\label{sec:evaluation-setup}

We evaluate \methodname with three complementary settings: simulation experiments on synthetic benchmarks and real-world applications (Section~\ref{sec:experimental-results}), a human-subject user study (Section~\ref{sec:user-study}), and a high-stakes cervical cancer brachytherapy case study (Section~\ref{sec:case-study}). The simulations enable controlled comparison against baseline feedback mechanisms at scale, the user study captures effects that rule-based simulation cannot, and the case study validates the framework in a realistic safety-critical setting. Before presenting results, we briefly clarify what each of our experiments isolates.

Our main simulation results (Figures~\ref{fig:full-unit-feedback-results} and~\ref{fig:eval-results}) evaluate complete deployed systems, pairing each baseline with information gain (IG) for active query selection, as IG is standard for preference learning and has been shown to outperform alternatives such as volume removal in both convergence and query ease \citep{biyik_asking_2020}. Pairing each baseline with its standard acquisition function reflects practical deployment and avoids hand-tying baselines to a criterion not designed for them.

To isolate the contribution of our active sampling, Figure~\ref{fig:full-ablation-results} ablates it against random sampling with the rest of the framework held fixed. To further isolate the effect of the feedback mechanism itself, Figure~\ref{fig:active-sample-baselines-max-min} equips all baselines with our active query sampling method. As described in Appendix~\ref{appendix:simulation-setup}, our simulator follows fixed rules for bound adjustment, whereas real decision-makers exhibit flexibility, priors, and confidence-driven exploration. Our user study (Section~\ref{sec:user-study}) addresses these simulation limitations directly.

\section{Simulation Experiments}\label{sec:experimental-results}

This section presents a comprehensive empirical evaluation of \methodname against common feedback mechanisms using simulated DM interactions across synthetic benchmarks and real-world applications. Our comparisons include pairwise feedback, complete ranking \citep{plackett_analysis_1975, luce_individual_1959}, partial k-ranking (k=3) \citep{guiver_bayesian_2009}, and random selection \citep{biyik_asking_2020}. For these baselines (excluding random), preference queries were selected via an information gain-based objective \citep{biyik_asking_2020}. Furthermore, we conduct extensive ablation studies to assess the impact of each component of \methodfullname. For clarity, \methodname integrated with \trustmethod is referred to as \methodfullname. 
\textit{Additional experiments and setup details are available in Appendix \ref{appendix:simulation-results}}.

\subsection{Simulation Setup}\label{sec:simulation-setup}

To evaluate our framework, we simulated DM interactions using a ground-truth ideal point $\pmb{y}^*$, derived from a randomly sampled preference vector $\pmb{\lambda}^*$ and soft-hard bounds $\pmb{\alpha}^*$ to ensure $\pmb{y}^*$ resides within a high-utility region. The goal is to identify $\pmb{y}^*$ over 10 feedback iterations, and we measure interaction efficiency using the SHF Utility Ratio (Eq. \ref{shf-utility-ratio}), calculated against these ground-truth values. \textit{Further details on the simulation setup and additional experiments accounting for cognitive complexity of the feedback mechanisms are in Appendix \ref{appendix:simulation-setup}}.

\subsection{Simulation Results}

\textbf{Branin-Currin, Four Bar Truss, DTLZ2.}
We leverage both the Branin-Currin (two objectives) and DTLZ2 (three objectives) problems provided in the
BoTorch framework \citep{balandat2020botorch}. We further evaluate our approach on a MOO structural engineering problem: the Four Bar Truss design problem from REPROBLEM \citep{tanabe_easy--use_2020}. This system presents a bi-objective optimization task with four continuous design variables and exhibits a convex Pareto frontier \citep{cheng_generalized_1999}.

\begin{figure}[ht]
    \centering
    \includegraphics[width=1.0\linewidth]{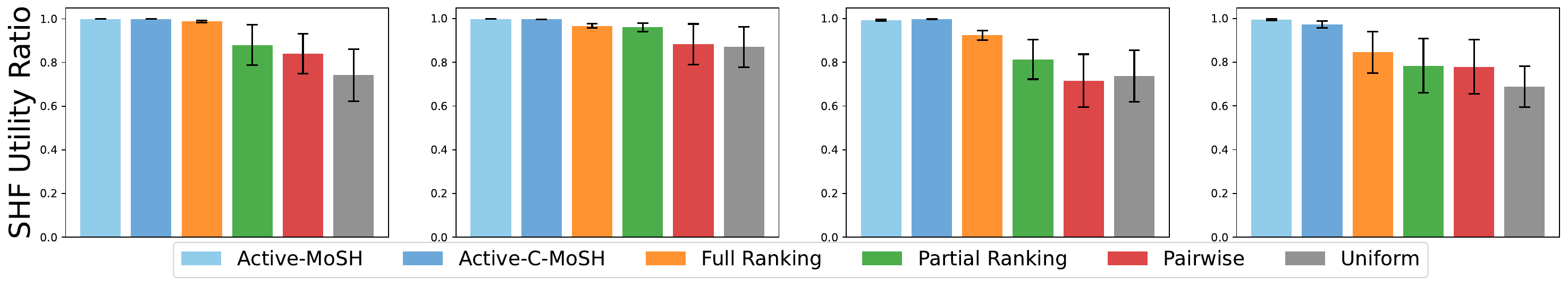}
    \caption{Evaluation results with 10 units for Branin-Currin, Four Bar Truss, Brachytherapy, and DTLZ2 applications (from left to right). 10 independent trials.}
    \label{fig:full-unit-feedback-results}
\end{figure}

\textbf{Brachytherapy.}
We validate our framework using a real clinical case for brachytherapy treatment planning. In this setting, we present it as a two-objective optimization problem: maximizing dose to the tumor while minimizing radiation exposure to the bladder. All results are shown in Figure \ref{fig:full-unit-feedback-results}, illustrating that \methodname and \methodfullname both often result in higher SHF Utility Ratio values compared to all baselines. More details in Appendix \ref{appendix:brachytherapy-additional-details}.

\section{User Study Evaluation}\label{sec:user-study}

\textbf{Setup.}
To evaluate the practical utility of \methodname, we conducted an IRB-approved human-subject user study. We designed a task involving the selection of AI-generated magazine cover photos, which serves as a controlled proxy for complex multi-objective decision-making (see Figure~\ref{fig:user-study-feedback-interface}). Participants were tasked with using different feedback types to balance two competing objectives, realism and color vividness, to match a specified editorial goal (e.g., high realism, balanced, or high vividness). Each participant interacted with multiple task instances, each targeting a different region of the tradeoff space. \textit{Further details in Appendix~\ref{appendix:user-study-details}}.

We compared \methodfullname against pairwise, full ranking, and partial ranking, all using information gain. We emphasize that our evaluation focuses on the general applicability of the interactive preference learning framework; thus, we do not compare against methods tailored specifically to generative AI, such as direct prompting.

We recruited 21 Amazon Mechanical Turk participants. In the primary study, each worker completed one task consisting of four instances, each using a distinct feedback type. For each instance, workers iteratively used the mechanism until they identified a tradeoff point they felt matched the task description, then selected a final image from any prior iteration. To further investigate decision confidence and learning effects, we conducted two supplementary experiments: an ablation study isolating the global decision confidence component and a longitudinal study where participants performed four consecutive trials using only \methodfullname. Post-study, a 5-point Likert survey assessed each mechanism's perceived (1) mental effort, (2) expressiveness, and (3) trustworthiness (confidence).

\subsection{User Study Hypotheses and Metrics}\label{sec:user-study-hypotheses}
\textbf{Hypotheses.} \textbf{H1.} Active-C-MoSH leads to faster convergence to the optimal tradeoff point.	\textbf{H2.} Active-C-MoSH is helpful for building confidence in the DM process. \textbf{H3.} Active-C-MoSH is more cognitively demanding. \textbf{H4.} Active-C-MoSH is more expressive of user’s preferences.

To evaluate \textbf{H1}, we used the SHF Utility Ratio at the second feedback iteration. Due to the subjective nature of \textbf{H2}, \textbf{H3} and \textbf{H4}, we evaluated them with MTurk worker Likert scale responses. To further quantitatively assess \textbf{H2} (confidence), we propose the Iteration Stop Efficiency (ISE) metric, a behavioral proxy for decision-maker confidence. The underlying assumption is that a DM's confidence in the system directly influences their interaction behavior. Backed by literature in decision theory, a DM who is confident that the system is effectively exploring the most relevant tradeoffs or presenting optimal solutions is more likely to halt the process once a satisfactory ("good enough") point is found \citep{clausen_reasoning_2007, nandy_adopting_2025}. Conversely, a DM lacking confidence may continue iterating excessively, fearing that superior alternatives exist just beyond the presented options. ISE operationalizes this concept by measuring how quickly a user finds a satisfactory point relative to the total number of interactions. 
A point $y=f(x)$ is deemed ``good'' if its SHF Utility Ratio $U(\{x\})$ satisfies $U(\{x\}) \geq (1-\delta) U(\{x^*\})$ for a threshold $\delta \in [0, 1)$, where $U(\{x^*\})$ is the utility of the ideal optimal point. ISE is then $1 - (M - M_G) / M$ where $M$ is the total number of feedback iterations, and $M_G$ is the first iteration index ($1 \leq M_G \leq M$) where the set of evaluated points $X_{M_G}$ contains at least one ``good point.'' Formally, $M_G = \min\{M' \in \{1,...,M\} \mid \exists x \in X_{M'} \text{ s.t. } U(\{x\}) \geq (1-\delta) U(\{x^*\})\}$.

\begin{figure}[hbt!]
    \centering
    \includegraphics[width=1.0\linewidth]{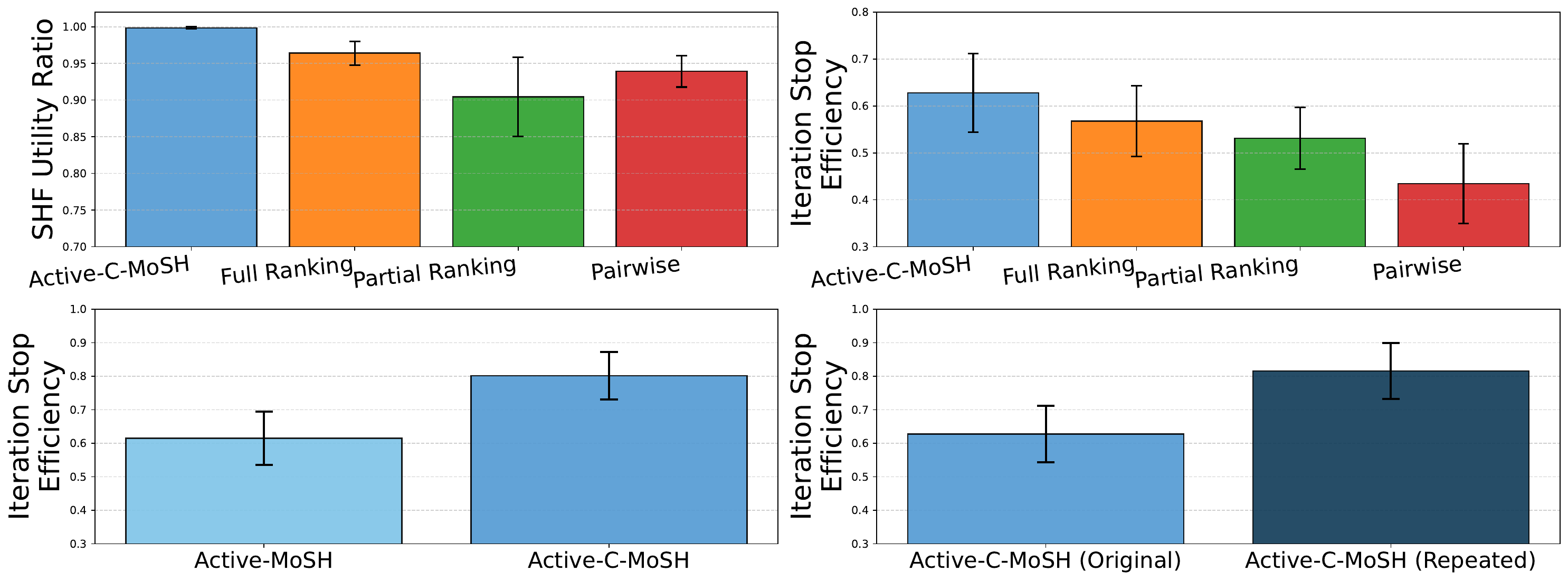}
    \caption{User Study Results. Top (left to right): SHF Utility Ratio at second iteration and Iteration Stop Efficiency Metric (ISE) for all feedback types. Bottom: ISE results for the \trustmethod ablation study and the longitudinal familiarity study.}
    \label{fig:user-study-results}
\end{figure}

\subsection{User Study Results}\label{sec:user-study-results}

We present the results of our user study in Figure~\ref{fig:user-study-results}.
Regarding \textbf{H1} (faster convergence), an analysis of the SHF Utility Ratio at the second feedback iteration \textit{supported that \methodfullname{} enabled participants to reach the optimal tradeoff point more effectively}. Pairwise comparisons revealed that \methodfullname{} achieved significantly higher utility ratios than pairwise feedback ($\text{adj. p} = 0.05$) and full ranking ($\text{adj. p} = 0.05$). The difference compared to partial ranking was not statistically significant ($\text{adj. p} = 0.07$). 

For \textbf{H2} (building confidence), we analyzed ISE, Likert-scale survey responses, and our two supplementary experiments. Our strongest evidence came from the post-study Likert-scale survey. When asked to rate their trust in each method, participants rated \methodfullname significantly higher than full ranking (adj. p = 0.01), partial ranking (adj. p = 0.05), and pairwise feedback (adj. p = 0.04), providing direct support for H2 (results in Appendix~\ref{appendix:user-study-results}). 

While ISE results showed a positive trend in the comparative study, the difference was not initially statistically significant. We attributed this to user unfamiliarity with the novel soft-hard bound mechanism, which was also rated as more complex (see \textbf{H3}). To test this hypothesis, our longitudinal study tracked ISE over four consecutive trials of \methodfullname. By the last trial, once participants had familiarized themselves with the interface, \methodfullname achieved an ISE score (mean=0.82) that was statistically significantly greater than that of all other baseline feedback mechanisms (p < 0.04), confirming that the behavioral signal of confidence strengthens as the learning curve is overcome. Furthermore, to isolate the source of this confidence, our ablation study compared \methodfullname against \methodname. The full \methodfullname yielded significantly higher ISE scores (mean=0.80 vs. 0.62, p=0.04), confirming that the sensitivity analysis provided by C-MoSH is the primary driver for building decision confidence.

Finally, subjective Likert-scale survey responses addressed \textbf{H3} (cognitive demand) and \textbf{H4} (expressiveness). For cognitive load, higher scores indicate more effort. \methodfullname{} (mean=$3.75$) was rated as requiring more mental effort compared to full ranking (mean=$2.25$, $\text{adj. p} =0.00$), partial ranking (mean=$3.00$, $\text{adj. p} = 0.02$), and pairwise feedback (mean=$2.75$, $\text{adj. p}=0.00$). This is not unexpected, as \methodfullname{} is a novel feedback structure and may require more familiarization than conceptually simpler mechanisms like pairwise or ranking. Another part of this may be attributed to the nature of providing exact numerical values using slider bars, which may require more cognitive load than discrete values for rankings or buttons for pairwise selection. For \textbf{H4}, \textit{\methodfullname{} (mean=$4.25$) was rated as significantly more expressive} than full ranking (mean=$2.58$, $\text{adj. p}= 0.00$), partial ranking (mean=$3.58$, $\text{adj. p}= 0.06$), and pairwise feedback (mean=$3.20$, $\text{adj.p}=0.02$).

\section{Case Study: Cervical Cancer}\label{sec:case-study}

We evaluated \methodname within a high-stakes, high-dimensional decision support tool for cervical cancer brachytherapy treatment planning. The setup assumes tradeoffs among six competing objectives: maximizing radiation dose to the tumor while minimizing exposure to five healthy regions \citep{chen_almo_2026}. Using three retrospective patient datasets obtained from a local hospital, a domain expert attempted to identify treatment plans comparable to clinically approved reference plans within a strict 30-minute window. Note that standard manual planning typically requires 30-60 minutes \cite{deufel_pnav_2020, michaud_workflow_2016}.

We quantified performance using a scalar utility score derived by normalizing each objective to $[0, 1]$ (where higher is better), dividing by the corresponding reference plan metric, and averaging across dimensions. \methodname achieved a mean utility score of $1.11 \pm 0.02$ across the three cases, significantly outperforming full ranking ($0.82 \pm 0.02$), partial ranking ($0.80 \pm 0.01$), and pairwise feedback ($0.80 \pm 0.01$).

To illustrate the clinical significance of these results, we focus on two critical objectives for clarity: tumor coverage (maximize) and Bladder dose (minimize). In one patient case with reference values of (95.6\%, 457.8 cGy), the expert using \methodname identified a dominating plan with metrics of (96.0\%, 436.2 cGy). This solution offers improved tumor control while simultaneously reducing toxicity risks to healthy tissue. In contrast, using pairwise feedback, it proved difficult to identify a plan superior to (89.0\%, 556.3 cGy). The bladder dose of 556.3 cGy approaches standard safety limits and could greatly increase the risk of severe complications \citep{romano_high_2018, carrara_comparison_2017}. These findings underscore the necessity of expressive feedback mechanisms in safety-critical domains. Additional details for this case study are provided in Appendix \ref{app:case-study}.

\section{Related Works}\label{sec:related-works}

\textbf{Interactive Feedback Mechanisms.} Research in interactive learning has explored diverse feedback modalities. These include learning from corrections \cite{zhang_learning_2019, losey_including_2018}, ordinal feedback \cite{chu_gaussian_2005, chu_preference_2005, ozaki_multi-objective_2023, giovanelli_interactive_2024, wilde_learning_2022}, choice functions \cite{benavoli_learning_2023}, and critiques \cite{argall_learning_2007}. Within MOO, \cite{ozaki_multi-objective_2023} introduced improvement requests for specific objectives. \cite{racca_interactive_2020} used slider bars for interactive parameter tuning in robotics; however, their focus was on single-objective problems. To the best of our knowledge, \methodname is the first interactive framework for MODM that directly allows iterative refinement of experts' dual-level preferences, specifically their aspirational targets (soft bounds) and non-negotiable limits (hard bounds).

\textbf{Active Sampling} generally aims to select the most informative queries to present to a user, thereby maximizing learning from a limited number of interactions \citep{biyik_asking_2020}. One prominent line of research focuses on maximizing the volume removed from the hypothesis space (volume removal) \citep{sadigh_active_2017, palan_learning_2019, biyik_batch_2018, golovin_adaptive_2017, biyik_green_2019, katz_learning_2019, basu_active_2019}. Another seeks to directly maximize the information gained from each query (e.g., via information gain criteria) \citep{golovin_near-optimal_2010, zheng_efficient_2012, houlsby_bayesian_2011, mackay_information-based_1992, krishnapuram_semi-supervised_2004, lawrence_fast_2002}. In contrast, our active sampling approach is specifically tailored for MODM by explicitly considering the high cost of DM validation for each presented point and aiming for diverse coverage of the DM's evolving region of interest.

\textbf{Building Confidence in AI Systems.} The challenge of building user confidence, and trust, in AI systems is a longstanding area of research. Leveraging perturbations to enhance model interpretability, and consequently trust, has been explored in various contexts, such as influence functions and explainable AI \citep{cook_detection_1977, koh_understanding_2017, ribeiro_why_2016, lundberg_unified_2017}. Various measures of trust have also been proposed. \citet{schmidt_quantifying_2019} measured trust via information transfer rates during ML annotation tasks, while \citet{jiang_trust_2018} defined a trust score based on agreement between a classifier and a modified nearest-neighbor classifier to assess prediction reliability. \citet{irshad_leveraging_2024} equates trust to fidelity in a multi-fidelity setting. To the best of our knowledge, we are the first to introduce and empirically validate a quantifiable notion of trust within an interactive MODM context.

\textit{Additional related works may be found in Appendix \ref{app:related-works}}.

\section{Conclusion}

We introduced \methodname, an interactive multi-objective preference learning framework. Its \textit{local component} uses probabilistic soft-hard bounds and active sampling to efficiently refine the search for optimal tradeoffs. The \textit{global component}, \trustmethod, builds DM confidence via multi-objective sensitivity analysis to quantify robustness and uncover overlooked high-value regions. Evaluations across synthetic benchmarks, a cancer treatment case study, and a user study confirm \methodname accelerates convergence to preferred solutions and boosts confidence over baselines. Key limitations (Appendix~\ref{app:limitations}) include computational scalability and reliance on consistent DM feedback.

\begin{acknowledgements} %

    We thank Elizabeth Kidd, Thomas Niedermayr, and members of the Guestrin lab for helpful discussions throughout this project. This research was supported in part by the Stanford Institute for Human-Centered Artificial Intelligence and the Chan Zuckerberg Biohub. SK acknowledges support by NSF 2046795 and 2205329,
    IES R305C240046, ARPA-H, the MacArthur Foundation,
    Schmidt Sciences, and HAI.
\end{acknowledgements}

\bibliography{uai2026-template,references}

\newpage

\onecolumn

\title{Interactive Multi-Objective Probabilistic Preference \\ Learning with Soft and Hard Bounds \\(Supplementary Material)}
\maketitle

\appendix

\section{Multi-Objective Preference Learning with Soft-Hard Bounds}

\subsection{Soft-Hard Utility Functions}\label{appendix:shfs}

\begin{figure}[hbt!]
    \centering
    \includegraphics[width=0.40\linewidth]{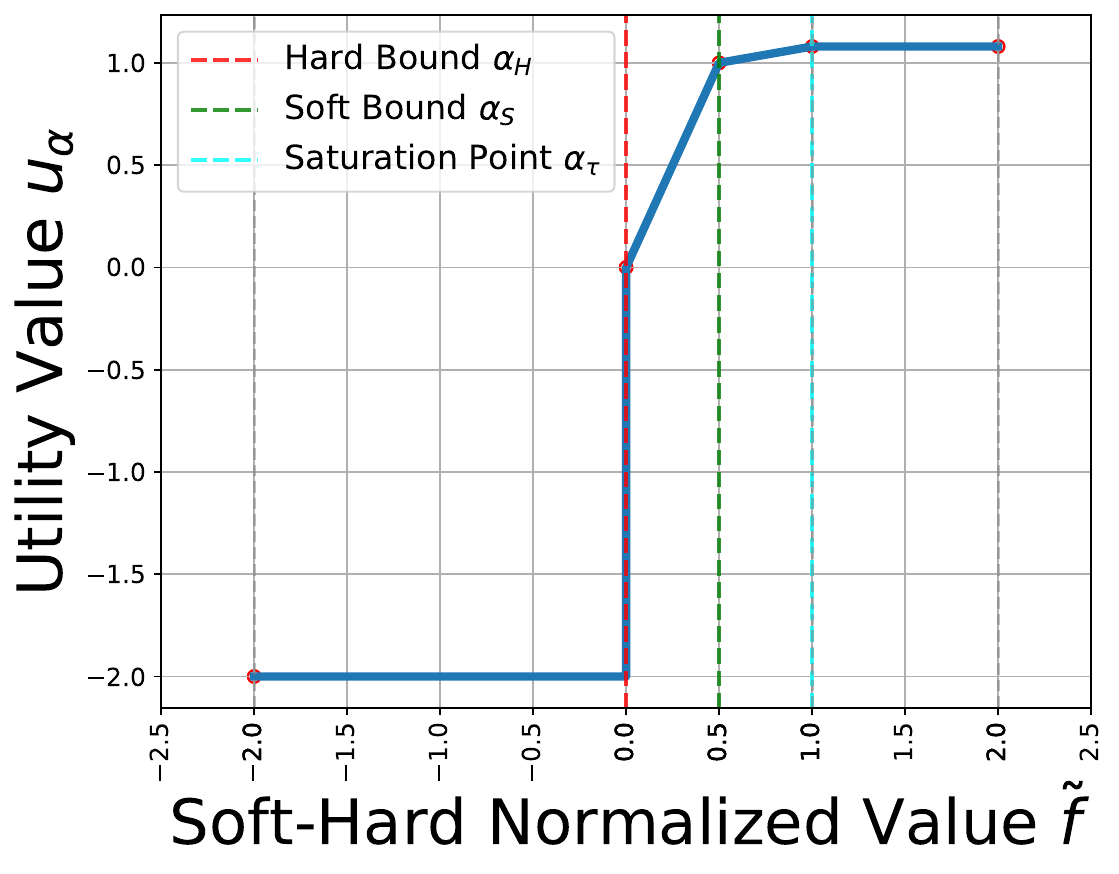}
    \caption{Example of a normalized soft-hard bounded utility function. The dashed vertical bars highlight the regions where the values correspond to the hard, soft, and saturated regions. The utility value associated with points below the hard bound is shown as -2 for illustration purposes only.}
    \label{fig:puf-figure}
\end{figure}

We follow \citet{chen_modeling_2026} and use a similar form for the soft-hard utility functions, as follows:

\begin{equation}\label{eq:shf}
    u_{\alpha}(x) = \begin{cases}
        1 + 2 \beta \times (\tilde{\alpha}_\tau - \tilde{\alpha}_S) & f(x) \geq \alpha_{\tau} \\
        1 + 2 \beta \times (\tilde{f}(x) - \tilde{\alpha}_S) & \alpha_{S} < f(x) < \alpha_{\tau} \\
        \text{1} & f(x) = \alpha_{S} \\
        \text{2 } \times \tilde{f}(x) & \alpha_{H} < f(x) < \alpha_{S} \\
        0 & f(x) = \alpha_{H} \\
        -\infty & f(x) < \alpha_{H} \\
    \end{cases}
\end{equation}

where $\tilde{f}(x)$ and $\tilde{\alpha}$ are the soft-hard bound normalized values, $\alpha_{\tau}$, the saturation point, determines where the utility values begin to saturate, and $\beta$ $\in [0, 1]$ determines the fraction of the original rate of utility, in $[\alpha_{H},\alpha_{S}]$, obtained within $[\alpha_{S}, \alpha_{\tau}]$. Normalization, for value $z$, is performed according to the soft and hard bounds, $\alpha_{S}$ and $\alpha_{H}$, respectively, using: $\tilde{z} = ((z-\alpha_{H}) \division (\alpha_{S}-\alpha_{H})) * 0.5$. In practice, we determine $\alpha_{\tau}$ to be $\alpha_{H}$+$\zeta(\alpha_{S}-\alpha_{H})$, for $\zeta = 3.9$. We use $\zeta = 3.9$, as opposed to $\zeta = 2.0$ (as described in \citet{chen_modeling_2026}) since we found a higher value to be more robust to optimization in interactive settings. Figure \ref{fig:puf-figure} depicts an example of the soft-hard utility function $u_{\pmb{\alpha}(x)}$. For our experiments, we used $\beta = 0.25$. Additional details may be found in \citep{chen_modeling_2026}.

\section{\methodname: Probabilistic Modeling and Sampling}\label{appendix:prob-modeling-sampling}

\subsection{Modeling Soft and Hard Bounds Feedback}\label{appendix:modeling-soft-hard-bounds}

We motivate the specific ranking interpretations for different feedback scenarios (illustrated in Figure~\ref{fig:feedback-interp}) as follows:

\begin{enumerate}
    \item \textbf{Hard Bound Tightened ($\alpha_{\ell,H,m} > \alpha_{\ell,H,m-1}$):}
    When the DM makes a hard bound more restrictive (e.g., requiring a \textit{higher} minimum for tumor coverage), they are effectively shrinking the feasible space. We interpret this as a strong signal that their interest lies in solutions that satisfy this new, more stringent requirement. The most informative points in $Y_m$ are those that now lie near, or just satisfy, this new hard bound. Therefore, points are ranked based on their values in the adjusted dimension: those that meet or exceed the new $\alpha_{\ell,H,m}$ are prioritized, with ranking determined by proximity to $\alpha_{\ell,H,m}$. The critical signal is that the DM wants to ensure this new hard constraint is met, and solutions near this boundary are key to understanding the tradeoffs under this tighter condition.

    \item \textbf{Hard Bound Relaxed ($\alpha_{\ell,H,m} < \alpha_{\ell,H,m-1}$):}
    Conversely, when the DM relaxes a hard bound (e.g., accepting a \textit{lower} minimum tumor coverage), they are expanding the feasible region. This action signals an interest in exploring solutions within this newly accessible part of the tradeoff space, perhaps because previous constraints were too restrictive and prevented desirable solutions from being considered. The critical signal is the DM's willingness to explore what was previously infeasible. Points in $Y_m$ that fall within this newly opened region, or are close to the now-relaxed $\alpha_{\ell,H,m}$, become highly relevant. Ranking therefore prioritizes points that benefit from this relaxation, i.e. those closer to the new (more permissive) $\alpha_{\ell,H,m}$ within the expanded feasible space.

    \item \textbf{Soft Bound Adjusted ($\alpha_{\ell,S,m} \neq \alpha_{\ell,S,m-1}$):}
    An adjustment to a soft bound represents a direct refinement of the DM's aspirational target for an objective. Unlike hard bounds which define feasibility, soft bounds define desirability. Therefore, when a DM moves a soft bound, we take this as an explicit indication that they are now most interested in solutions whose value for that objective is as close as possible to the new $\alpha_{\ell,S,m}$. The critical signal is precisely this new target value. Consequently, points in $Y_m$ are ranked directly by their Euclidean distance to the adjusted $\alpha_{\ell,S,m}$ in the modified dimension—the closer a point is, the higher its rank.

    \item \textbf{No Change in Bounds:}
    If the DM makes no changes to any bounds after reviewing $Y_m$, we interpret this as an implicit confirmation that their current preference model (including the existing bound distributions) adequately captures their region of interest, or at least that $Y_m$ did not provide sufficient new information to warrant an adjustment. In our probabilistic framework (Section~\ref{sec:prob-modeling}), this lack of change translates to an update that places higher certainty (i.e., reduces variance) on the existing parameters of the unmodified bounds' distributions, reinforcing the current belief about the DM's preferences.
    
\end{enumerate}

\textbf{Implementation Details.}
In practice, the ranking of points in $Y_m$ is determined by their Euclidean distance to the bound being modified at the end of iteration $m$, following the intuition described in Section \ref{sec:model-soft-hard-bounds}. The prior distribution over the preferences, $p(\pmb{\lambda})$, is set to a uniform distribution. To compute the posterior distribution over the preferences (Equation \ref{eq:lambda-bayes}), we leveraged the Metropolis-Hastings algorithm with 20 burn-in steps and with the Dirichlet distribution as the proposal distribution \cite{chib_understanding_1995}. To implement the steps described in Section \ref{sec:active-sampling} to obtain the preference queries, we leveraged the BoTorch library v0.10.0 \citep{balandat2020botorch}. \textit{All details are also provided in our provided code.}

\subsection{Active Sampling For Preference Queries}\label{appendix:active-sampling-for-queries}

Additional details about the active sampling method may be found below.

\textbf{Step 1: Dense Set of Points.}

For our experiments, we follow \citet{chen_modeling_2026} and use the Upper Confidence Bound (UCB) heuristic. We define acq($u, \pmb{\lambda_t}, \pmb{\alpha_t}, x$) $= s_{\pmb{\lambda_t}} (u_{\pmb{\alpha_t}(x)})$ where $f(x) = \mu_t(x) + \sqrt{\beta_t} \sigma_t({x})$ (Equation \ref{eq:shf}) and $\beta_t = \sqrt{0.125 \times \log(2\times t + 1)}$. For $\beta_t$, we followed the optimal suggestion in \citet{paria_flexible_2019}. 

\textbf{Step 2: Sparse Set of Points.}

\begin{algorithm}[H]
    \caption{Greedy Submodular Partial Cover (GPC) Algorithm \citep{krause_robust_2008}}
    \begin{algorithmic}[1]
    \Procedure{GPC}{$\Bar{F}_q, q$}
        \State $C = \emptyset$
        \While{$\Bar{F}_q(C) < q$}
            \State \textbf{foreach} $c \in D \backslash C$ \textbf{do} $\delta_c = \Bar{F}_q(C \cup \{c\}) - \Bar{F}_q(C)$
            \State $C = C \cup \{\argmax_c \delta_c\}$
        \EndWhile
    \EndProcedure
    \end{algorithmic}
    \label{algorithm:gpc}
\end{algorithm}

\section{\trustmethod: Enhancing Decision-Maker Confidence with Sensitivity Analysis}\label{appendix:t-mosh}

In practice, we found it difficult to display proper objective points to highlight, i.e. within the soft-hard bounds $\pmb{\alpha}$ and exceeding the current best $f_l(x^*)$ for objective $l$ -- especially at earlier iterations. As a result, we solve for Equation \ref{eq:t-mosh-formulation} multiple times (10) and filter out the obtained objective points which violate the SHF or do not exceed the current best point.

\section{Simulation Results}\label{appendix:simulation-results}

\subsection{Simulation Setup}\label{appendix:simulation-setup}
All simulations used a ground-truth ideal point $\pmb{y}^* = \max_{x \in X} s_{\pmb{\lambda}^*} (u_{\pmb{\alpha}^*}(x))$, derived from a randomly sampled ground-truth preference vector $\pmb{\lambda}^* \in \Delta(L)$, where $\Delta(L)$ is the $L$-dimensional probability simplex, and randomly sampled set of soft and hard bounds $\pmb{\alpha}^*$. To obtain $\pmb{\lambda}^*$ and $\pmb{\alpha}^*$, we go through the following procedure: (1) sample a set of hard bounds $\alpha_{\ell,H}^*$, for each dimension $\ell$, in [0, 0.95], (2) sample a set of soft bounds $\alpha_{\ell,S}^*$, for each dimension $\ell$, in [$\alpha_{\ell,H}^*+0.5$, 1.0], (3) sample $\pmb{\lambda}^*$ using the heuristic $\pmb{\lambda}^* = \textbf{u} \division \lVert \textbf{u} \rVert_1$, where $\textbf{u}_\ell \sim \text{N}(\alpha_{\ell,S}, |\alpha_{\ell,H}-\alpha_{\ell,S}| \division 3)$. This ensures that the simulated $\pmb{\lambda}^*$ and $\pmb{y}^*$ correspond to the high-utility regions of the sampled SHF. Throughout our experiments, for the scalarization function we used the augmented Chebyshev scalarization function $s_{\pmb{\lambda}}(y) = -\max_{\ell\in[L]}\{\pmb{\lambda}_{\ell}\lvert y_{\ell} - z^{*}_{\ell}\rvert\} - \gamma\sum_{\ell=1}^L \lvert y_{\ell} - z^{*}_{\ell}\rvert$ where $z_\ell^*$ is the ideal point for objective $\ell$.

\textbf{Baseline Feedback Simulations.} To simulate pairwise, ranking, and partial ranking feedback, we utilized $\pmb{\lambda}^*$ and $\pmb{\alpha}^*$ with added Gaussian noise $\epsilon \sim N(0, \sigma^2)$, where $\sigma=0.01$, when determining preferences between alternatives. Specifically, for pairwise comparisons, the hidden noisy utilities were computed as $\upsilon_i = s_{\pmb{\lambda}^*}(u_{\pmb{\alpha}^*}(x_i)) + \epsilon_i$, where $u_{\pmb{\alpha}^*}(x_i)$ represents the soft-hard utility values for alternative $i$. The alternative with the higher noisy utility value was selected as preferred. Ranking and partial ranking-based feedback follow similarly. Sample queries were selected using an information gain sampling strategy. Since the active sampling component of \methodname implicitly selects Pareto-optimal points, we also only select samples from the Pareto frontier for the baselines (to ensure fairer comparisons).

\textbf{\methodname Feedback Simulation.} For \methodname feedback, we simulated DM interaction in response to the displayed query points $Y_m$ at iteration $m$. The DM is assumed to respond based on a \textit{reference point} within $Y_m$. For instance, a DM might observe points at the extremes of $Y_m$ to determine if a bound modification is warranted. When simulating \methodfullname (i.e., \methodname with \trustmethod), we model the DM's enhanced confidence from \trustmethod's sensitivity analysis by allowing for larger magnitudes in their bound adjustments. \textit{Note: We assert that it is difficult to accurately simulate the effects of \trustmethod. This is our attempt at doing so. However, even without such simulations of \trustmethod, we observe enhanced performance. We believe \trustmethod is better understood through our user study.}

We assume a default magnitude of adjustment for the soft and hard bounds, 0.3 and 0.45, respectively. \textit{Although we chose those as the default values, we did find our results to be consistent across others of similar magnitude.} For the final value to adjust the bound with, we weighted it by a value sampled from a Gaussian distribution centered at the difference between the reference point and $\pmb{y}^*$, with a fixed variance ($\sigma=0.01$).

\begin{itemize}
    \item When $\pmb{y}^*$ falls outside some of the current bounds $\pmb{\alpha}_{H,m}$, the simulated DM selects the bound associated with the objective $\ell$ that violates the value of $\pmb{y}^*$ the most and adjusts those bounds to bring $\pmb{y}^*$ closer to being within bounds. The adjustment magnitude is determined by the reference point, which we assume to be the point in $Y_m$ closest to $\pmb{y}^*$.
    \item When $\pmb{y}^*$ is within all bounds $\pmb{\alpha}_{H,m}$, the DM selects the bound furthest from $\pmb{y}^*$ and, we assume, attempts to narrow the feasible space around $\pmb{y}^*$. For this case, we assume the reference point to be the point in $Y_m$ furthest from $\pmb{y}^*$.
    \item We assume the DM initially leverages the hard bounds, to ensure that $\pmb{y}^*$ is within the desired region. For some objective $\ell$ that is selected, if the soft and hard bounds are too close in proximity, i.e. $\alpha_{\ell,S,m} - \alpha_{\ell,H,m} < \beta$, we assume the DM then leverages the soft bounds to \textit{fine tune} the desired region points.
\end{itemize}

Gaussian noise is added to the observations of points in $Y_m$ at each iteration $m$. The magnitude of bound adjustments is sampled from a Gaussian distribution centered at the deviation between $y^*$ and the reference objective point, with added noise. When $y^*$ is outside the bounds and our proposed method \trustmethod promotes an improved point, we assume the DM has enhanced confidence and increases the magnitude with which they modify the soft or hard bound for that iteration. We assume this increase to be 5\%.

Each simulator maintains consistent noise parameters and sampling procedures to ensure fair comparisons across feedback types. This simulation framework allows us to systematically evaluate how different types of preference feedback and varying levels of DM noise affect the convergence and effectiveness of our proposed approach. The setup enables controlled experiments while capturing realistic aspects of human decision-making behavior such as inconsistency and imprecision in feedback.

In terms of compute, all simulations were run on an internal cluster with CPUs. The flagship run for each simulation experiment typically finished within a day.

\subsection{Performance Evaluation}\label{appendix:performance-evaluation}

To present a different perspective on the simulated experiments, we attempt to account for cognitive complexity of the feedback mechanisms using \textit{interaction units}. In general, the literature has no general consensus on feedback complexity since it is contextual to the task and visualization \citep{nesbitt_comparison_2024}. We propose a notion of feedback complexity specific to our task and then summarize supporting literature for our mapping and provide analysis from our user study.

\begin{figure}[hbt!]
    \centering
    \includegraphics[width=0.85\linewidth]
    {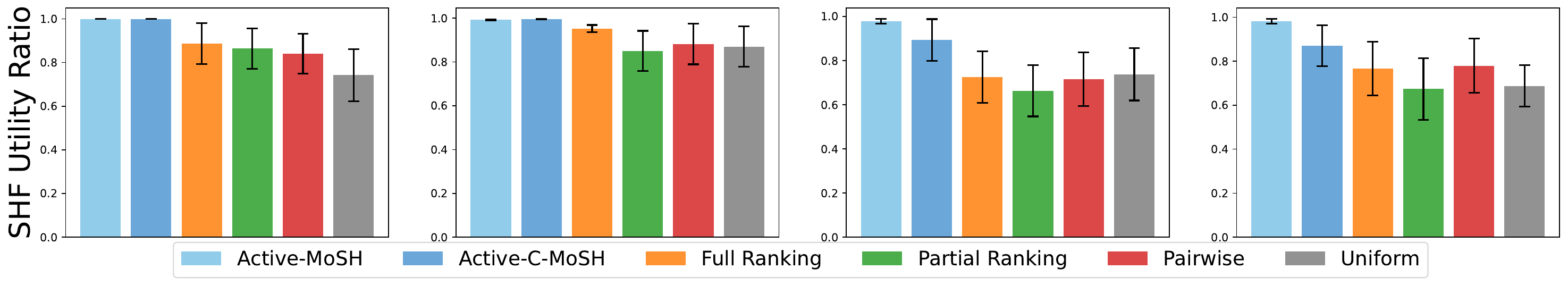}
    \caption{Evaluation results with 10 interaction units for the Branin-Currin synthetic function, Four Bar Truss engineering design, Brachytherapy Treatment Planning, and DTLZ2 (four objectives) applications (from left to right). The cognitive complexity of the feedback mechanisms were taken into consideration, using interaction units. The mean and standard error results were all computed over 10 independent trials.}
    \label{fig:eval-results}
\end{figure}

Pairwise feedback is assigned 1 unit as it is generally treated as the baseline unit for low-complexity feedback \citep{wilde_learning_2022, wirth_survey_2017}. \methodname is 2 units, as it requires the DM to (1) determine the dimension and bound to modify, and (2) specify the modification's magnitude and direction. Slider and scale (not soft-hard bounds) feedback was studied in \citep{wilde_learning_2022}, which decomposes it into two stages (direction identification, then adjustment), aligning with our identification and quantification mapping of 2 units. Ranking-based feedback is assigned $k$ units, where $k$ is the number of points displayed at iteration $m$\footnote{$k \log k$ is also a valid interpretation for ranking-based interaction units; we use $k$ because the cognitive load of ranking is not always uniform across points and for simplicity.}. For ranking, \citet{jamieson_active_2011} notes a baseline $k \log k$ upper bound; the structured nature of our displayed rankings (two objectives, sorted) supports a tighter lower bound of k \citep{jamieson_active_2011}. This method provides a standardized basis for comparing preference elicitation efficiency (Figure~\ref{fig:eval-results}).

We provide further motivation for the specific unit assignments below:

\begin{itemize}
    \item \textbf{Pairwise Feedback (1 unit):} Selecting one preferred item from a pair is considered a relatively simple cognitive task. It involves a single comparative judgment and is thus assigned as our baseline of 1 interaction unit.

    \item \textbf{\methodname{} Feedback (2 units):} Providing feedback in \methodname{} involves a two-step cognitive process by the DM:
    \begin{enumerate}
        \item \textit{Identification:} The DM must first identify which of the $L$ objective dimensions they wish to adjust, and then decide whether to modify the soft or hard bound for that dimension. This requires understanding the current state of the tradeoff space.
        \item \textit{Quantification:} The DM must then specify the magnitude and direction of the change for the selected bound (e.g., how much to increase the soft bound or decrease the hard bound). This involves a quantitative judgment about their desired preference shift.
    \end{enumerate}
    Given these distinct decision-making steps, we assign 2 interaction units to a single \methodname{} feedback action.

    \item \textbf{Ranking-based Feedback ($k$ units):} When a DM is asked to rank $k$ presented items, the cognitive effort generally scales with $k$. While the exact cognitive complexity can be debated (e.g., it might not be strictly linear, and the difficulty of ranking can depend on the similarity of items), assigning $k$ units reflects that the DM performs $k-1$ implicit pairwise comparisons to establish a full order, or at least makes $k$ evaluations to place each item. As noted in the footnote, $k \log k$ is another plausible model reflecting sorting complexity. However, we opt for $k$ units due to (1) the cognitive load of ranking not always being uniform across all $k$ items (e.g., identifying the best and worst might be easier than ordering middle items), and (2) for simplicity in our comparative framework. This assignment acknowledges that processing and ordering $k$ items is substantially more demanding than a single pairwise choice.
\end{itemize}

For mechanisms like ranking that are assigned multiple interaction units ($k>1$) to represent a single complete feedback instance (e.g., one full ranking of $k$ items), we assume that the performance metric (e.g., SHF Utility Ratio as per Equation~\ref{shf-utility-ratio}) remains constant across these constituent interaction units. The metric is only updated once the DM has completed the entire feedback action (e.g., submitted the full ranking). This accounting method, with results presented in Figure~\ref{fig:eval-results}, provides a standardized basis for comparing the efficiency of different preference elicitation approaches, considering both the quality of learned preferences and the human effort involved. Even within such a setting, Figure \ref{fig:eval-results} shows that \methodname and \methodfullname consistently outperform other feedback mechanisms in terms of the SHF Utility Ratio.

\textbf{Empirical Validation of Interaction Units.} To assess whether our theoretical interaction unit assignments reflect actual decision-maker effort, we measured the per-action feedback times recorded during our user study (Section~\ref{sec:user-study}). Table~\ref{tab:feedback-times} reports the mean and standard deviation of the time taken to complete a single feedback action for each mechanism.

\begin{table}[hbt!]
    \centering
    \caption{Mean feedback time per action (in seconds) across feedback mechanisms, measured during the user study. Values are reported as mean $\pm$ standard deviation.}
    \label{tab:feedback-times}
    \begin{tabular}{lc}
        \toprule
        Method & Time (s) \\
        \midrule
        Pairwise & $2.8 \pm 1.4$ \\
        Soft-Hard & $9.7 \pm 4.7$ \\
        Partial Ranking & $16.7 \pm 14.0$ \\
        Full Ranking & $11.5 \pm 12.3$ \\
        \bottomrule
    \end{tabular}
\end{table}

Treating pairwise feedback as the baseline of 1 unit, the empirical time ratios suggest approximate effective costs of 3 units for soft-hard bounds, 4 units for full ranking, and 5 units for partial ranking. The soft-hard mapping is broadly consistent with our theoretical assignment of 2 units, with the modest increase plausibly attributable to the cognitive overhead of specifying exact numerical values via slider bars (as also discussed for H3 in Section~\ref{sec:user-study}). The largest deviation from our theoretical mapping occurs for partial ranking, likely due to the added complexity of first selecting which subset of points to rank before ordering them, an additional decision step not present in full ranking.

\textbf{Re-running Simulations with Empirical Units.} To verify that our conclusions are robust to this empirically grounded accounting, we re-ran our simulation experiments using the empirically derived interaction units in place of the theoretical assignments. Table~\ref{tab:empirical-units} reports the resulting SHF Utility Ratio values for the Four Bar Truss (FBT) and Brachytherapy (Brachy) applications.

\begin{table}[hbt!]
    \centering
    \caption{SHF Utility Ratio under empirically derived interaction units, for the Four Bar Truss (FBT) and Brachytherapy applications. Values are reported as mean $\pm$ standard error over 10 independent trials.}
    \label{tab:empirical-units}
    \begin{tabular}{lcc}
        \toprule
        Method & FBT & Brachytherapy \\
        \midrule
        \methodname & $.97 \pm .03$ & $.97 \pm .03$ \\
        \methodfullname & $\mathbf{.99 \pm .01}$ & $\mathbf{.99 \pm .01}$ \\
        Partial Ranking & $.96 \pm .01$ & $.79 \pm .17$ \\
        Full Ranking & $.92 \pm .06$ & $.80 \pm .18$ \\
        Pairwise & $.97 \pm .03$ & $.77 \pm .10$ \\
        \bottomrule
    \end{tabular}
\end{table}

Under this empirically grounded unit mapping, \methodfullname continues to achieve the highest SHF Utility Ratio across both applications, consistent with our original results in Figure~\ref{fig:eval-results}. This indicates that the observed performance advantages of \methodname and \methodfullname are not an artifact of our specific theoretical interaction unit assignments, and persist when feedback costs are calibrated against measured decision-maker effort.

\subsection{Branin-Currin, Four Bar Truss, and DTLZ2}\label{appendix:synthetic-problems-details}

Branin-Currin and DTLZ2 are standard benchmark problems provided in the BoTorch library \citep{balandat2020botorch}. Four Bar Truss represents a realistic multi-objective optimization problem within the domain of engineering design. Its objectives involve the simultaneous minimization of structural volume and joint displacement, while the decision variables parameterize the lengths of the individual truss members. All the above benchmark problems are open-source.

\subsection{Cervical Cancer Brachytherapy Treatment Planning}\label{appendix:brachytherapy-additional-details}

For the cervical cancer brachytherapy treatment planning application, we leveraged a real and anonymized clinical case which had been performed in the clinic previously. As a result, we had access to the patient CT data (in the form of DICOM files) and the final treatment plan which had been administered to the patient. For computational consistency, we reformulate the minimization objectives for the OARs as maximization problems through appropriate transformations. The decision variables serve as inputs to an epsilon-constraint optimization program \citep{deufel_pnav_2020}, which enables systematic exploration of the treatment planning trade-off space. The three parameters to that linear program, which implicitly control the weights of the different objectives (radiation dosage to the cancer tumor and the bladder), were employed as the decision variables. The Pareto frontier for this problem is non-convex. Due to patient privacy issues, we do not plan to publicly release this clinical case.

\subsection{Additional Experiments and Ablation Studies}\label{appendix:additional-ablations}

To provide a different perspective of the results among the different methods, we use the \textit{area under the SHF Utility Ratio - Feedback curve} as the metric for our ablations below. The x-axis is the number of feedback units and the y-axis is the change in the SHF Utility Ratio, which produces a curve. By taking the area under that curve, we can compare how quickly each variation is able to obtain a high SHF Utility Ratio.

\begin{figure}[hbt!]
    \centering
    \includegraphics[width=0.9\linewidth]{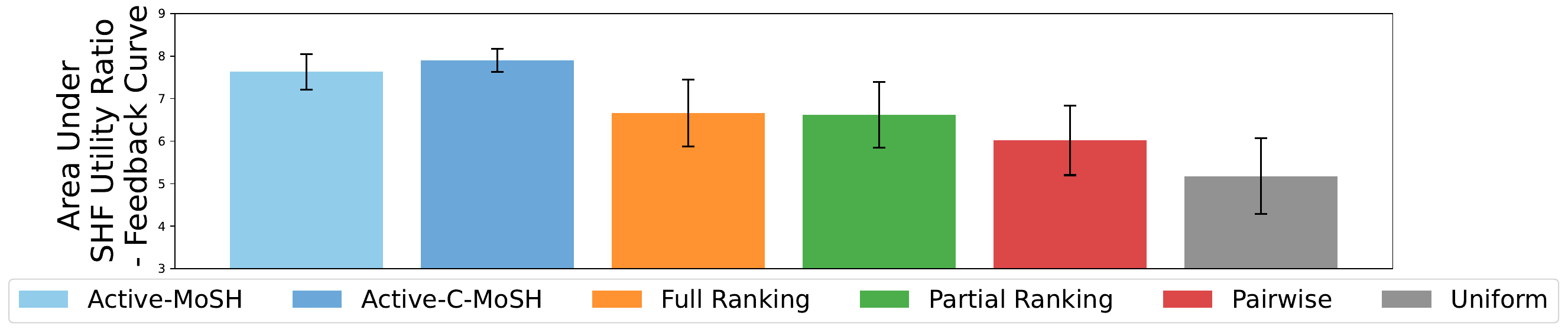}
    \caption{Area under the SHF Utility Ratio-Feedback Curve results over 10 interaction units for the Branin-Currin synthetic function. The mean and standard error results were all computed over 10 independent trials.}
    \label{fig:auc-bc-results}
\end{figure}

\begin{figure}[hbt!]
    \centering
    \includegraphics[width=0.9\linewidth]{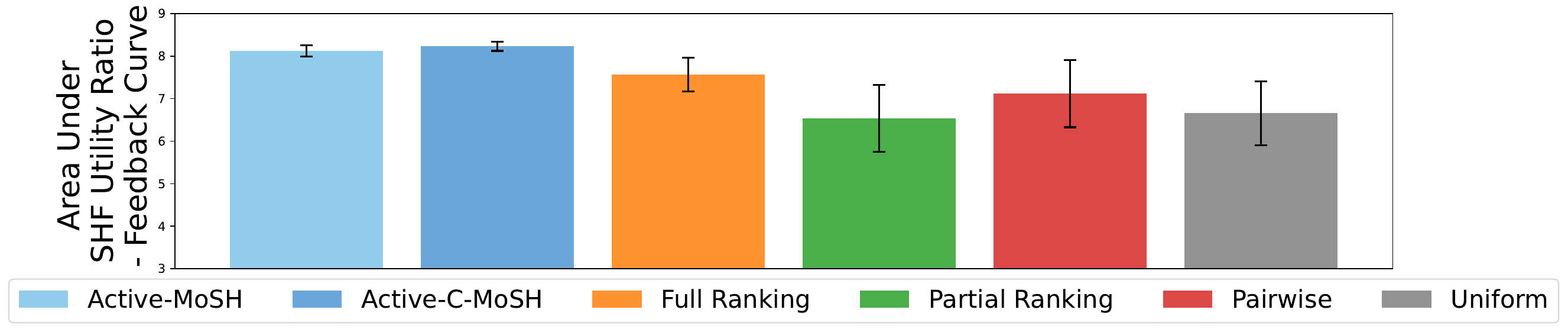}
    \caption{Area under the SHF Utility Ratio-Feedback Curve results over 10 interaction units for the Four Bar Truss function. The mean and standard error results were all computed over 10 independent trials.}
    \label{fig:auc-fbt-results}
\end{figure}

\textbf{Area Under the SHF Utility Ratio-Feedback Curve Results for Branin-Currin and Four Bar Truss.} Figures \ref{fig:auc-bc-results} and \ref{fig:auc-fbt-results} illustrates our results for the area under the SHF Utility Ratio-Feedback Curve with both the Branin-Currin and Four Bar Truss problem. This provides a different perspective on our results as it illustrates the convergence rates with the different methods, further supporting the efficiency of \methodfullname and \methodname.

\begin{figure}[ht]
    \centering
    \includegraphics[width=1.0\linewidth]{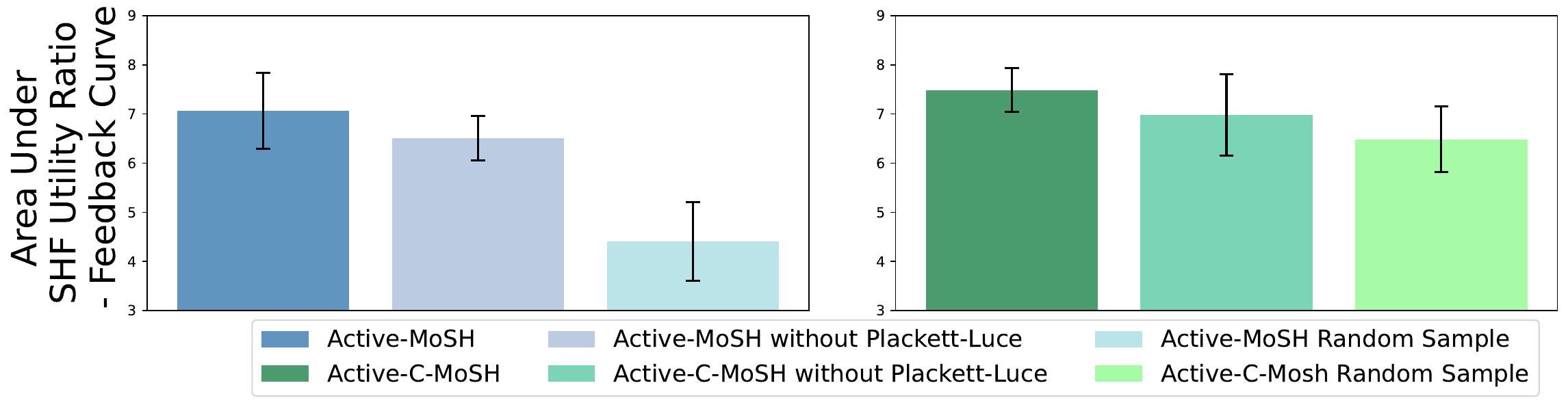}
    \caption{Ablation Results. (\textbf{Left}) Ablation Results for \methodname. (\textbf{Right}) Ablation Results for \methodfullname. \textit{Without Plackett-Luce} use a uniform distribution for p($\pmb{\lambda}$). \textit{Random Sample} obtains subsequent queries with random sampling, as opposed to our active query sampling approach. 10 independent trials.}
    \label{fig:full-ablation-results}
\end{figure}

\textbf{Preference Modeling Ablation.} Figure \ref{fig:full-ablation-results} illustrates our results when ablating both \methodname and \methodfullname with and without the Plackett-Luce likelihood model (which we refer to as \textit{\methodname (\methodfullname) without Plackett-Luce}). For \methodname (\methodfullname) without Plackett-Luce, we instead use a uniform distribution for p($\pmb{\lambda}$). From the results, we notice that solely using the uniform distribution greatly degrades the average performance, while often also increasing the variance of the results. This supports the importance of our preferences modeling component.

\textbf{Active Query Sampling Method Ablation.} Figure \ref{fig:full-ablation-results} illustrates our results when ablating both \methodname and \methodfullname when just using random sampling for the query sampling method, rather than our proposed two-step active querying method. From the results, we notice that simply using random sampling to obtain the queries hurts the converge of \methodname and \methodfullname significantly, while also often leading to higher variance. This supports the importance of our active sampling approach.

\begin{figure}[hbt!]
    \centering
    \includegraphics[width=0.9\linewidth]{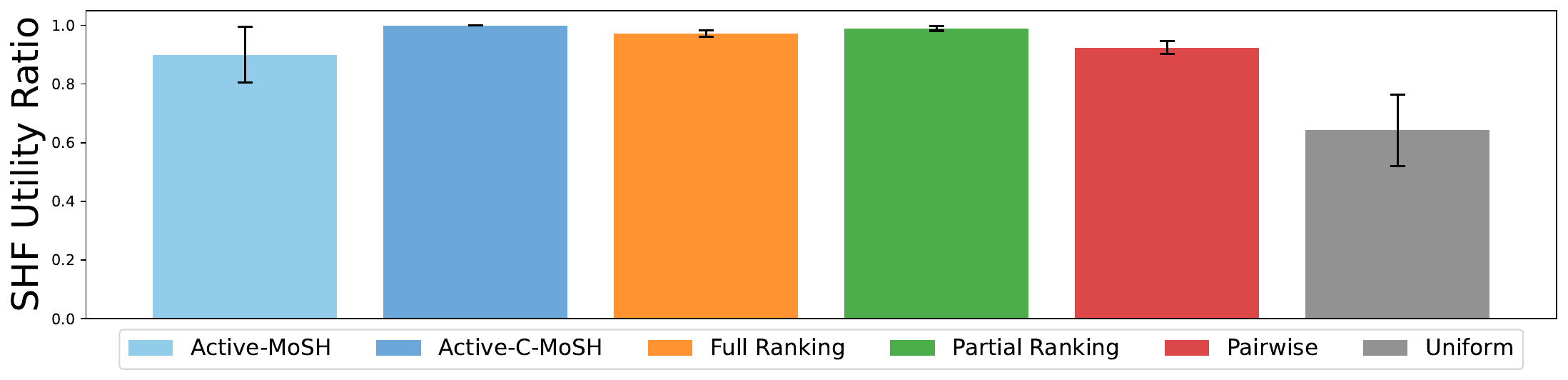}
    \caption{Evaluation results at 10 interaction units for the Branin-Currin synthetic function, where the baseline feedback mechanisms (besides \textit{uniform}) are all using our proposed active query sampling method. The mean and standard error results were all computed over 10 independent trials.}
    \label{fig:active-sample-baselines-max-min}
\end{figure}

\textbf{Observing All With Same Active Querying Method.} We also include additional experiments comparing \methodname and \methodfullname against the baseline feedback mechanisms, all using our proposed active query sampling method, as opposed to the information gain sampling method. In doing so, we seek to understand the impact of the active query sampling method on the SHF Utility Ratio values. We show the results in Figure \ref{fig:active-sample-baselines-max-min}. According to the results, the baseline feedback mechanisms observe higher SHF Utility Ratio values by the end of 10 feedback units (as compared to the results shown in Figure \ref{fig:eval-results}). This supports the improvement which our active sampling method provides for our problem setting. It also further highlights the superiority of \methodfullname over the other baseline feedback mechanisms, while having a lower variance. However, we would like to emphasize that this result is obtained from our simulation setup, which may have limitations related to the simulated feedback. For instance, while we use a set of rules for \methodname, we would expect for a decision-maker to be more flexible in their feedback adjustments, potentially obtaining greater utility over the other feedback mechanisms. This may also be seen in our user study results (Section \ref{sec:user-study-results}).

\section{User Study}\label{appendix:user-study-details}

\subsection{User Study Setup}\label{appendix:user-study-setup-details}

\begin{figure}[hbt!]
    \centering
    \includegraphics[width=0.75\linewidth]{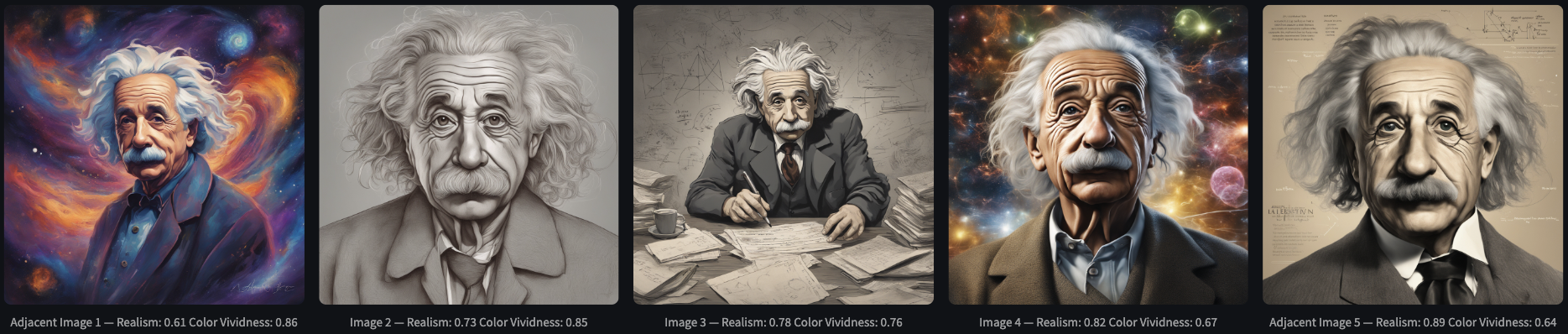}
    \caption{Sample query images from the user study for the task related to Albert Einstein.}
    \label{fig:user-study-feedback-interface}
\end{figure}

As mentioned, we designed an educational magazine cover photo selection task using AI-generated images. Interested participants were first given detailed instructions and were required to pass a five-question screening quiz ($\geq$3 correct answers) to qualify. Successful participants were then given a magazine topic description and instructed to use various feedback mechanisms to select the optimal cover photo. Each task instance's description implicitly defined a tradeoff between two objectives: realism and color vividness, where higher values for both were desirable, aligning with our multi-objective setting. We created six task instances in total, combining two topics (Albert Einstein, Barack Obama) with three distinct tradeoff points. Participants were informed about the specific feedback mechanism to use for each task.

\textbf{Generating the Pareto Frontier of AI Images for User Study.} For the user study (Section~\ref{sec:user-study}), we required a set of AI-generated images that systematically trade off between two competing objectives: realism and color vividness. To achieve this, we controlled the output of a text-to-image generative model by manipulating its effective input prompt representation using proxy tuning \citep{liu_tuning_2024, mitchell_emulator_2023, shi_decoding-time_2024}.

Proxy tuning steers a large pre-trained language model ($M$) by incorporating signals from smaller, specialized models: an expert model ($M^+$) tuned for a desired attribute, and an anti-expert model ($M^-$), typically the untuned version of $M^+$. Following the notation of \citet{liu_tuning_2024}, the output distribution of the proxy-tuned model ($\tilde{M}$) at decoding step $t$, conditioned on the preceding token sequence $x_{<t}$, is given by:

\begin{equation*}
\begin{split}
    p_{\tilde{M}}(X_t | x_{<t}) = \text{softmax}\biggl[ & s_{M}(X_t | x_{<t}) \\
    & + \sum_{i=1}^2 \theta_i \Bigl(s_{M^+_i}(X_t | x_{<t}) \\
    & - s_{M^-_i}(X_t | x_{<t})\Bigr)\biggr]
\end{split}
\end{equation*}
where $s_M$, $s_{M_i^+}$, and $s_{M_i^-}$ are the logit scores from the respective models. The parameters $\theta_i$ are controllable weights applied to the logit difference for expert $M^+_i$. In our two-objective setting, $M^+_1$ was tuned for realism and $M^+_2$ for color vividness. By varying $\theta_1$ and $\theta_2$ at decoding time, we effectively controlled the text prompt representation fed into the text-to-image model, thereby generating images with varying tradeoffs between realism and color vividness.

For our experiments, we utilized models from the T\"{U}LU-2 suite \citep{ivison_camels_2023}: T\"{U}LU-2-13B as the base model $M$, and T\"{U}LU-2-DPO-7B for both the expert ($M^+$) and anti-expert ($M^-$) models. The expert models, $M^+_1$ (realism) and $M^+_2$ (color vividness), were independently fine-tuned using Direct Preference Optimization (DPO) \citep{rafailov_direct_2024} on custom preference datasets. To construct these datasets, we first obtained names of historical figures from the Pantheon 1.0 dataset \citep{yu_pantheon_2016}. For each figure and each objective (realism, color vividness), we used GPT-4o-mini to expand a seed prompt (the figure's name) into multiple textual variations representing positive, neutral, and negative degrees of the target attribute (see Figures~\ref{fig:llm-degree-prompt-realism-pos-1}--\ref{fig:llm-degree-prompt-color-negative} for exact prompts). We generated two positive variations for each. Binary preference pairs for DPO were then formed as (positive variation 1, positive variation 2), (positive variation 1, neutral), (positive variation 1, negative), and (neutral, negative). For pairs not involving two positive variations, the more positive prompt was designated as chosen. For (positive 1, positive 2) pairs, we generated images from both prompts using Stable Diffusion XL \citep{podell_sdxl_2023} and then used GPT-4o-mini with a detailed grading rubric (Figures~\ref{fig:llm-image-eval-prompt-part-1}--\ref{fig:llm-image-eval-prompt-part-4}) to evaluate the resulting images on the relevant attribute (realism or color vividness); the prompt corresponding to the higher-scoring image was marked as preferred.

With the two fine-tuned expert models ($M^+_1, M^+_2$), we systematically varied $\theta_1$ and $\theta_2$ over the range $[-0.5, 1.0]$ with a step size $\delta = 0.05$. For each $(\theta_1, \theta_2)$ pair, we obtained a proxy-tuned prompt which was then used as input to Stable Diffusion XL to generate an image. The realism and color vividness scores (0-100 scale) for each generated image were subsequently evaluated by GPT-4o-mini using the aforementioned detailed grading rubric. This process yielded a diverse set of images for Albert Einstein and Barack Obama, populating different regions of the realism-vividness tradeoff space.

The fine-tuning of expert models and image generation were conducted on an internal cluster using three NVIDIA A100 GPUs (82GB VRAM). Flash Attention 2 \citep{dao_flashattention-2_2023} was employed during DPO fine-tuning, enabling completion of each expert model's flagship run within approximately 12 hours.

\begin{figure}[htbp]
    \centering
    \includegraphics[width=0.45\textwidth]{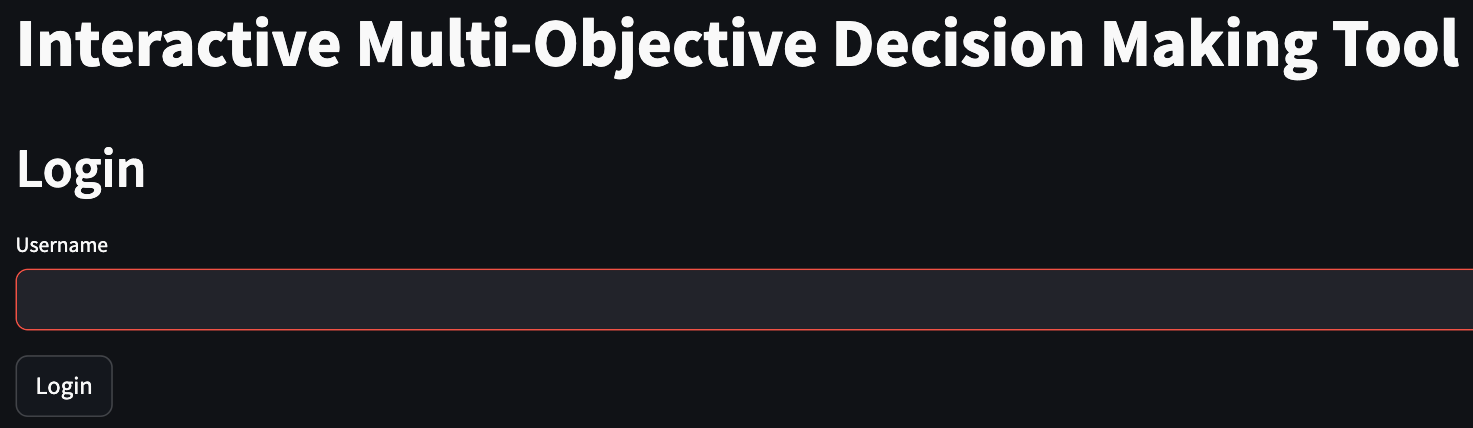} %
    \caption{Login interface for the Interactive Multi-Objective Decision Making Tool used in the user study. Participants entered their MTurk username to begin.}
    \label{fig:user_study_login}
\end{figure}

\begin{figure}[htbp]
    \centering
    \includegraphics[width=0.45\textwidth]{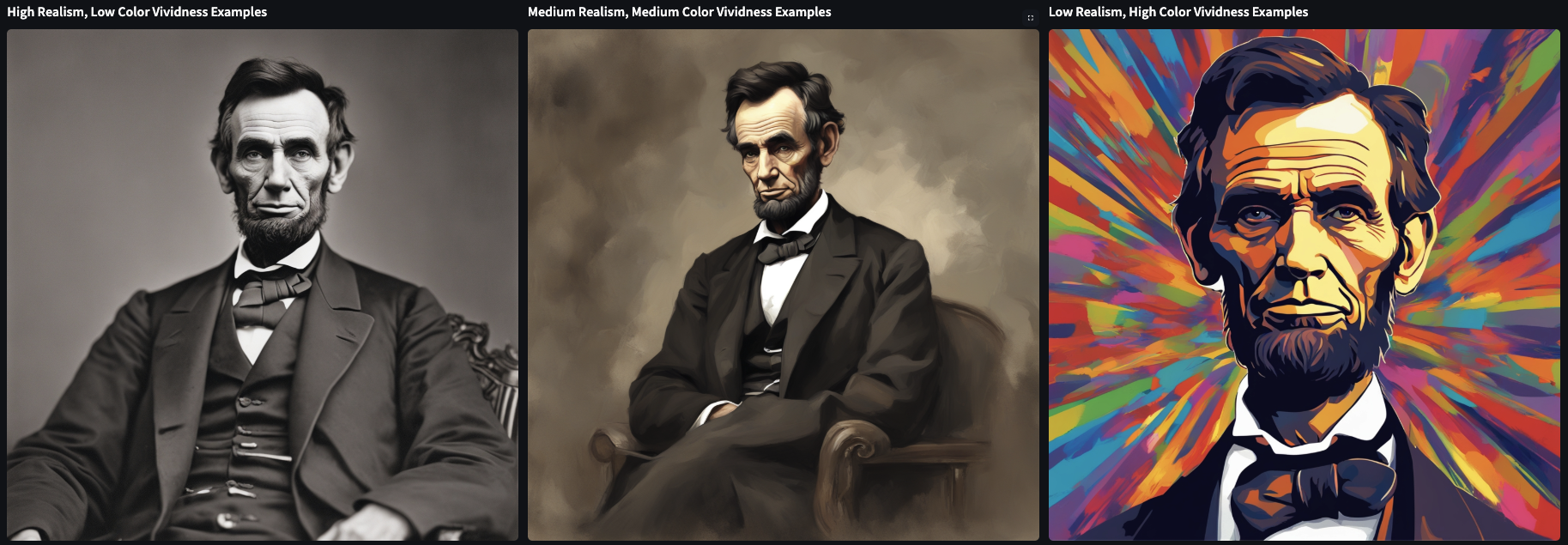} %
    \caption{Instructional material presented to user study participants, illustrating the tradeoff between ``Realism'' and ``Color Vividness'' using AI-generated images of Abraham Lincoln. Examples show (L-R): High Realism/Low Color Vividness, Medium/Medium, and Low Realism/High Color Vividness.}
    \label{fig:user_study_instructions_lincoln}
\end{figure}

\begin{figure}[htbp]
    \centering
    \includegraphics[width=0.45\textwidth]{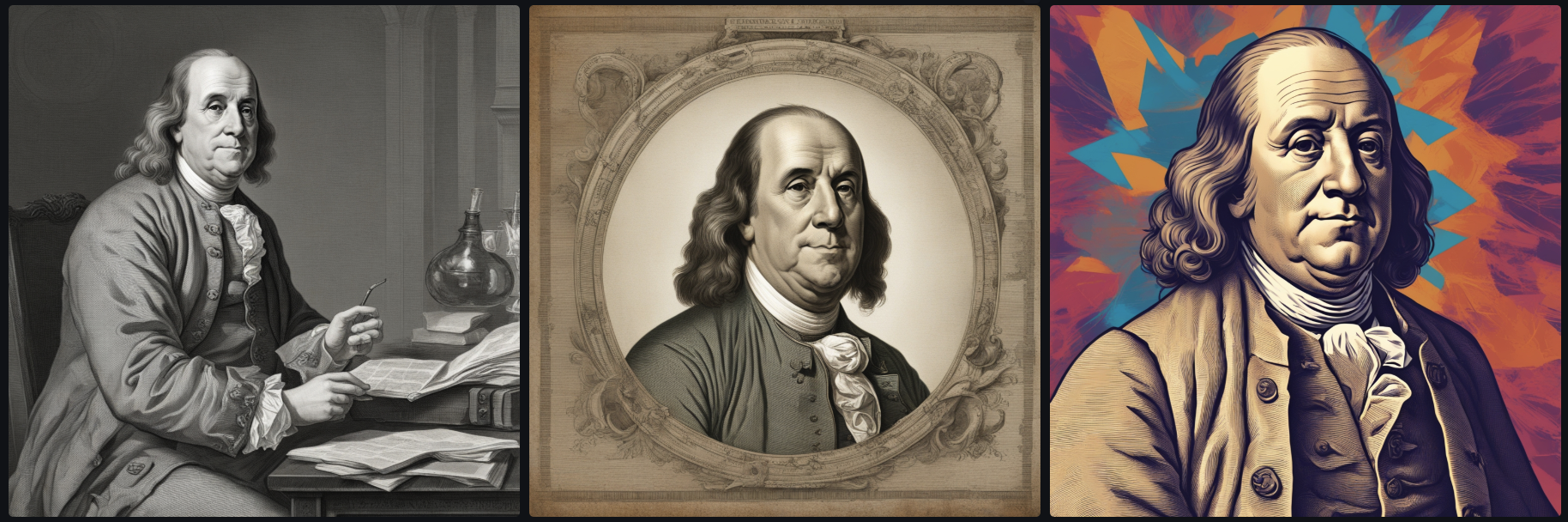} %
    \caption{Additional instructional examples illustrating the realism and color vividness tradeoff, using AI-generated images of Benjamin Franklin, provided to user study participants.}
    \label{fig:user_study_instructions_franklin}
\end{figure}

\begin{figure*}[htbp]
    \centering
    \includegraphics[width=\textwidth]{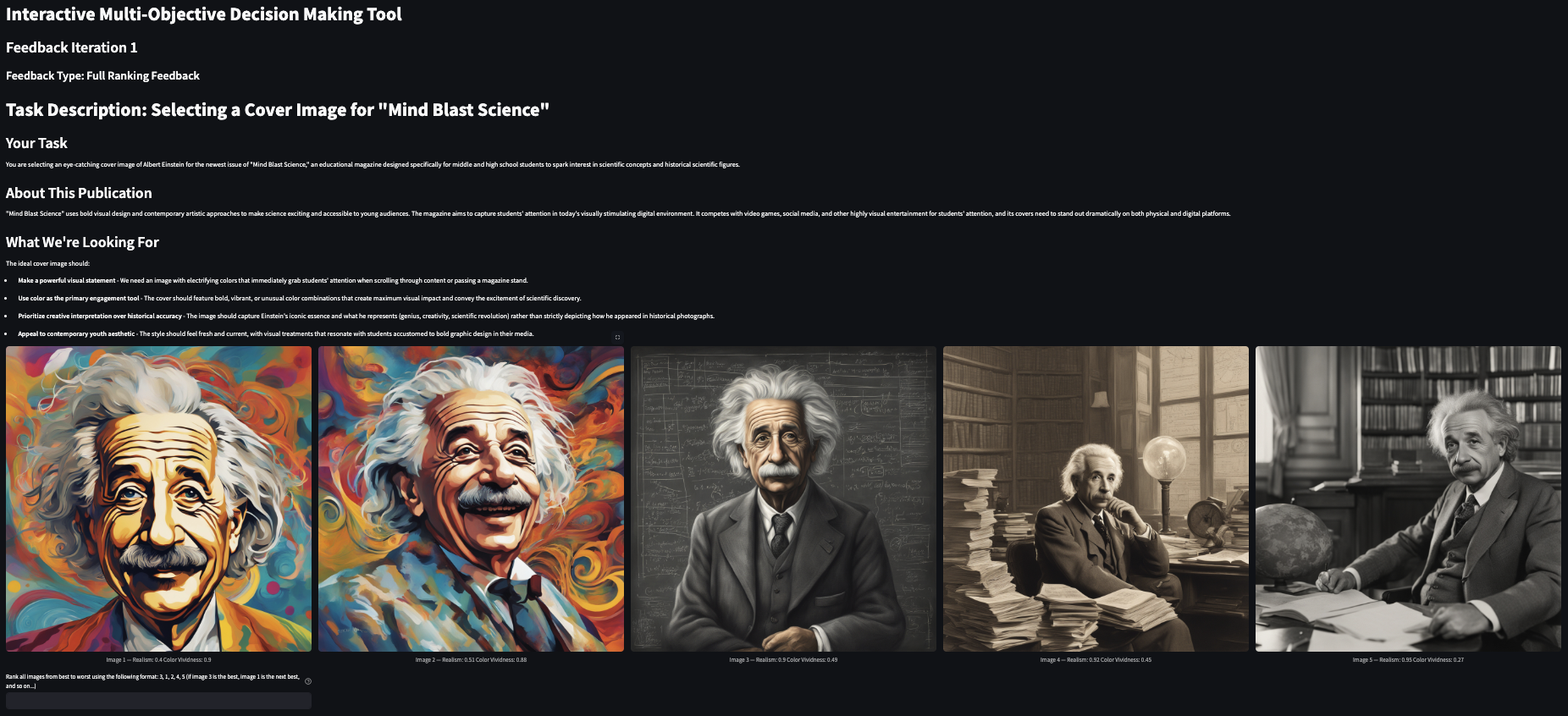} %
    \caption{User interface for the Full Ranking feedback mechanism. Participants are presented with a task description (e.g., ``Selecting a Cover Image for 'Mind Blast Science'''), details about the publication, desired image qualities, and a set of AI-generated images (Albert Einstein in this example) to rank from most to least preferred.}
    \label{fig:user_study_full_ranking}
\end{figure*}

\begin{figure*}[htbp]
    \centering
    \includegraphics[width=\textwidth]{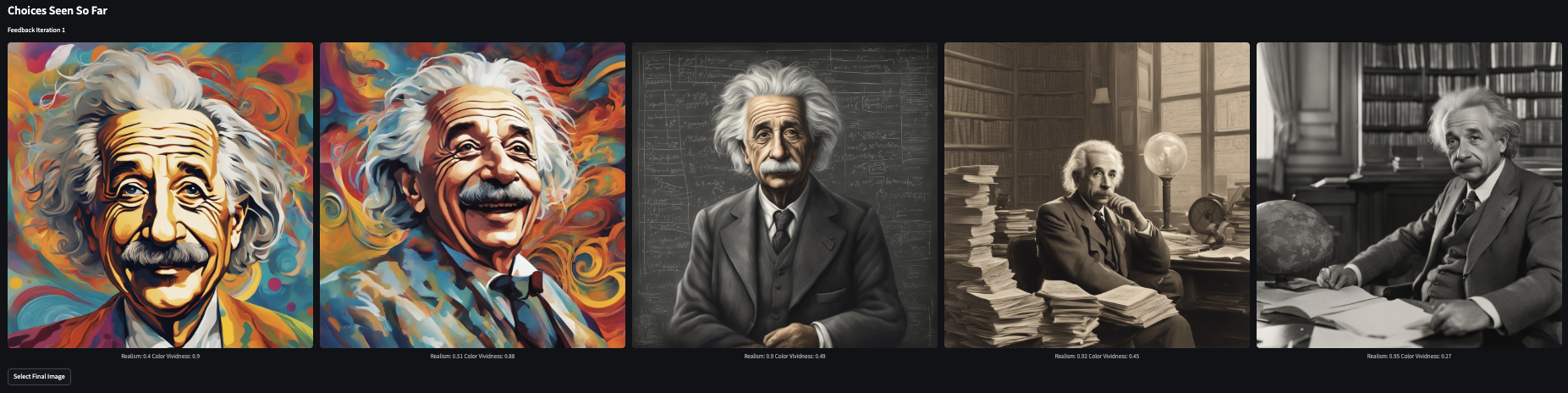} %
    \caption{Interface displaying ``Choices Seen So Far'' during a task instance. This screen allows participants to review all images encountered up to that point before making a final selection or continuing with more feedback iterations.}
    \label{fig:user_study_choices_seen}
\end{figure*}

\begin{figure*}[htbp]
    \centering
    \includegraphics[width=\textwidth]{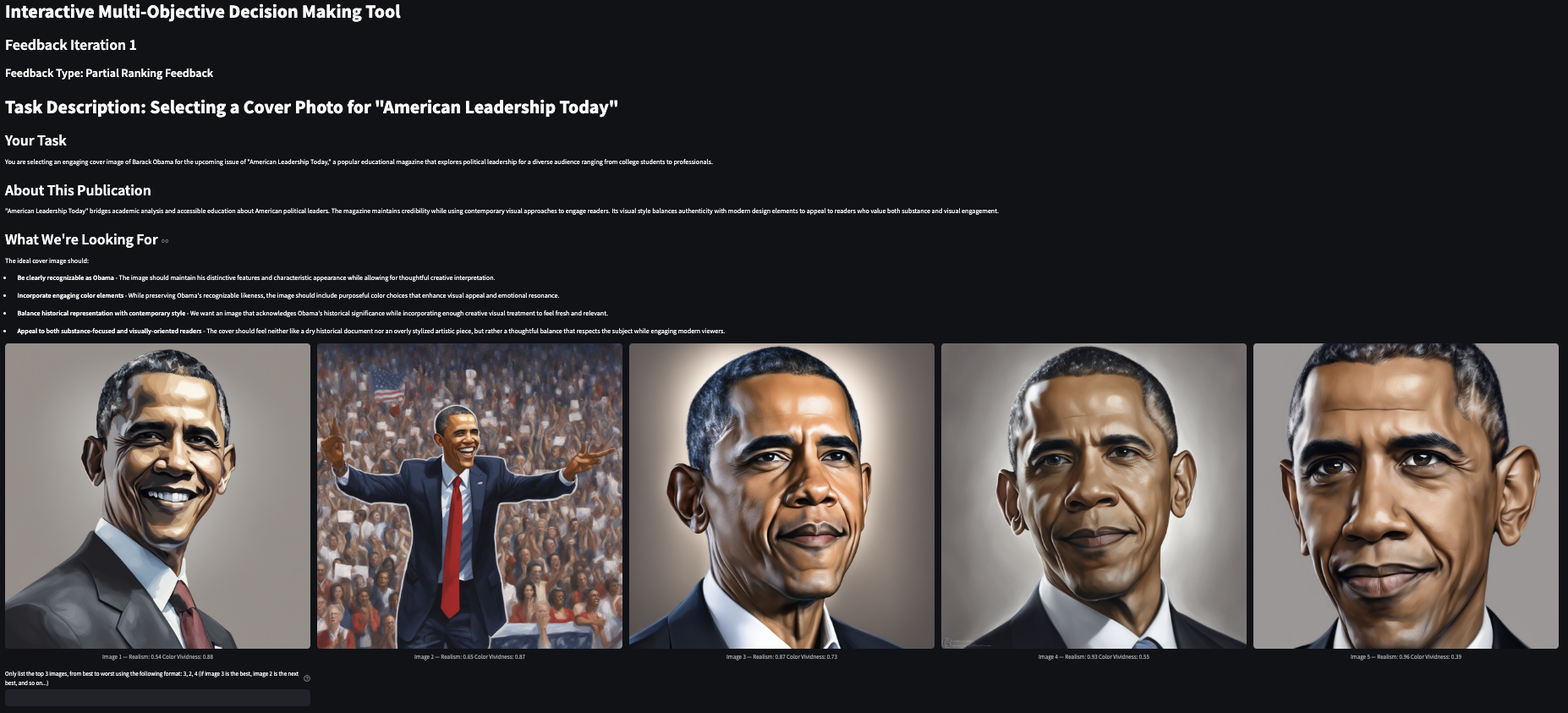} %
    \caption{User interface for the Partial Ranking feedback mechanism. Participants are tasked with selecting a cover photo (Barack Obama for ``American Leadership Today'' magazine in this instance) by ranking only their top three preferred images from the presented set.}
    \label{fig:user_study_partial_ranking}
\end{figure*}

\begin{figure*}[htbp]
    \centering
    \includegraphics[width=\textwidth]{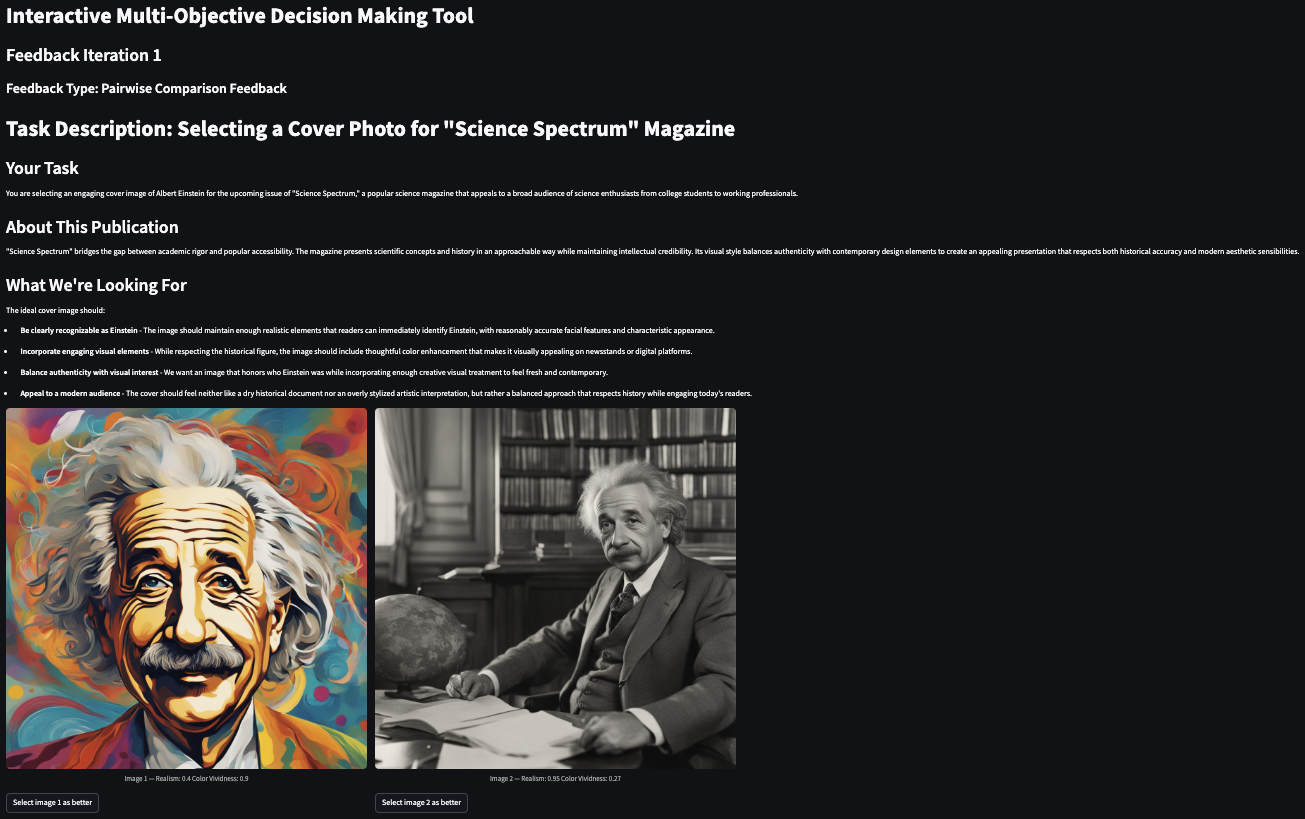} %
    \caption{User interface for the Pairwise Comparison feedback mechanism. Participants are shown two AI-generated images (Albert Einstein for ``Science Spectrum'' magazine here) and must select the one they prefer more based on the task description.}
    \label{fig:user_study_pairwise_comparison}
\end{figure*}

\begin{figure*}[htbp]
    \centering
    \includegraphics[width=\textwidth]{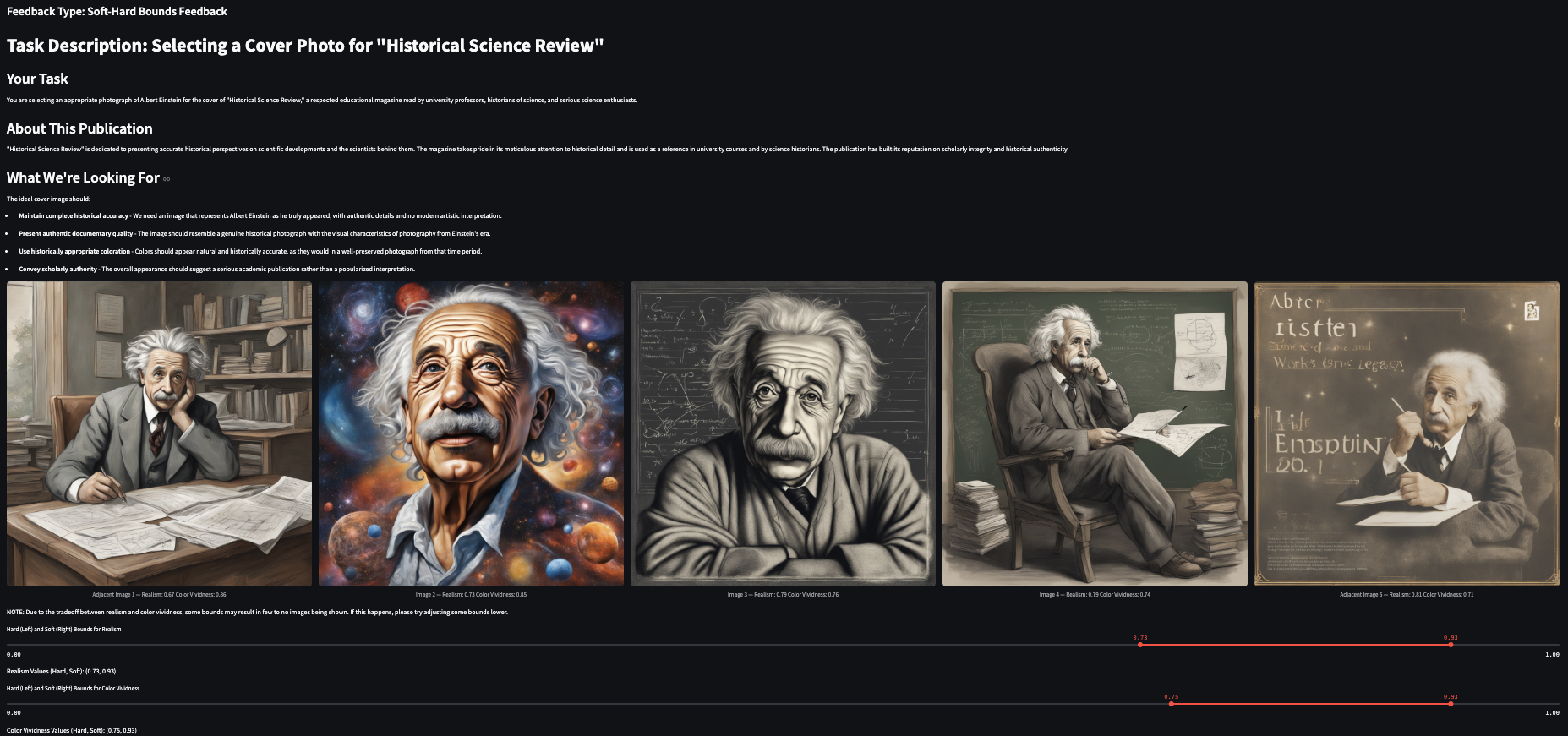} %
    \caption{User interface for the Soft-Hard Bounds feedback mechanism (\methodfullname). Participants adjust sliders to set soft (preferred) and hard (minimum acceptable) bounds for image attributes (realism and color vividness). Images shown are for Barack Obama for ``Future Leaders: Youth Politics Magazine''.}
    \label{fig:user_study_soft_hard_bounds}
\end{figure*}

\textbf{User Study Components and Screenshots.} Here we outline the set of components we used for the user study:

\begin{enumerate}
    \item \textbf{Instructions.} We provided each MTurk participant with a detailed set of instructions covering the tradeoff between the objectives, the different feedback mechanisms and how to use them, a walkthrough of the entire workflow, and other details. We provide the entire set of instructions in Figures \ref{fig:mturk_instructions_part1}-\ref{fig:mturk_instructions_part6}.
    \item \textbf{Screening quiz.} We gave interested MTurk participants a five-question screening quiz on the concepts which were explained in the set of instructions, mainly for ensuring understanding of the feedback types. We provide our screening quiz questions in Figures \ref{fig:mturk_screening_quiz_part1} and \ref{fig:mturk_screening_quiz_part2}. The screening quiz interface was developed using Streamlit \citet{streamlit} and the data was stored using Supabase \citep{supabase}.
    \item \textbf{User study tool.} Each Human Intelligence Task (HIT) consisted of four different task instances. For each task instance, the MTurk participants were provided with the task description, along with an interface to provide their feedback on a set of query images at each iteration. The MTurk participants were instructed to provide feedback on the set of query images until they felt they had found the optimal image for the educational magazine's cover photo in the task description. At the end of the study, each MTurk participant was asked to respond to three different sets of questions regarding the mental effort, expressiveness, and trustworthiness of each of the feedback mechanisms. We provide screenshots of the relevant components in Figures \ref{fig:user_study_login}-\ref{fig:user_study_soft_hard_bounds}. Figures \ref{fig:task_einstein_hr_lc}-\ref{fig:task_obama_lr_hc} depict all of the exact task descriptions we used. The survey questions at the end are shown in Figure \ref{fig:mturk_survey_questions}. The user study tool interface was developed using Streamlit \citet{streamlit} and the data was stored using Supabase \citep{supabase}.
\end{enumerate}

\subsection{User Study Procedure}\label{appendix:user-study-procedure}

We randomized the order of these task instances for each MTurk worker to prevent any bias. Each MTurk worker was permitted to complete the user study twice. For their second time with the user study, the MTurk worker was presented with a random ordering of the last two unseen task instances and two of the earlier task instances, for a total of the four feedback mechanisms.

Each MTurk worker was compensated with \$4.00 for completing the screening quiz and was awarded an additional \$1.50 for passing it. For those who passed the screening quiz, they were compensated with \$19.00 for completing each HIT of the user study. We attempted to ensure that the time in between each step of the user study (e.g. screening quiz and user study HIT) was as short as possible, to retain familiarity; most MTurk participants completed both within 24 hours.

\subsection{User Study Results}\label{appendix:user-study-results}

\begin{figure}[hbt!]
    \centering
    \includegraphics[width=0.9\linewidth]{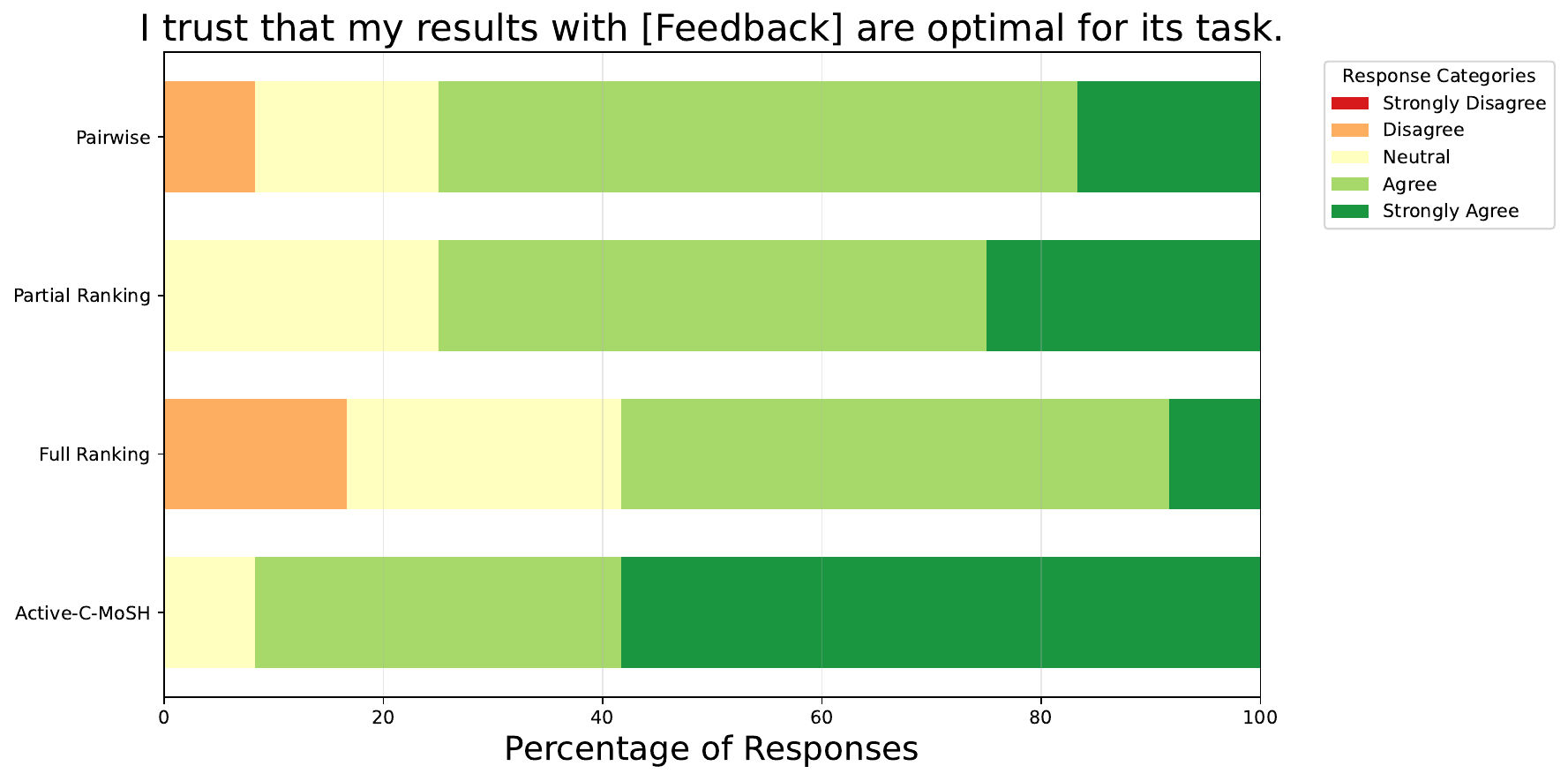}
    \caption{Distribution of Likert-scale survey responses for our third end-of-study question for the MTurk participants.}
    \label{fig:user-study-q3-results}
\end{figure}

Since each of the MTurk participants were given the opportunity to participate in the user study for a total of two times, we used the results from only their second trial for Figures \ref{fig:user-study-results} and \ref{fig:user-study-q3-results}, to account for the novelty effect \citep{elston_novelty_2021}. For hypothesis testing, we leveraged the one-sided Welch's T-Test for \textbf{H1} and \textbf{H2} and the Mann-Whitney U Test for the Likert-scale responses part of \textbf{H2}, \textbf{H3}, and \textbf{H4}. We used a statistical significance level of 0.05 and leveraged the Holm-Bonferroni method to control the Family-Wise Error Rate \citep{holm_simple_1979}.

\begin{figure}[hbt!]
    \centering
    \includegraphics[width=0.9\linewidth]{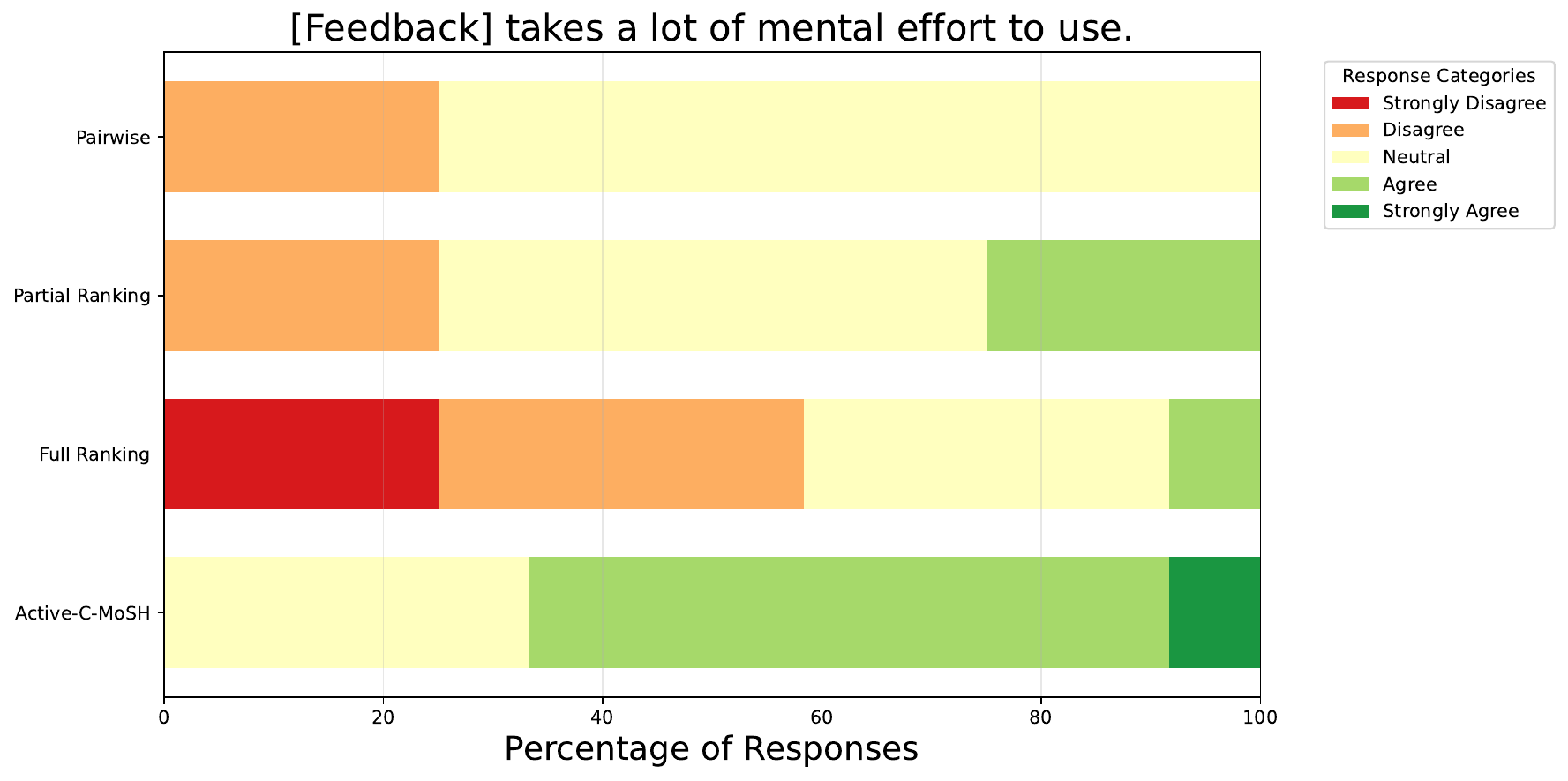}
    \caption{Distribution of Likert-scale survey responses for our first end-of-study question for the MTurk participants.}
    \label{fig:user-study-q1-results}
\end{figure}

\begin{figure}[hbt!]
    \centering
    \includegraphics[width=0.9\linewidth]{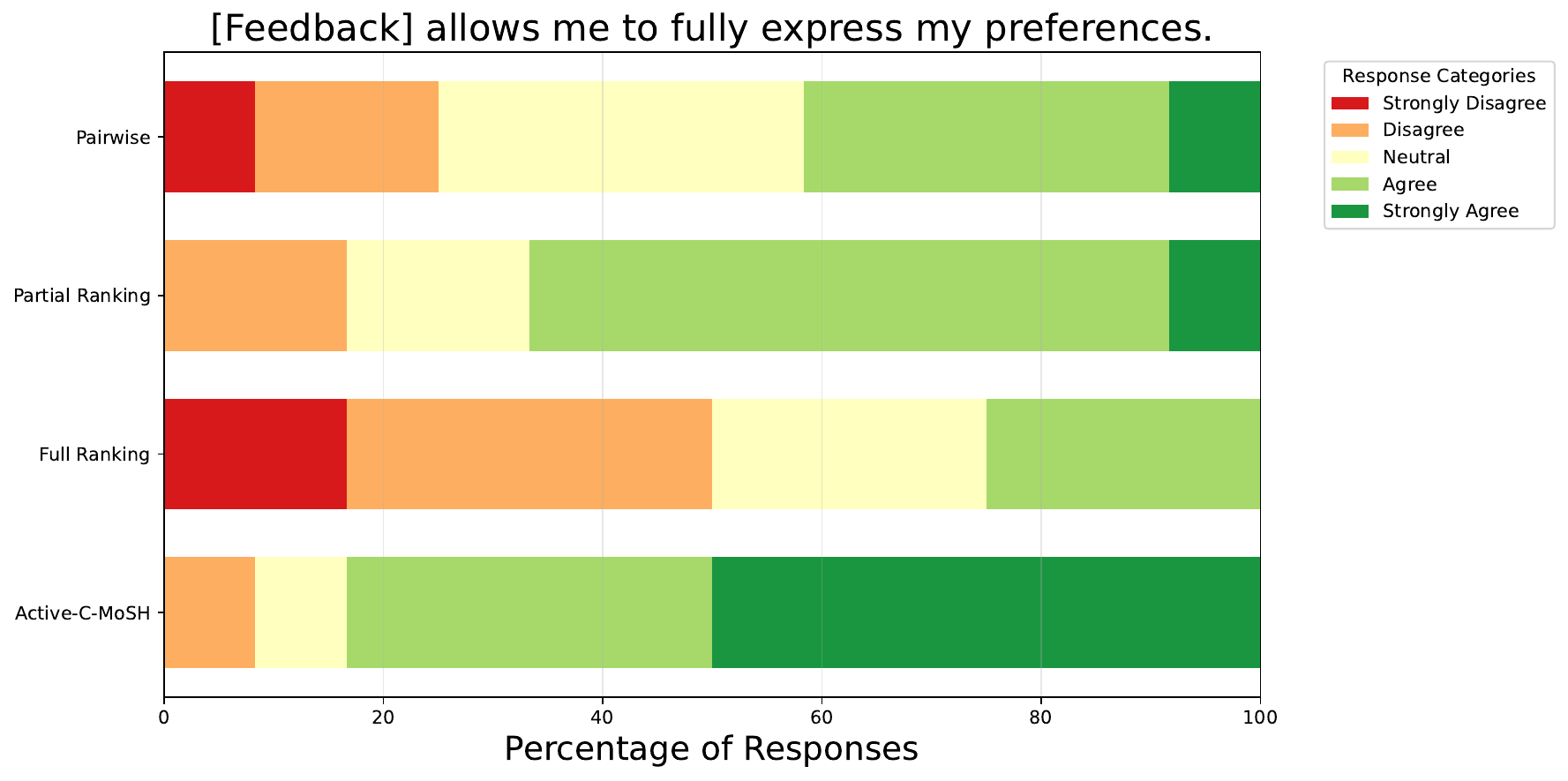}
    \caption{Distribution of Likert-scale survey responses for our second end-of-study question for the MTurk participants.}
    \label{fig:user-study-q2-results}
\end{figure}

Although our statistical hypothesis testing on the Iteration Stop Efficiency (ISE) metric results from \ref{sec:user-study-results} were not statistically significant, we did notice there was a trend towards supporting our \textbf{H2} regarding \methodfullname building more confidence. To further evaluate \textbf{H2}, we also leveraged statistical hypothesis testing on our third end-of-study question asking MTurk participants about how they view each of the feedback types, in terms of trust. In this case, to simplify the question we equate trust with decision confidence. Higher scores indicate more decision confidence. \methodfullname (mean=$4.50$) was rated as ensuring more confidence compared to full ranking (mean=$3.50$, $\text{adj. p} = 0.01$), partial ranking (mean=$4.00$, $\text{adj. p} = 0.05$), and pairwise feedback (mean=$3.83$, $\text{adj. p} = 0.04$). We believe part of this may be attributed to the samples from \trustmethod, which were often presented to the MTurk participant. Another possible reasoning could be the additional familiarity the MTurk participants gained with \methodfullname, better understanding how to interpret it and, therefore, be confident in its results. The distribution of all Likert-scale responses are shown in Figures \ref{fig:user-study-q1-results}, \ref{fig:user-study-q2-results}, \ref{fig:user-study-q3-results}.

\section{Case Study: Cervical Cancer}\label{app:case-study}

\begin{figure}[hbt!]
    \centering
    \includegraphics[width=0.9\linewidth]{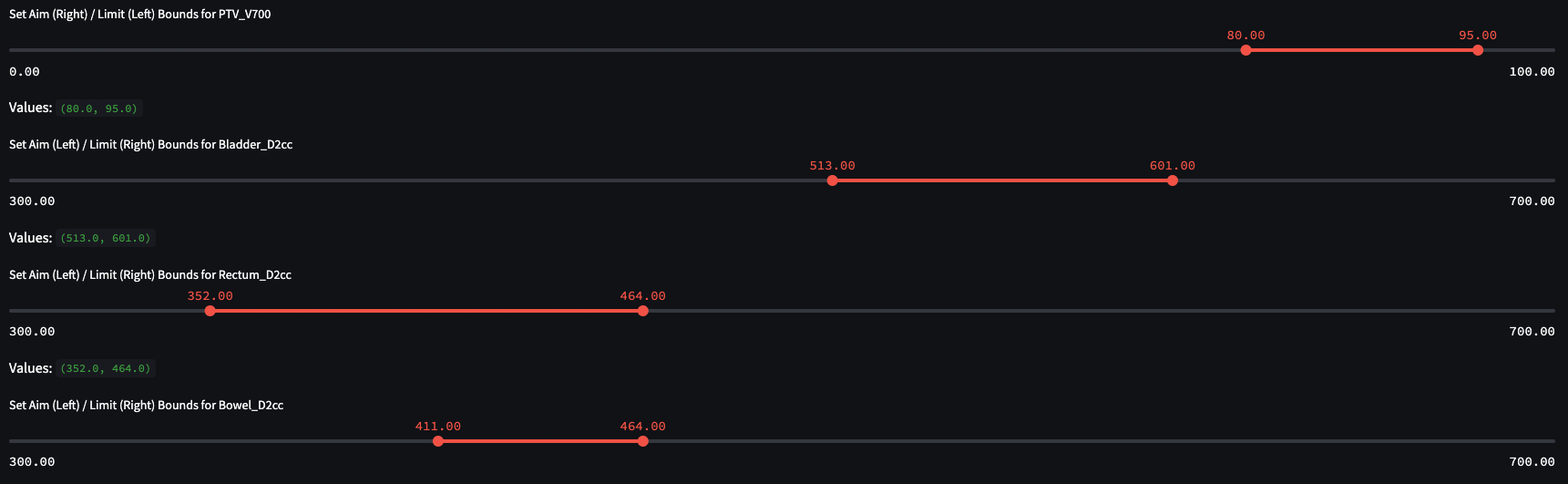}
    \caption{Soft-Hard Bounds Feedback Mechanism for Cervical Cancer Brachytherapy Treatment Planning Case Study.}
    \label{fig:case-study-soft-hard-bounds}
\end{figure}

\begin{figure}[hbt!]
    \centering
    \includegraphics[width=0.9\linewidth]{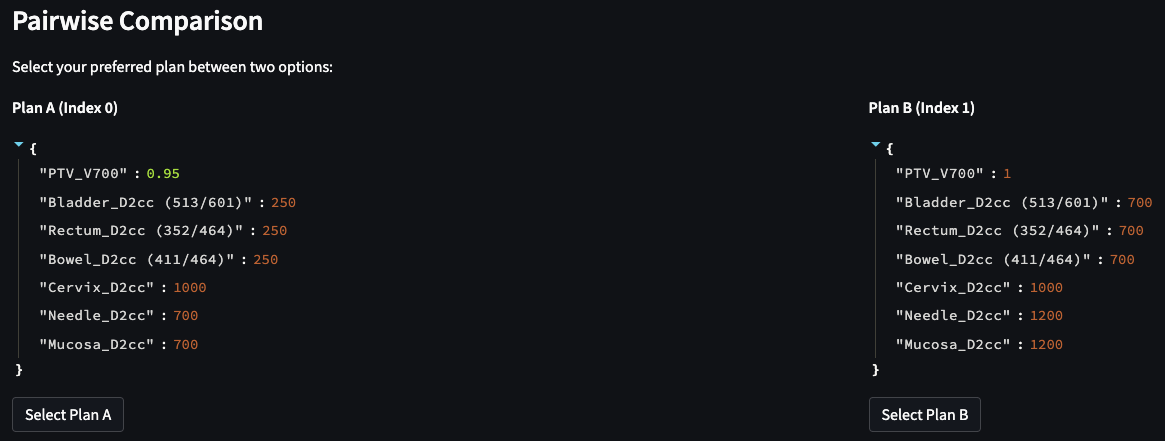}
    \caption{Pairwise Feedback Mechanism for Cervical Cancer Brachytherapy Treatment Planning Case Study. Only default treatment plan metrics are shown. The $\text{Needle}_{\text{D2cc}}$ objective is not used for evaluation.}
    \label{fig:case-study-pairwise}
\end{figure}

\begin{figure}[hbt!]
    \centering
    \includegraphics[width=0.9\linewidth]{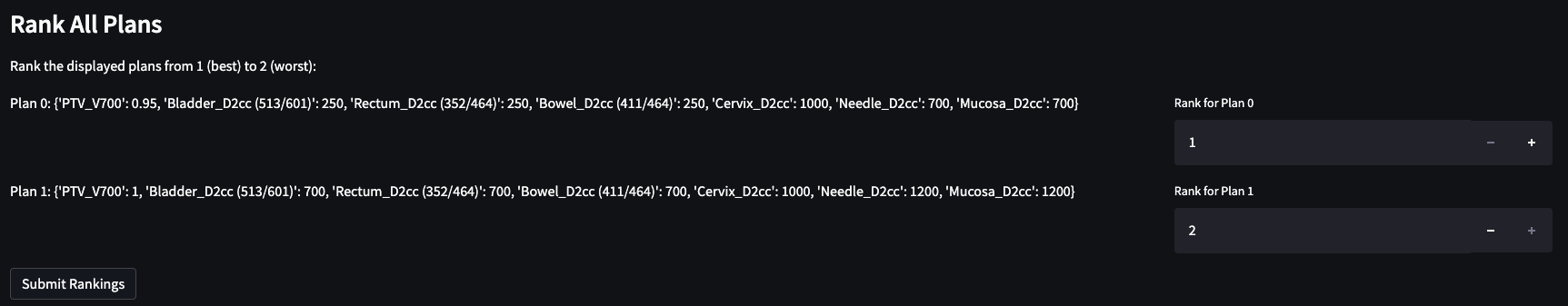}
    \caption{Ranking-Based Feedback Mechanism for Cervical Cancer Brachytherapy Treatment Planning Case Study. Only default treatment plan metrics are shown. The $\text{Needle}_{\text{D2cc}}$ objective is not used for evaluation.}
    \label{fig:case-study-ranking}
\end{figure}

To conduct this case study, we designed a simple interface for \methodname and all the baseline feedback mechanisms. This was built using Streamlit \citep{streamlit}. The six objectives which were being traded off include the following: $\text{PTV}_{\text{V700}}$, $\text{Bladder}_{\text{D2cc}}$, $\text{Rectum}_{\text{D2cc}}$, $\text{Bowel}_{\text{D2cc}}$, $\text{Cervix}_{\text{D2cc}}$, and $\text{Mucosa}_{\text{D2cc}}$. Figures of the interface are shown in Figures \ref{fig:case-study-soft-hard-bounds}, \ref{fig:case-study-pairwise}, \ref{fig:case-study-ranking}. We followed \citet{chen_almo_2026} in modeling the optimization problem; the decision variables served as inputs to an epsilon-constraint optimization program \citep{deufel_pnav_2020}. Due to the high-dimensionality of this domain and the need for guardrails, the decision variable space was discretized with a step size of 0.05 for $\text{PTV}_{\text{V700}}$, $\text{Bladder}_{\text{D2cc}}$, $\text{Rectum}_{\text{D2cc}}$, $\text{Bowel}_{\text{D2cc}}$ and a step size of 0.30 for the rest. The input range of each of the variables was determined by a simple variation of the Automated Planning Parameter Setup described in \citet{chen_almo_2026}.

We gathered three real retrospective anonymized patient cases from a local hospital from which we were able to determine the clinical reference treatment plan metrics. The six metric dimensions for each of the treatment plans were in slightly different regions of the tradeoff frontier, since the optimal treatment plan is highly dependent on patient physical conditions \citep{deufel_pnav_2020}.

\section{Additional Related Works}\label{app:related-works}

\textbf{Interactive Feedback Mechanisms.} Another stream of diverse feedback modalities includes learning from demonstrations \cite{abbeel_apprenticeship_2004, gonzalez_modeling_2018, ziebart_maximum_2008, shaikh_aligning_2024}. For multiple human inputs, \cite{myers_learning_2022} proposed multimodal rewards.

\textbf{Pareto Frontier Population Mechanisms.} Many MOO works focus on populating the entire Pareto frontier \citep{campigotto_active_2014, ponweiser_multiobjective_2008, emmerich_computation_2008, picheny_multiobjective_2015, hernandez-lobato_predictive_2016, zhang_multiobjective_2009}, sometimes via random scalarizations \citep{knowles_parego_2006, paria_flexible_2019} or by seeking sparse coverage, e.g., through level sets \citep{zuluaga_e-pal_2016, malkomes_beyond_2021}. We instead use soft and hard bounds to actively direct exploration to high-utility, preference-aligned areas, reducing computational and cognitive load.

\textbf{Building Confidence in AI Systems.}
\citet{broderick_toward_2023} developed a taxonomy of trust failure points in data analysis platforms, identifying perturbation analysis as a critical component. Within decision-making specifically, \citet{johnson_unifying_2024} studied the interplay between decision confidence, rewards, and priors. While \citet{insua_framework_1991, insua_sensitivity_1990} proposed a general framework for sensitivity analysis in discrete multi-objective decision-making, their work did not connect it to specific interactive feedback mechanisms or practical implementations. On the other hand, our framework offers an actionable multi-objective decision-making system where sensitivity analysis (\trustmethod) is directly integrated to enhance DM confidence in the soft-hard bounds feedback process. 

\textbf{Bayesian Decision Theory.} A related line of work frames exploration through Bayes-Adaptive Markov Decision Processes (BAMDPs) for non-myopic Bayes-optimal exploration, e.g., via meta-learning approaches such as variBAD \citep{zintgraf_varibad_2021}. Our setting differs: we have no task distribution to meta-train over (a single DM and single MOO problem), our unknowns are the DM's latent $(\pmb{\lambda}^*, \pmb{\alpha}^*)$ rather than MDP dynamics, and DM queries are expensive and budget-limited. Our regime instead aligns with active preference learning and preferential Bayesian optimization \citep{biyik_asking_2020, ozaki_multi-objective_2023}, where myopic, posterior-aware acquisition is standard. We follow this approach and extend it to soft-hard bounds with \trustmethod for confidence.

\section{Limitations}\label{app:limitations}
While \methodname demonstrates promising results, several limitations warrant discussion.

\textbf{Computational Cost.} The computational cost of maintaining and updating probabilistic models (e.g., Gaussian Processes for surrogate modeling of objectives, posteriors over $\pmb{\lambda}$ and $\pmb{\alpha}$) can become significant with a large number of objectives $L$ or many iterations $M$. While active sampling aims to be efficient, the overhead of GP training and acquisition function optimization in the local component, and sensitivity analysis in \trustmethod, could be a bottleneck for very high-dimensional problems or extremely rapid interaction requirements.

\textbf{Cognitive Load.} Although \methodname is designed to manage cognitive load, the interactive nature still requires DM engagement. The effectiveness of the framework relies on the DM's ability to provide meaningful feedback on soft-hard bounds. DM fatigue, inconsistency, or difficulty in articulating complex multi-level preferences (as indicated by the user study results in Figure \ref{fig:user-study-results}) could impact performance. Our assumption of a single bound modification per iteration simplifies feedback but might be restrictive for DMs wishing to express more complex preference changes simultaneously.

\textbf{Generalizability of Evaluations.} The goal of our simulation-based experiments was to mimic humans in decision-making scenarios as closely as possible. While we followed the standard in current literature, we acknowledge that humans often have complex decision-making patterns which are difficult to perfectly replicate, especially when it comes to novel feedback mechanisms such as soft-hard bounds. For instance, regarding our simulation setup, it is difficult to simulate the effects of \trustmethod. However, even without such simulations of \trustmethod, we observe enhanced performance. Finally, our user study (Section~\ref{sec:user-study}), while insightful, was conducted on a specific AI-generated image selection task with a particular DM pool (MTurk workers). Our quantitative proxy for confidence, Iteration Stop Efficiency (ISE), may possibly be confounded with the MTurk workers' fatigue or any time constraints. Further studies across diverse domains and with domain experts may help to fully assess the generalizability and practical utility of \methodname and \trustmethod.

\section{Broader Impacts}\label{app:broader-impacts}
The \methodname framework has the potential to positively impact high-stakes decision-making in various domains by enabling more efficient and trustworthy navigation of complex tradeoff spaces. In healthcare (e.g., brachytherapy treatment planning) or engineering design, where multiple competing objectives and expensive evaluations are common, our method could lead to better-informed decisions that more closely align with true stakeholder preferences, potentially improving outcomes and resource allocation. The explicit mechanism for building DM confidence (\trustmethod, Section~\ref{sec:t-mosh}) is particularly valuable, as it can increase user adoption and confidence in AI-assisted decision support systems.

However, potential negative societal impacts or misuse should also be considered. If the DM's underlying goals or specified bounds are unethical or biased, \methodname, like any optimization tool, could inadvertently facilitate undesirable outcomes. The accessibility of such advanced decision-support tools might also be a concern, potentially creating disparities if not made widely available or if requiring significant expertise to use effectively.

\begin{figure}[H]
    \centering
    \begin{tcolorbox}[colback=blue!5!white,title=GPT-4o-mini Input Prompt For Realism: Positive 1]\label{llm-degree-prompt-realism-pos-1}
    ``You are an expert in photorealistic prompt engineering for AI image generation. Your task is to expand the following subject into a detailed prompt that would guide an AI to generate an extremely photorealistic image. 
    \\
    Focus on:
    \begin{itemize}
        \item Photographic details (lighting, shadows, depth of field)
        \item Physical accuracy (correct anatomy, proportions, physics)
        \item Technical specifications (8K resolution, professional photography terms)
        \item Realistic materials, textures, and surface details
        \item Natural environment and context
    \end{itemize}
    
    Only output the expanded output prompt directly. Do NOT output anything else.
    \\
    \\
    \text{\{input prompt\}}''
\end{tcolorbox}
    \caption{Input prompt passed into GPT-4o-mini to generate the expanded prompt for an image with high realism, used for the realism objective preference dataset.}
    \label{fig:llm-degree-prompt-realism-pos-1}
\end{figure}

\begin{figure}[H]
    \centering
    \begin{tcolorbox}[colback=blue!5!white,title=GPT-4o-mini Input Prompt For Realism: Positive 2]\label{llm-degree-prompt-realism-pos-2}
    ``You are an expert in photorealistic prompt engineering for AI image generation. Your task is to expand the following subject into a detailed prompt that would guide an AI to generate a photorealistic image. 
    \\
    \\
    Only output the expanded output prompt directly. Do NOT output anything else.
    \\
    \\
    \text{\{input prompt\}}''
    \end{tcolorbox}
    \caption{Input prompt passed into GPT-4o-mini to generate the expanded prompt for an image with high realism, used for the realism objective preference dataset.}
    \label{fig:llm-degree-prompt-realism-pos-2}
\end{figure}

\begin{figure}[H]
    \centering
    \begin{tcolorbox}[colback=blue!5!white,title=GPT-4o-mini Input Prompt: Neutral]\label{llm-degree-prompt-realism-neutral}
    ``You are an expert in prompt engineering for AI image generation. Your task is to expand the following subject into a detailed prompt that would guide an AI to generate an image.
    \\
    \\
    Only output the expanded output prompt directly. Do NOT output anything else.
    \\
    \\
    \text{\{input prompt\}}
    ''
    \end{tcolorbox}    
    \caption{Input prompt passed into GPT-4o-mini to generate the expanded prompt for an image with neutral qualities, used for both the realism and color vividness preference datasets.}
    \label{fig:llm-degree-prompt-realism-neutral}
\end{figure}

\begin{figure}[H]
    \centering
    \begin{tcolorbox}[colback=blue!5!white,title=GPT-4o-mini Input Prompt For Realism: Negative]\label{llm-degree-prompt-realism-negative}
    ``You are a creative prompt engineer for AI image generation. Your task is to expand the following subject into a prompt that would guide an AI to generate a highly artistically stylized image.
    \\
    Focus on:
    \begin{itemize}
        \item Artistic styles (cartoon, illustration, abstract, etc.)
        \item Creative liberties with appearance and physics
        \item Stylized exaggerations or simplifications
        \item Fantasy or surreal elements
        \item Bold colors or unrealistic visual treatments
    \end{itemize}
    Only output the expanded output prompt directly. Do NOT output anything else.
    \\
    \text{\{input prompt\}}
    ''
    \end{tcolorbox}
    \caption{Input prompt passed into GPT-4o-mini to generate the expanded prompt for an image with low realism, used for the realism objective preference dataset.}
    \label{fig:llm-degree-prompt-realism-negative}
\end{figure}

\begin{figure}[H]
    \centering
    \begin{tcolorbox}[colback=blue!5!white,title=GPT-4o-mini Input Prompt For Color Vividness: Positive 1]\label{llm-degree-prompt-color-pos-1}
    ``You are an expert in vibrant color styling for AI image generation. Your task is to expand the following subject into a detailed prompt that would guide an AI to generate an image with extraordinarily bold, vivid, and diverse colors.
    \\
    Focus on:
    \begin{itemize}
        \item Bold color choices (saturated hues, neon elements, vibrant tones)
        \item Strategic color relationships (complementary pairs, triadic schemes, high-contrast combinations)
        \item Technical color specifications (maximum saturation, color vibrance, enhanced chroma)
        \item Unusual or unexpected color applications (color blocking, gradient shifts, non-traditional color choices)
        \item Lighting techniques that amplify color impact (rim lighting, color gels, dramatic color contrast)
        \item Color diversity across the image (wide color gamut, maximum color range, varied color temperature)
        \item Visual color impact (eye-catching color focal points, color that creates visual tension or harmony)
    \end{itemize}
    Only output the expanded output prompt directly. Do NOT output anything else.
    \\
    \text{\{input prompt\}}
    ''
    \end{tcolorbox}
    \caption{Input prompt passed into GPT-4o-mini to generate the expanded prompt for an image with high color vividness, used for the color vividness objective preference dataset.}
    \label{fig:llm-degree-prompt-color-pos-1}
\end{figure}

\begin{figure}[H]
    \centering
    \begin{tcolorbox}[colback=blue!5!white,title=GPT-4o-mini Input Prompt For Color Vividness: Positive 2]\label{llm-degree-prompt-color-pos-2}
    ``You are an expert in vibrant color styling for AI image generation. Your task is to expand the following subject into a detailed prompt that would guide an AI to generate an image with bold, vivid, and diverse colors.
    \\
    \\
    Only output the expanded output prompt directly. Do NOT output anything else.
    \\
    \\
    \text{\{input prompt\}}
    ''
    \end{tcolorbox}
    \caption{Input prompt passed into GPT-4o-mini to generate the expanded prompt for an image with high color vividness, used for the color vividness objective preference dataset.}
    \label{fig:llm-degree-prompt-color-pos-2}
\end{figure}

\begin{figure}[H]
    \centering
    \begin{tcolorbox}[colback=blue!5!white,title=GPT-4o-mini Input Prompt For Color Vividness: Negative]\label{llm-degree-prompt-color-negative}
    ``You are an expert in subdued color styling for AI image generation. Your task is to expand the following subject into a detailed prompt that would guide an AI to generate an image with exceptionally restrained, subtle, and minimal color presence.
    \\
    Focus on:
    \begin{itemize}
        \item Restrained color choices (desaturated tones, muted hues, subtle color variations)
        \item Monochromatic or limited color schemes (single-color palettes, tonal variations, grayscale with minimal accents)
        \item Technical color specifications (reduced saturation, lowered chroma, subtle tonal shifts)
        \item Sophisticated color restraint (elegant neutrals, barely-there color hints, delicate color transitions)
        \item Lighting techniques that minimize color impact (diffused lighting, low contrast, subtle gradations)
        \item Color harmony through absence (minimalist color approach, understated color presence, thoughtful color reduction)
        \item Visual impact through subtlety (tonal relationships, texture over color, emphasis on light and shadow rather than hue)
    \end{itemize}
    Only output the expanded output prompt directly. Do NOT output anything else.
    \\
    \text{\{input prompt\}}
    ''
    \end{tcolorbox}
    \caption{Input prompt passed into GPT-4o-mini to generate the expanded prompt for an image with low color vividness, used for the color vividness objective preference dataset.}
    \label{fig:llm-degree-prompt-color-negative}
\end{figure}

\begin{figure}[H]
    \centering
    \begin{tcolorbox}[colback=blue!5!white,title=GPT-4o-mini Input Prompt For Image Evaluation (Part 1 of 4)]
    ``Evaluate this image using the following calibrated scale for realism and color palette boldness/vividness. Your scores will help establish a Pareto frontier of optimal tradeoffs between these competing objectives.

    IMPORTANT SCORING GUIDELINES:
    \begin{itemize}
        \item Both scales use the FULL range from 0.0-100.0 with decimal precision (e.g., 76.3, 42.8)
        \item The two scores typically follow an approximate inverse relationship
        \item As one score increases, the other typically decreases
        \item Roughly speaking, realism score + color boldness score often falls between 75-110
        \item Rare exceptional images might score highly on both dimensions (but these are uncommon)
        \item Both extremely low scores on both dimensions indicates poor image quality, not a valid tradeoff
        \item Your scores should reflect relative positioning compared to other images in this domain
    \end{itemize}
    
    FUNDAMENTAL RELATIONSHIP:
    \begin{itemize}
        \item Realism focuses on accurate, photographic representation of reality with true-to-life colors
        \item Color boldness/vividness measures how extreme, vibrant, unusual, or attention-grabbing the color palette is
        \item For most images, increasing color boldness typically reduces realism, creating a tradeoff relationship
        \item The optimal frontier represents the best possible combinations of these competing values
        \item Images can be Pareto-optimal at different points (high realism/low color, balanced, high color/low realism)
    \end{itemize}
    
    REALISM SCALE (0.0-100.0):
    \begin{itemize}
        \item 95.0-100.0: Perfect photographic realism - completely indistinguishable from a professional photograph
        \item 85.0-94.9: High photorealism - extremely realistic with minimal stylistic elements
        \item 75.0-84.9: Strong realism - convincing real-world representation with occasional minor stylistic choices
        \item 65.0-74.9: Moderate realism - clearly recognizable with some noticeable stylistic elements
    \end{itemize} 
    
    ''
    \end{tcolorbox}
    \caption{Input prompt passed into GPT-4o-mini to evaluate the AI-generated images in terms of both realism and color vividness and obtain two numerical scores (part 1 of 4).}
    \label{fig:llm-image-eval-prompt-part-1}
\end{figure}

\begin{figure}[H]
    \centering
    \begin{tcolorbox}[colback=blue!5!white,title=GPT-4o-mini Input Prompt For Image Evaluation (Part 2 of 4)]
    ``
    \begin{itemize}
        \item 55.0-64.9: Balanced realism - equal emphasis on realism and interpretation
        \item 45.0-54.9: Limited realism - recognizable subjects with significant stylization
        \item 35.0-44.9: Partial realism - subjects are recognizable but heavily stylized
        \item 25.0-34.9: Minimal realism - heavily stylized with basic real-world references
        \item 15.0-24.9: Trace realism - predominantly abstract with subtle real-world hints
        \item 0.0-14.9: No realism - completely abstract, no recognizable elements
    \end{itemize}
    
    COLOR BOLDNESS/VIVIDNESS SCALE (0.0-100.0):
    \begin{itemize}
        \item 95.0-100.0: Extreme color intensity - extraordinarily vibrant, possibly fluorescent or neon
        \item 85.0-94.9: Highly vivid colors - very bold, saturated palette that immediately draws attention
        \item 75.0-84.9: Bold and vivid - distinctly vibrant palette with enhanced saturation
        \item 65.0-74.9: Moderately enhanced colors - noticeably vibrant with deliberate enhancement
        \item 55.0-64.9: Balanced color approach - moderately enhanced colors without overwhelming
        \item 45.0-54.9: Slightly enhanced colors - subtle enhancement beyond natural appearance
        \item 35.0-44.9: Natural color plus - largely natural palette with occasional subtle enhancement
        \item 25.0-34.9: Natural color range - colors as they would appear in standard photography
        \item 15.0-24.9: Subdued natural colors - slightly desaturated or muted natural colors
        \item 0.0-14.9: Minimal color presence - extremely desaturated or limited color range
    \end{itemize}
    
    RELATIONSHIP PATTERNS (not strict rules):
    \begin{itemize}
        \item Professional documentary photography often scores: Realism 85-95, Color Boldness 20-40
        \item Technical/scientific imagery often scores: Realism 80-95, Color Boldness 10-30
    \end{itemize}
      
    ''
    \end{tcolorbox}
    \caption{Input prompt to evaluate the AI-generated images in terms of both realism and color vividness and obtain scores (part 2 of 4).}
    \label{fig:llm-image-eval-prompt-part-2}
\end{figure}

\begin{figure}[H]
    \centering
    \begin{tcolorbox}[colback=blue!5!white,title=GPT-4o-mini Input Prompt For Image Evaluation (Part 3 of 4)]
    ``
    \begin{itemize}
        \item Commercial product photography often scores: Realism 80-90, Color Boldness 40-60
        \item Fashion photography often scores: Realism 75-85, Color Boldness 50-70
        \item Modern digital art often scores: Realism 40-70, Color Boldness 60-90
        \item Pop art style often scores: Realism 30-50, Color Boldness 80-100
        \item Psychedelic art often scores: Realism 20-40, Color Boldness 85-100
        \item Impressionist style often scores: Realism 40-60, Color Boldness 60-80
        \item Realistic paintings often score: Realism 60-80, Color Boldness 40-60
    \end{itemize}
    HOW TO EVALUATE COLOR BOLDNESS/VIVIDNESS:
    Consider these factors when scoring color boldness/vividness:
    \begin{enumerate}
        \item Saturation levels - how pure and intense the colors appear
        \item Color contrast - the degree of difference between adjacent colors
        \item Relationship to natural color - how much colors deviate from their appearance in nature
        \item Use of complementary colors - pairing colors from opposite sides of the color wheel
        \item Color harmony vs. discord - whether colors create pleasing relationships or intentional tension
        \item Overall color impact - how immediately the colors capture attention
        \item Unusual or unexpected color choices - colors applied in unconventional ways
        \item Color temperature extremes - very warm or cool color schemes, or dramatic contrasts between warm and cool
    \end{enumerate}
    
    ''
    \end{tcolorbox}
    \caption{Input prompt passed into GPT-4o-mini to evaluate the AI-generated images in terms of both realism and color vividness and obtain two numerical scores (part 3 of 4).}
    \label{fig:llm-image-eval-prompt-part-3}
\end{figure}

\begin{figure}[H]
    \centering
    \begin{tcolorbox}[colback=blue!5!white,title=GPT-4o-mini Input Prompt For Image Evaluation (Part 4 of 4)]
    ``
    EVALUATION INSTRUCTIONS:
    \begin{enumerate}
        \item First assess each dimension independently using their specific criteria
        \item Then consider if the scores reflect an appropriate tradeoff relationship
        \item Use decimal precision (e.g., 82.7 rather than just 83)
        \item For images that seem to defy the usual tradeoff, verify if it truly represents exceptional quality
        \item Consider how this image would position relative to others on a Pareto frontier
        \item Ensure you're using the full scale range appropriately
    \end{enumerate}
    
    Output format:
    \\
    **Realism Score: [0.0-100.0]**
    \\
    **Color Boldness/Vividness Score: [0.0-100.0]**
    \\
    **Reasoning:** [Your detailed explanation, referencing specific visual elements and explaining how the image represents a specific tradeoff point between these objectives]
    ''
    \end{tcolorbox}
    \caption{Input prompt passed into GPT-4o-mini to evaluate the AI-generated images in terms of both realism and color vividness and obtain two numerical scores (part 4 of 4).}
    \label{fig:llm-image-eval-prompt-part-4}
\end{figure}

\begin{figure}[H] %
    \centering
    \begin{tcolorbox}[
        colback=blue!5!white,
        title=MTurk User Study Instructions (Part 1 of 6): Introduction \& Overview,
    ]
    \textbf{Image Selection User Study: Instructions}

    \textbf{Introduction}

    Thank you for your interest in participating in our image selection user study. This document provides detailed instructions for the study tasks. Please read these instructions carefully as you will be asked to complete a short quiz on this material before beginning the actual study.

    \textbf{Study Overview}

    In this study, you will take on the role of a cover photo designer for an educational magazine. You will evaluate and select AI-generated images based on specific magazine requirements, using different feedback mechanisms to guide the selection process. Your participation will help us understand which image selection approaches are most effective for creative professionals.

    \textbf{Your Task}

    As a cover photo designer, you will:
    \begin{enumerate}
        \item Read a specific magazine's requirements for their next cover image.
        \item View several AI-generated images.
        \item Use a type of feedback to guide the system toward better images.
        \item Select a final image that best matches the magazine's requirements.
        \item Repeat steps 1-4 using four different types of feedback.
        \item Complete a survey about your experience.
    \end{enumerate}
    \end{tcolorbox}
    \caption{MTurk User Study Instructions: Introduction, Study Overview, and Task Summary.}
    \label{fig:mturk_instructions_part1}
\end{figure}

\begin{figure}[H]
    \centering
    \begin{tcolorbox}[
        colback=blue!5!white,
        title=MTurk User Study Instructions (Part 2 of 6): Image Attributes,
    ]
    \textbf{Key Image Attributes}

    The images in this study vary along two important dimensions:
    \begin{itemize}
        \item \textbf{Realism}
        \begin{itemize}
            \item Measures how photorealistic and accurate the image appears.
            \item Higher realism means the image looks more like a real photograph.
            \item Lower realism means the image appears more stylized or artistic.
        \end{itemize}
        \item \textbf{Color Vividness}
        \begin{itemize}
            \item Measures how bold, vibrant, and colorful the image appears.
            \item Higher color vividness means more saturated, eye-catching colors.
            \item Lower color vividness means more subdued, natural, or muted colors.
        \end{itemize}
    \end{itemize}

    \textbf{The Critical Tradeoff}

    Both high realism AND high color vividness are highly desired qualities for magazine covers. However, there is often a tradeoff between these objectives. Images that are extremely realistic typically have more muted colors, while images with vibrant colors often appear less realistic. Your challenge is to find the optimal balance based on the specific magazine's requirements. Please disregard factors in the image unrelated to the aforementioned two tradeoffs.
    \smallskip %

    \end{tcolorbox}
    \caption{MTurk User Study Instructions: Description of Key Image Attributes and the Realism-Vividness Tradeoff.}
    \label{fig:mturk_instructions_part2}
\end{figure}

\begin{figure}[H]
    \centering
    \begin{tcolorbox}[
        colback=blue!5!white,
        title=MTurk User Study Instructions (Part 3 of 6): Feedback Mechanisms I,
    ]
    \textbf{Feedback Mechanisms}

    You will try four different feedback approaches during the study:

    \textbf{1. Full Ranking}
    \begin{itemize}
        \item \textit{What it is:} You will order all images from most preferred to least preferred, according to the magazine’s requirements.
        \item \textit{How to use it:} Enter the numbers corresponding to each image in your preferred order.
        \item \textit{Example:} If you have five images and prefer them in the order 3, 1, 4, 2, 5, enter: \texttt{3,1,4,2,5}
        \item \textit{Important:} Make sure to format your input correctly with commas between numbers.
    \end{itemize}
    \smallskip

    \textbf{2. Partial Ranking}
    \begin{itemize}
        \item \textit{What it is:} You will rank only the top three images based on your strongest preferences according to the magazine’s requirements.
        \item \textit{How to use it:} Enter only the numbers for the top three images you have clear preferences about.
        \item \textit{Example:} If there is a total of five images and if you strongly prefer image 3, then image 1, then image 2, enter: \texttt{3,1,2}
    \end{itemize}
    \smallskip

    \textbf{3. Pairwise Preferences}
    \begin{itemize}
        \item \textit{What it is:} You will choose which image you prefer between two options at a time, according to the magazine’s requirements.
        \item \textit{How to use it:} Simply click on the image you prefer when presented with two options. Even if you do not like any of the images, please select the image you prefer more.
    \end{itemize}
    \end{tcolorbox}
    \caption{MTurk User Study Instructions: Description of Full Ranking, Partial Ranking, and Pairwise Preferences feedback mechanisms.}
    \label{fig:mturk_instructions_part3}
\end{figure}

\begin{figure}[H]
    \centering
    \begin{tcolorbox}[
        colback=blue!5!white,
        title=MTurk User Study Instructions (Part 4 of 6): Feedback Mechanisms II,
    ]
    \textbf{4. Soft \& Hard Bounds}
    \begin{itemize}
        \item \textit{What it is:} You'll set minimum acceptable levels (hard bounds) and preferred levels (soft bounds) for image attributes.
        \item \textit{How to use it:} For each image quality attribute (realism and color vividness):
        \begin{itemize}
            \item The left slider handle sets a \textbf{hard bound} (minimum acceptable level).
            \item The right slider handle sets a \textbf{soft bound} (preferred level).
        \end{itemize}
        \item \textit{What it represents:}
        \begin{itemize}
            \item Hard bounds: ``I reject any image below this threshold.''
            \item Soft bounds: ``I prefer images above this threshold, but will accept lower values.''
        \end{itemize}
        \item \textit{Important Notes:} 
        \begin{itemize}
            \item The first iteration of this feedback type will allow you to initialize the region of the bounds, based on our guess from the task description. You can perform multiple actions for this iteration.
            \item After initialization, the system only allows you to perform \textbf{one action} (one modification of one of the bounds) for each set of results. Please do not attempt to perform multiple actions for each iteration.
            \item After performing one action, images labeled as ``Adjacent Image'' may sometimes appear in the next set of results. These images are used to help you understand what might happen if the hard bounds are adjusted to be lower for one or both objectives. For example, due to the competing tradeoffs, if you were to decrease the hard bound for “Realism”, an image with a much higher value of “Color Vividness” may appear in the next set of results. An “Adjacent Image” with a much higher value of “Realism” lets you know that if you decrease the hard bound for “Color Vividness”, you might observe images such as that.
        \end{itemize}
    \end{itemize}
    \smallskip

    \end{tcolorbox}
    \caption{MTurk User Study Instructions: Description of the Soft \& Hard Bounds feedback mechanism.}
    \label{fig:mturk_instructions_part4}
\end{figure}

\begin{figure}[H]
    \centering
    \begin{tcolorbox}[
        colback=blue!5!white,
        title=MTurk User Study Instructions (Part 5 of 6): Study Process,
    ]
    \textbf{Study Process}

    For each feedback mechanism, you will:
    \begin{enumerate}
        \item Read the specific magazine requirements carefully. \textbf{This will change for each type of feedback!}
        \begin{itemize}
            \item \textit{Example prompt interpretation: The example task description in an image (not shown here) will be looking for an image which provides a balance between realism and color vividness, so we should strive to look for such images.}
        \end{itemize}
        \item Review the initial set of images.
        \begin{itemize}
            \item The realism and color vividness scores are displayed below each image.
        \end{itemize}
        \item Provide feedback using the current mechanism.
        \item Review new images generated based on your feedback.
        \item Using the same mechanism, provide feedback (i.e. pairwise, ranking, etc.) until you find a satisfactory image.
        \item Select your final image by clicking ``Select Final Image'' at the bottom of the page.
        \begin{itemize}
            \item This will lead you to a page with all of the images you’ve seen displayed. Please enter in your level of satisfaction (0-100) with the final image, in terms of how appropriate you feel it is for the task description, and click on “Select this image”.
        \end{itemize}
        \item Move on to the next feedback mechanism.
    \end{enumerate}
    \smallskip

    \end{tcolorbox}
    \caption{MTurk User Study Instructions: Step-by-step study process.}
    \label{fig:mturk_instructions_part5}
\end{figure}

\begin{figure}[H]
    \centering
    \begin{tcolorbox}[
        colback=blue!5!white,
        title=MTurk User Study Instructions (Part 6 of 6): Tips \& Final Notes,
    ]
    \textbf{Tips for Effective Image Selection}
    \begin{itemize}
        \item Keep the magazine's requirements in focus throughout the selection process.
        \item Explore different options before making your final selection.
        \item Consider the tradeoff between realism and color vividness based on the magazine's needs.
        \item Be consistent in your preferences.
        \item You can stop and select your final image whenever you feel you've found a suitable match.
    \end{itemize}
    \smallskip

    \textbf{Technical Requirements}
    \begin{itemize}
        \item Please use a desktop or laptop computer (not a mobile device).
        \item Ensure your screen brightness is adjusted for optimal image viewing.
        \item Complete the study in a single session if possible.
    \end{itemize}
    \smallskip

    Thank you for your participation! Your feedback will help us understand which approaches are most effective for image selection tasks in creative professional contexts.
    \smallskip

    \textbf{Note:} Each user may complete this task up to two times.
    \end{tcolorbox}
    \caption{MTurk User Study Instructions: Tips for selection, technical requirements, and concluding remarks.}
    \label{fig:mturk_instructions_part6}
\end{figure}

\begin{figure}[H] %
    \centering
    \begin{tcolorbox}[
        colback=blue!5!white, %
        title=MTurk User Study: Post-Mechanism Survey Questions for User Study,
    ]
    At the conclusion of interacting with each of the four feedback mechanisms, participants were presented with the following statements and asked to rate their agreement on a 5-point Likert scale (1-Strongly Disagree, 2-Disagree, 3-Neutral, 4-Agree, 5-Strongly Agree). The placeholder ``\{\}'' was replaced with the name of the specific feedback mechanism being evaluated (e.g., ``Soft-Hard Bounds'', ``Pairwise Preferences'').

    \bigskip %

    \begin{enumerate}
        \item \textbf{Mental Effort:} ``\{\} takes a lot of mental effort to use.''
        \item \textbf{Expressiveness:} ``\{\} allows me to fully express my preferences.''
        \item \textbf{Trust:} ``I trust that my results with \{\} are optimal for its task.''
    \end{enumerate}
    \end{tcolorbox}
    \caption{Survey questions presented to MTurk participants after their interaction with each feedback mechanism. The placeholder ``\{\}'' was substituted with the name of the mechanism in question.}
    \label{fig:mturk_survey_questions}
\end{figure}

\begin{figure}[H] %
    \centering
    \begin{tcolorbox}[
        colback=blue!5!white,
        title=MTurk User Study: Screening Quiz Questions (Part 1 of 2),
    ]
    Participants were required to answer the following screening questions correctly (at least 3 out of 5) to proceed with the user study.

    \bigskip

    \textbf{Question 1: Soft-Hard Bounds}
    
    \textit{Instruction:} In the soft-hard bounds feedback mechanism state below, please drag the circle on the left to a value less than 0.2. What does that action represent?
    
    \smallskip
    \textit{[Visual cue: A slider labeled ``Example: Hard (Left) and Soft (Right) Bounds for Realism'' is displayed, with range 0.0 to 1.0, and current values (0.3, 0.7).]}
    \smallskip

    \textit{Answer Options:}
    \begin{enumerate}
        \item[A:] I increased the soft bound, the preferred amount, for realism.
        \item[B:] I decreased the hard bound, the minimum acceptable amount, for realism down to below 0.2.
        \item[C:] I adjusted the medium bound, the neutral position for realism.
        \item[D:] Something else.
    \end{enumerate}
    \textit{(Correct Answer: B)}

    \bigskip %

    \textbf{Question 2: Soft-Hard Bounds}

    \textit{Instruction:} In the soft-hard bounds feedback mechanism state below, please drag the circle on right to a value greater than 0.85. What does that action represent?

    \smallskip
    \textit{[Visual cue: A slider labeled ``Example: Hard (Left) and Soft (Right) Bounds for Color Vividness'' is displayed, with range 0.0 to 1.0, and current values (0.4, 0.8).]}
    \smallskip

    \textit{Answer Options:}
    \begin{enumerate}
        \item[A:] I increased the hard bound, the minimum acceptable amount, for color vividness. I prefer to see images with color vividness above 0.85.
        \item[B:] I increased the soft bound, the preferred amount, for color vividness. I prefer to see images with color vividness above 0.85.
        \item[C:] I increased the medium bound, the neutral position, for realism.
        \item[D:] Something else.
    \end{enumerate}
    \textit{(Correct Answer: B)}

    \bigskip

    \end{tcolorbox}
    \caption{MTurk User Study Screening Quiz: Questions 1-3, focusing on understanding the Soft-Hard Bounds mechanism and related interface elements. Correct answers are indicated for reference.}
    \label{fig:mturk_screening_quiz_part1}
\end{figure}

\begin{figure}[H] %
    \centering
    \begin{tcolorbox}[
        colback=blue!5!white,
        title=MTurk User Study: Screening Quiz Questions (Part 2 of 2),
    ]
    \textbf{Question 3: Soft-Hard Bounds' Adjacent Images}

    \textit{Instruction:} What does the image labeled as ``Adjacent Image'' represent in the example below?

    \smallskip
    \textit{[Visual cue: Three images are displayed side-by-side. Image 1 caption: ``Image 1 -- Realism: 0.40 Color Vividness 0.86''. Image 2 caption: ``Image 2 -- Realism: 0.65 Color Vividness 0.63''. Image 3 caption: ``Adjacent Image 3 -- Realism: 0.95 Color Vividness 0.30''. Below these, two disabled sliders are shown for Realism and Color Vividness, both set to (0.4, 0.8).]}
    \smallskip
    
    \textit{Answer Options:}
    \begin{enumerate}
        \item[A:] The image that is most preferred in the ranking.
        \item[B:] An example of an image I might see if I adjust the color vividness hard bound to be slightly lower.
        \item[C:] An image that is adjacent to the current image in the ranking.
        \item[D:] An example of an image I might see if I adjust the realism soft bound to be slightly higher.
    \end{enumerate}
    \textit{(Correct Answer: B, based on typical interpretation of adjacent images in such contexts)}

    \bigskip
    
    \textbf{Question 4: Partial Ranking}

    \textit{Instruction:} In ``Partial Ranking'', what does the task require you to do?
    
    \smallskip
    \textit{Answer Options:}
    \begin{enumerate}
        \item[A:] Rank all images from best to worst.
        \item[B:] Rank only your top three preferred images.
        \item[C:] Select the single best image.
        \item[D:] Group images into categories.
    \end{enumerate}
    \textit{(Correct Answer: B)}

    \bigskip %

    \textbf{Question 5: Image Attributes}

    \textit{Instruction:} According to the study instructions, what is the relationship between ``Realism'' and ``Color Vividness'' in the images?
    
    \smallskip
    \textit{Answer Options:}
    \begin{enumerate}
        \item[A:] They are completely unrelated attributes.
        \item[B:] Higher realism always leads to higher color vividness.
        \item[C:] They often represent a tradeoff - increasing one may decrease the other.
        \item[D:] They are different names for the same quality.
    \end{enumerate}
    \textit{(Correct Answer: C)}

    \end{tcolorbox}
    \caption{MTurk User Study Screening Quiz: Questions 4-5, assessing understanding of the Partial Ranking mechanism and the core image attribute tradeoff. Correct answers are indicated for reference.}
    \label{fig:mturk_screening_quiz_part2}
\end{figure}

\begin{figure}[htbp]
    \centering
    \begin{tcolorbox}[
        colback=blue!5!white,
        title=User Study Task Description: Einstein for "Historical Science Review",
        width=\textwidth %
    ]
    \textbf{\Large Task Description: Selecting a Cover Photo for "Historical Science Review"}

    \medskip
    \textbf{\large Your Task}
    
    You are selecting an appropriate photograph of Albert Einstein for the cover of "Historical Science Review," a respected educational magazine read by university professors, historians of science, and serious science enthusiasts.

    \medskip
    \textbf{\large About This Publication}
    
    "Historical Science Review" is dedicated to presenting accurate historical perspectives on scientific developments and the scientists behind them. The magazine takes pride in its meticulous attention to historical detail and is used as a reference in university courses and by science historians. The publication has built its reputation on scholarly integrity and historical authenticity.

    \medskip
    \textbf{\large What We're Looking For}
    
    The ideal cover image should:
    \begin{itemize}
        \item \textbf{Maintain complete historical accuracy} - We need an image that represents Albert Einstein as he truly appeared, with authentic details and no modern artistic interpretation.
        \item \textbf{Present authentic documentary quality} - The image should resemble a genuine historical photograph with the visual characteristics of photography from Einstein's era.
        \item \textbf{Use historically appropriate coloration} - Colors should appear natural and historically accurate, as they would in a well-preserved photograph from that time period.
        \item \textbf{Convey scholarly authority} - The overall appearance should suggest a serious academic publication rather than a popularized interpretation.
    \end{itemize}
    \end{tcolorbox}
    \caption{User study task description for selecting an Albert Einstein cover image for "Historical Science Review." This task targets a high realism ($\sim$90) and low color vividness ($\sim$5) tradeoff point.}
    \label{fig:task_einstein_hr_lc}
\end{figure}

\begin{figure}[htbp]
    \centering
    \begin{tcolorbox}[
        colback=blue!5!white,
        title=User Study Task Description: Einstein for "Mind Blast Science",
        width=\textwidth
    ]
    \textbf{\Large Task Description: Selecting a Cover Image for "Mind Blast Science"}

    \medskip
    \textbf{\large Your Task}

    You are selecting an eye-catching cover image of Albert Einstein for the newest issue of "Mind Blast Science," an educational magazine designed specifically for middle and high school students to spark interest in scientific concepts and historical scientific figures.

    \medskip
    \textbf{\large About This Publication}

    "Mind Blast Science" uses bold visual design and contemporary artistic approaches to make science exciting and accessible to young audiences. The magazine aims to capture students' attention in today's visually stimulating digital environment. It competes with video games, social media, and other highly visual entertainment for students' attention, and its covers need to stand out dramatically on both physical and digital platforms.

    \medskip
    \textbf{\large What We're Looking For}

    The ideal cover image should:
    \begin{itemize}
        \item \textbf{Make a powerful visual statement} - We need an image with electrifying colors that immediately grab students' attention when scrolling through content or passing a magazine stand.
        \item \textbf{Use color as the primary engagement tool} - The cover should feature bold, vibrant, or unusual color combinations that create maximum visual impact and convey the excitement of scientific discovery.
        \item \textbf{Prioritize creative interpretation over historical accuracy} - The image should capture Einstein's iconic essence and what he represents (genius, creativity, scientific revolution) rather than strictly depicting how he appeared in historical photographs.
        \item \textbf{Appeal to contemporary youth aesthetic} - The style should feel fresh and current, with visual treatments that resonate with students accustomed to bold graphic design in their media.
    \end{itemize}
    \end{tcolorbox}
    \caption{User study task description for selecting an Albert Einstein cover image for "Mind Blast Science." This task targets a low realism ($\sim$15) and high color vividness ($\sim$85) tradeoff point.}
    \label{fig:task_einstein_lr_hc}
\end{figure}

\begin{figure}[htbp]
    \centering
    \begin{tcolorbox}[
        colback=blue!5!white,
        title=User Study Task Description: Einstein for "Science Spectrum" Magazine,
        width=\textwidth
    ]
    \textbf{\Large Task Description: Selecting a Cover Photo for "Science Spectrum" Magazine}

    \medskip
    \textbf{\large Your Task}

    You are selecting an engaging cover image of Albert Einstein for the upcoming issue of "Science Spectrum," a popular science magazine that appeals to a broad audience of science enthusiasts from college students to working professionals.

    \medskip
    \textbf{\large About This Publication}

    "Science Spectrum" bridges the gap between academic rigor and popular accessibility. The magazine presents scientific concepts and history in an approachable way while maintaining intellectual credibility. Its visual style balances authenticity with contemporary design elements to create an appealing presentation that respects both historical accuracy and modern aesthetic sensibilities.

    \medskip
    \textbf{\large What We're Looking For}
    
    The ideal cover image should:
    \begin{itemize}
        \item \textbf{Be clearly recognizable as Einstein} - The image should maintain enough realistic elements that readers can immediately identify Einstein, with reasonably accurate facial features and characteristic appearance.
        \item \textbf{Incorporate engaging visual elements} - While respecting the historical figure, the image should include thoughtful color enhancement that makes it visually appealing on newsstands or digital platforms.
        \item \textbf{Balance authenticity with visual interest} - We want an image that honors who Einstein was while incorporating enough creative visual treatment to feel fresh and contemporary.
        \item \textbf{Appeal to a modern audience} - The cover should feel neither like a dry historical document nor an overly stylized artistic interpretation, but rather a balanced approach that respects history while engaging today's readers.
    \end{itemize}
    \end{tcolorbox}
    \caption{User study task description for selecting an Albert Einstein cover image for "Science Spectrum" Magazine. This task targets a medium realism and medium color vividness tradeoff point.}
    \label{fig:task_einstein_med_med}
\end{figure}

\begin{figure}[htbp]
    \centering
    \begin{tcolorbox}[
        colback=blue!5!white,
        title=User Study Task Description: Obama for "Presidential Archives Quarterly",
        width=\textwidth
    ]
    \textbf{\Large Task Description: Selecting a Cover Photo for "Presidential Archives Quarterly"}

    \medskip
    \textbf{\large Your Task}

    You are selecting an appropriate photograph of Barack Obama for the cover of "Presidential Archives Quarterly," a prestigious educational journal published by the National Historical Society that documents and analyzes American presidential administrations.

    \medskip
    \textbf{\large About This Publication}

    "Presidential Archives Quarterly" serves as a definitive historical record consulted by political scientists, historians, and academic institutions worldwide. The publication maintains strict standards for historical accuracy and documentary authenticity, presenting unembellished historical documentation for scholarly analysis and historical preservation.

    \medskip
    \textbf{\large What We're Looking For}

    The ideal cover image should:
    \begin{itemize}
        \item \textbf{Maintain historical precision} - We need an image that depicts President Obama exactly as he appeared during his presidency, with complete accuracy in facial features, expressions, and physical details.
        \item \textbf{Present authentic documentary quality} - The image should have the characteristics of professional photojournalism or official White House photography, preserving the authentic visual record of his presidency.
        \item \textbf{Use minimal coloration} - Colors should appear minimal, as they would in official presidential photography, without enhancement or creative color styling.
        \item \textbf{Convey presidential gravitas} - The overall appearance should reflect the dignity and historical significance of the presidential office, suitable for archival documentation.
    \end{itemize}
    \end{tcolorbox}
    \caption{User study task description for selecting a Barack Obama cover image for "Presidential Archives Quarterly." This task targets a high realism ($\sim$90) and low color vividness ($\sim$5) tradeoff point.}
    \label{fig:task_obama_hr_lc}
\end{figure}

\begin{figure}[htbp]
    \centering
    \begin{tcolorbox}[
        colback=blue!5!white,
        title=User Study Task Description: Obama for "American Leadership Today",
        width=\textwidth
    ]
    \textbf{\Large Task Description: Selecting a Cover Photo for "American Leadership Today"}

    \medskip
    \textbf{\large Your Task}

    You are selecting an engaging cover image of Barack Obama for the upcoming issue of "American Leadership Today," a popular educational magazine that explores political leadership for a diverse audience ranging from college students to professionals.

    \medskip
    \textbf{\large About This Publication}

    "American Leadership Today" bridges academic analysis and accessible education about American political leaders. The magazine maintains credibility while using contemporary visual approaches to engage readers. Its visual style balances authenticity with modern design elements to appeal to readers who value both substance and visual engagement.

    \medskip
    \textbf{\large What We're Looking For}

    The ideal cover image should:
    \begin{itemize}
        \item \textbf{Be clearly recognizable as Obama} - The image should maintain his distinctive features and characteristic appearance while allowing for thoughtful creative interpretation.
        \item \textbf{Incorporate engaging color elements} - While preserving Obama's recognizable likeness, the image should include purposeful color choices that enhance visual appeal and emotional resonance.
        \item \textbf{Balance historical representation with contemporary style} - We want an image that acknowledges Obama's historical significance while incorporating enough creative visual treatment to feel fresh and relevant.
        \item \textbf{Appeal to both substance-focused and visually-oriented readers} - The cover should feel neither like a dry historical document nor an overly stylized artistic piece, but rather a thoughtful balance that respects the subject while engaging modern viewers.
    \end{itemize}
    \end{tcolorbox}
    \caption{User study task description for selecting a Barack Obama cover image for "American Leadership Today." This task targets a medium realism ($\sim$55) and medium color vividness ($\sim$60) tradeoff point.}
    \label{fig:task_obama_med_med}
\end{figure}

\begin{figure}[htbp]
    \centering
    \begin{tcolorbox}[
        colback=blue!5!white,
        title=User Study Task Description: Obama for "Future Leaders: Youth Politics Magazine",
        width=\textwidth
    ]
    \textbf{\Large Task Description: Selecting a Cover Image for "Future Leaders: Youth Politics Magazine"}

    \medskip
    \textbf{\large Your Task}

    You are selecting an eye-catching cover image of Barack Obama for the newest issue of "Future Leaders," an educational magazine designed to spark political interest and civic engagement among middle and high school students.

    \medskip
    \textbf{\large About This Publication}

    "Future Leaders" uses bold visual design and contemporary artistic approaches to make political history and civic education exciting and relevant to young audiences. The magazine competes for attention in students' visually saturated digital environment and needs covers that immediately capture interest on school library shelves, social media feeds, and digital platforms.

    \medskip
    \textbf{\large What We're Looking For}

    The ideal cover image should:
    \begin{itemize}
        \item \textbf{Create maximum visual impact} - We need an image with electrifying colors that immediately grab students' attention—think vibrant blues, bold reds, striking color contrasts, or unexpected color treatments.
        \item \textbf{Use color as a storytelling element} - The color palette should convey energy, inspiration, and the transformative potential of political engagement, making Obama's legacy feel relevant to today's youth.
        \item \textbf{Prioritize artistic interpretation over strict realism} - The image should capture Obama's iconic status and what he represents symbolically (change, hope, historical significance) rather than simply documenting how he appeared in photographs.
        \item \textbf{Connect with youth visual culture} - The style should incorporate graphic elements, color treatments, and visual approaches that resonate with students familiar with bold social media aesthetics and contemporary digital art.
    \end{itemize}
    \end{tcolorbox}
    \caption{User study task description for selecting a Barack Obama cover image for "Future Leaders: Youth Politics Magazine." This task targets a low realism ($\sim$15) and high color vividness ($\sim$85) tradeoff point.}
    \label{fig:task_obama_lr_hc}
\end{figure}

\end{document}